\documentclass[10pt,twocolumn,letterpaper]{article}

\usepackage{cvpr}              % To produce the CAMERA-READY version
\usepackage{wrapfig}
\usepackage[accsupp]{axessibility} % Improves PDF readability for those with disabilities.

\usepackage{float}
\usepackage{fancybox}
\usepackage{calc}
\usepackage{mathdots}
\usepackage{graphicx}
\usepackage{listings}
\usepackage{amsthm, amsmath, amssymb}
\usepackage{bm}
\usepackage{mathrsfs}
\usepackage{algorithm}
\usepackage{algorithmicx}
\usepackage{algpseudocode}
\usepackage{amsmath}
\usepackage{booktabs}
\usepackage{amsfonts}
\usepackage{array}
\usepackage{multirow}
\usepackage{dsfont}
\usepackage{caption}
\usepackage[font=small]{caption}
\usepackage{subcaption}
\usepackage{color}
\usepackage{bbm}
\usepackage{booktabs}
\usepackage{microtype}
\usepackage{threeparttable}
\usepackage{tabularx}
\newcommand{\D}{{\rm{d}}}

\newcommand{\la}{{\lambda}}

\newcommand{\x}{{\bm{x}}}
\newcommand{\y}{{\bm{y}}}
\newcommand{\z}{{\bm{z}}}

\newcommand{\q}{{\bm{q}}}

\newcommand{\I}{{\bm{I}}}
\newcommand{\s}{{\bm{s}}}
\newcommand{\m}{\bm{\mu}}
\newcommand{\The}{{\bm{\theta}}}
\newcommand{\eps}{{\bm{\epsilon}}}
\newcommand{\hx}{{\bm{\hat{x}}}}

\renewcommand{\geq}{\geqslant}
\renewcommand{\leq}{\leqslant}

\definecolor{cvprblue}{rgb}{0.21,0.49,0.74}
\usepackage[pagebackref,breaklinks,colorlinks,allcolors=cvprblue]{hyperref} 

\def\paperID{} % *** Enter the Paper ID here`
\def\confName{CVPR}
\def\confYear{2026}

\title{Elucidating the SNR-t Bias of Diffusion Probabilistic Models}

\author{
Meng Yu$^{1,2}$\thanks{Work done during the internship at AMAP Alibaba Group.},\ 
Lei Sun$^{2}$\thanks{Project leader.},\ 
Jianhao Zeng$^{2}$,\ 
Xiangxiang Chu$^{2}$,\ 
Kun Zhan$^{1}$\thanks{Corresponding author. Email: kzhan@lzu.edu.cn} \\
$^{1}$Lanzhou University,\ $^{2}$AMAP Alibaba Group
}

\begin{document}
\maketitle
\begin{abstract}
Diffusion Probabilistic Models have demonstrated remarkable performance across a wide range of generative tasks. However, we have observed that these models often suffer from a Signal-to-Noise Ratio-timestep (SNR-t) bias. This bias refers to the misalignment between the SNR of the denoising sample and its corresponding timestep during the inference phase. Specifically, during training, the SNR of a sample is strictly coupled with its timestep. However, this correspondence is disrupted during inference, leading to error accumulation and impairing the generation quality. We provide comprehensive empirical evidence and theoretical analysis to substantiate this phenomenon and propose a simple yet effective differential correction method to mitigate the SNR-t bias. Recognizing that diffusion models typically reconstruct low-frequency components before focusing on high-frequency details during the reverse denoising process, we decompose samples into various frequency components and apply differential correction to each component individually. Extensive experiments show that our approach significantly improves the generation quality of various diffusion models (IDDPM, ADM, DDIM, A-DPM, EA-DPM, EDM, PFGM++, and FLUX) on datasets of various resolutions with negligible computational overhead. The code is at https://github.com/AMAP-ML/DCW.

%Diffusion models have demonstrated remarkable performance across various generative tasks. However, we observe that they commonly suffer from a Signal-to-Noise Ratio–timestep (SNR-t) bias. SNR-t bias refers to the misalignment between the actual SNR of the model input sample and its corresponding timestep during inference phase. Theoretically, during training, the SNR of the sample is strictly coupled with its timestep. In contrast, this correspondence is disrupted during inference, leading to error accumulation and degradation of generation quality. We provide comprehensive empirical evidence and theoretical analysis to substantiate this phenomenon, and we propose a simple yet effective differential correction method to alleviate the SNR-t bias. Furthermore, considering that diffusion models tend to reconstruct the low-frequency component before focusing on high-frequency details during reverse denoising process, we decompose samples into different frequency components and perform differential correction on each component separately. Extensive experiments show that our approach significantly improves the generation quality of various diffusion models (IDDPM, ADM, DDIM, A-DPM, EA-DPM, EDM, DiT, PFGM++, and SDXL) on datasets of various resolutions with negligible computational overhead.
\end{abstract}    
\section{Introduction}
\label{sec:intro}

Due to their outstanding performance, Diffusion Probabilistic Models (DPMs)~\cite{sohl2015deep,ho2020denoising,song2021scorebased} have achieved remarkable success in various generative tasks, including image~\cite{dhariwal2021diffusion,rombach2022high}, audio~\cite{kongdiffwave, chen2021wavegrad}, and video~\cite{zheng2024open, blattmann2023align, khachatryan2023text2video} generation. DPMs typically consist of two processes. In the forward process, a data sample is progressively perturbed by Gaussian noise until it becomes the standard Gaussian noise. In the reverse process, DPMs iteratively denoise from the standard Gaussian noise to generate the clean data sample. Despite their significant success, we identify that DPMs suffer severely from a Signal-to-Noise Ratio–timestep (SNR-t) bias.

% \begin{figure}[!t]
% \centering
%    \includegraphics[width=0.60\linewidth]{figures/SNR-t_bias.pdf} 
% \caption{The schematic diagram of SNR-t bias in DPMs. During training, the SNR of perturbed samples is strictly tied to timesteps. During inference, however, due to network prediction errors and discretization errors in numerical solutions, the SNR of predicted samples no longer matches the preset timesteps.}
% \label{fig1.snr_t_bias}
% \end{figure}

\begin{figure*}[t]
\centering
\begin{subfigure}[b]{0.33\linewidth}
    \centering
    \includegraphics[width=\linewidth]{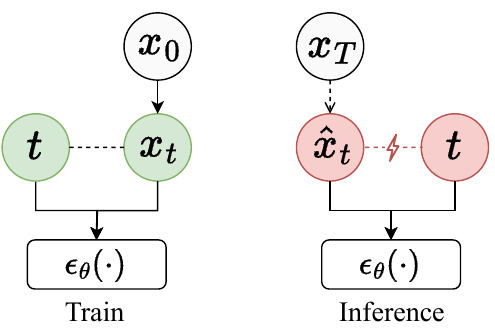}
    \caption{Schematic of SNR-t bias in DPMs.}
    \label{fig1a:snr_t_bias}
\end{subfigure}%
\begin{subfigure}[b]{0.33\linewidth}
    \centering
    \includegraphics[width=\linewidth]{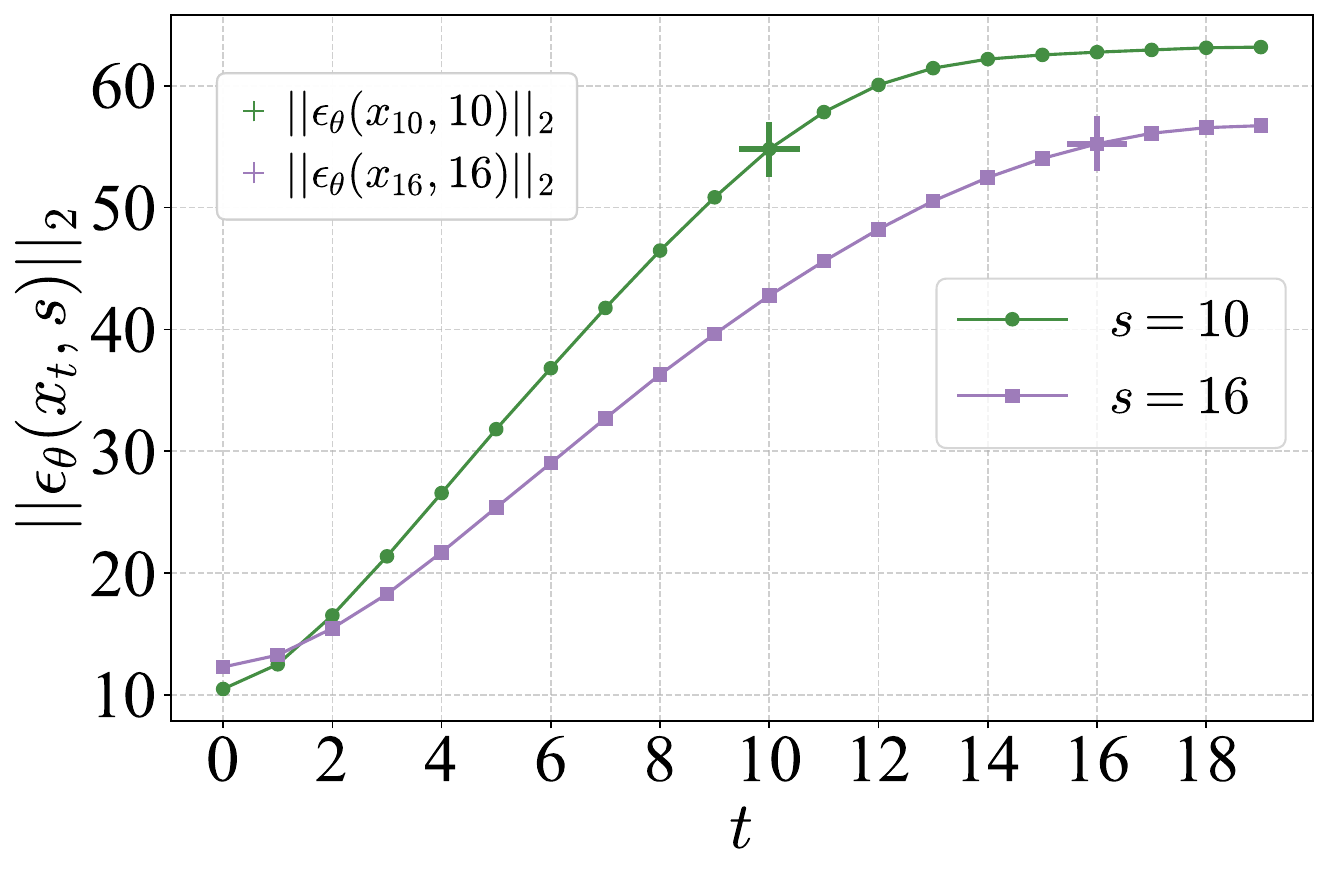}
    \caption{SNR-t bias causes inaccurate predictions.}
    \label{fig1b:network}
\end{subfigure}%
\begin{subfigure}[b]{0.33\linewidth}
    \centering
    \includegraphics[width=\linewidth]{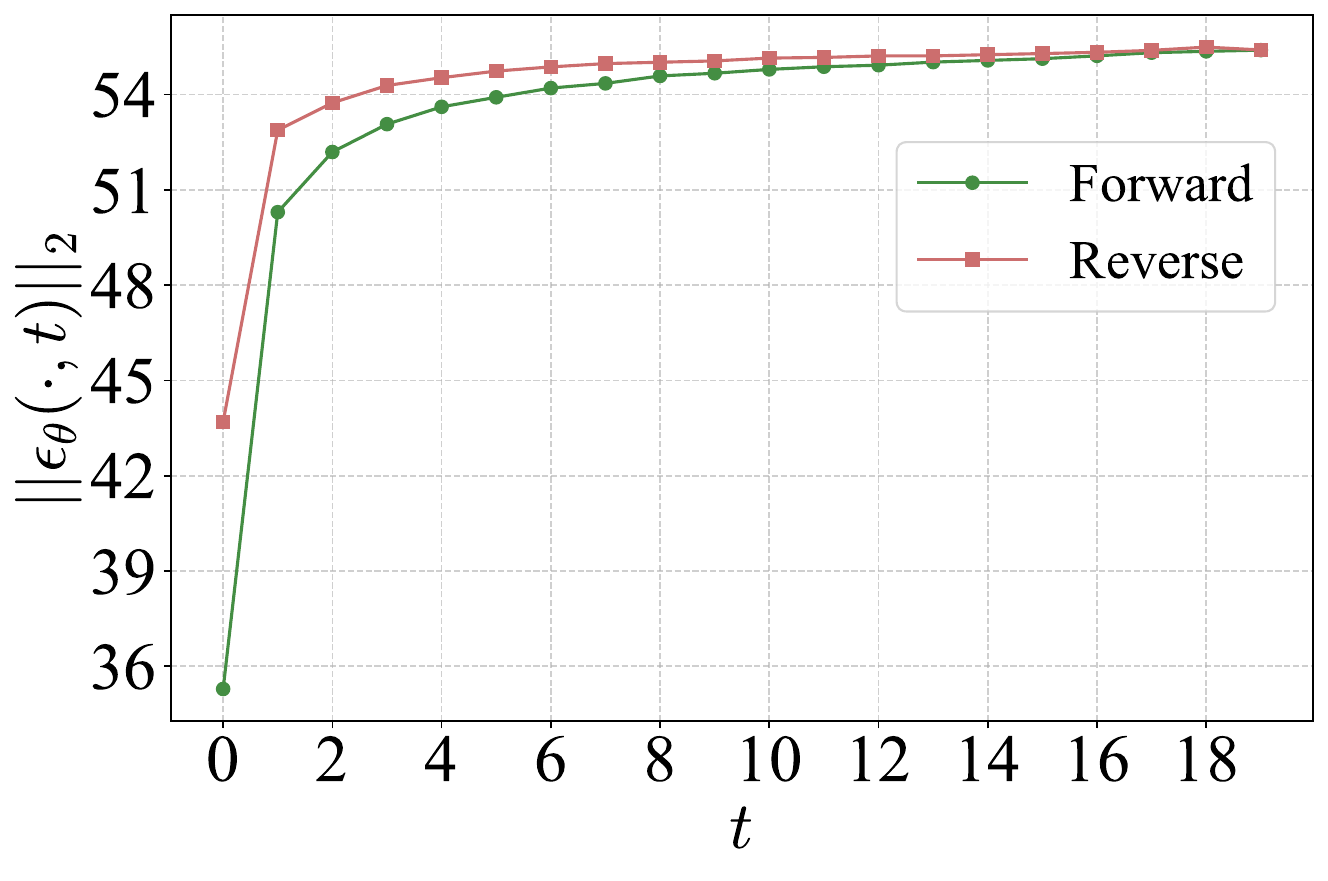}
    \caption{The reverse process exhibits lower SNR.}
    \label{fig1c:sample_bias}
\end{subfigure}
\caption{(a) During training, the SNR of perturbed sample $\x_t$ is strictly tied to timestep $t$. However, during inference, due to network prediction errors and discretization errors in numerical solvers, the SNR of predicted sample $\hx_t$ no longer matches the preset timestep $t$. 
(b) shows the network output $||\eps_\The(\x_t,s)||_2$ when a trained network $\eps_\The(\cdot,s)$ with fixed timestep $s$ receives samples $\x_t$ with mismatched SNR (samples are generated via forward process using Eq.~\ref{eq2:forward_onestep} with different $t$). 
(c) shows the network output $||\eps_\The(\cdot,t)||_2$ using forward samples and reverse predicted samples, respectively. $||\eps_\The(\hx_t,t)||_2$ is always larger than $||\eps_\The(\x_t,t)||_2$, which indicates that predicted samples exhibit lower SNR compared to forward samples at the same timestep. See the experiment details of (b) and (c) in Sec.~\ref{sec:4}.}

\label{fig2:time-snr}
\end{figure*}

The SNR-t bias refers to the misalignment between the SNR of predicted samples and their assigned timesteps during inference. Specifically, during training, the neural network is conditioned on both the perturbed sample and the corresponding timestep, establishing a deterministic correspondence between the SNR of the sample and the timestep. However, during inference, due to cumulative errors arising from both the model's predictions~\cite{kim2023refining} and the numerical solvers~\cite{song2021scorebased,lu2022dpmsolver}, the denoising trajectory inevitably deviates from the ideal path, causing a misalignment between the SNR of the predicted sample and its designated timestep, as shown in Fig.~\ref{fig1a:snr_t_bias}. 
Unlike previously studied exposure bias~\cite{ning2023input}, which focuses on inter-sample discrepancies, the SNR-t bias emphasizes the misalignment between the predicted sample and its corresponding timestep. We argue that the SNR-t bias is a more fundamental bias that can induce exposure bias and is prevalent in current DPMs.

%\caption{Harm of SNR-t bias.}
%\caption{Sample bias caused by SNR-t bias.}

%Expectation of $||\eps_\The(\cdot,t)||_2$ in the forward and reverse process on CIFAR10~\cite{krizhevsky2009learning}, with ADM~\cite{dhariwal2021diffusion} as the baseline model. 

% In (b) and (c), the network is frozen, with $T=20$.}

%In the forward process, a series of perturbed samples $\{\x_0, \x_1,\cdots,\x_t,\cdots,\x_T\}$ are generated via Eq.~\ref{eq2:forward_onestep}, then fed into the fixed-timestep network $\epsilon_\theta(\cdot,s)$, with $s$ as 10 and 16. Finally, expectation of $||\eps_\The(\x_t,s)||_2$ is computed.

We provide a comprehensive experimental analysis and theoretical justification for SNR-t bias. Our experiments reveal two key findings: (1) the network demonstrates significantly inaccurate predictions when processing samples with mismatched SNR and timesteps. Specifically, as illustrated in Fig.~\ref{fig1b:network}, samples with lower SNR tend to make the network produce larger noise predictions, while those with higher SNR yield smaller noise predictions. (2) Reverse denoising samples often exhibit lower SNR compared to their corresponding forward samples at the same timestep, as shown in Fig.~\ref{fig1c:sample_bias}.
These findings lead to a notable conclusion: the SNR-t bias severely degrades the model performance and often manifests as lower SNR for the corresponding timestep during the denoising process. To investigate the underlying mechanisms, we analyze the reverse process of DPMs and provide a theoretical proof of this bias, thereby offering a robust theoretical justification for our findings.

% from yum
%We provide comprehensive experimental analysis and theoretical justification for SNR-t bias. We obtain two critical findings: (1) when a sub-network $\eps_\The(\cdot,s)$ with fixed-timestep $s$ receives the perturbed sample whose SNR does not match the SNR corresponding to the timestep $s$, the prediction output of network exhibits significant errors, as shown in Fig.~\ref{fig1b:network}. Specifically, if SNR of the input sample is lower than the matched SNR for timestep, the network outputs higher amplitude, otherwise, it outputs lower amplitude. (2) The predicted sample has a lower SNR than the perturbed sample at the same timestep. Specifically, at each timestep, the network output for predicted samples is consistently higher than that for perturbed samples, as shown in Fig.~\ref{fig1c:sample_bias}. Combined with the finding (1), it is natural to conclude that the predicted samples have a lower SNR. Moreover, through analytical modeling of reverse process of DPMs, we derive the analytical expression for SNR of the predicted sample, further consolidating the conclusion of the lower SNR.

% As we all know, network of DPM can be viewed as a collection of $T$ sub-networks sharing the same parameters \cite{go2023addressing,hang2023efficient}
% Furthermore, through rigorous mathematical modeling, we derive the relationship between the reverse prediction and forward perturbation distributions at the same timestep, confirming reverse predicted samples indeed have lower SNR.

To mitigate the SNR-t bias, a natural solution is to align the distribution of reverse samples, which tends to have lower SNR, with the corresponding distribution of forward samples. Given the complexity of existing DPM frameworks, training or fine-tuning approaches would incur significant costs. Instead, we propose a dynamic differential correction method in the wavelet domain, which leverages the model's inherent capabilities to alleviate the bias without additional training.
Specifically, at each denoising step, we obtain the reconstruction sample, which directly predicts the clean sample from the current predicted sample. By analytically modeling the prediction distribution and the reconstruction distribution, we find that their difference signal contains gradient information that can guide the biased predicted sample toward the ideal perturbed sample. We incorporate this differential signal into each denoising step to ensure the predicted distribution aligns more closely with the perturbed distribution, thereby effectively mitigating the bias.

Additionally, to improve the correction effect, we introduce the method into the wavelet domain, allowing it to correct different frequency components of samples separately. This approach leverages the unique denoising characteristics~\cite{yi2024towards, qian2024boosting} of DPMs, which initially emphasize the reconstruction of low-frequency contours during the reverse process before focusing on restoring high-frequency details. Meanwhile, we assign dynamic weight coefficients to the correction operations for different components. By applying targeted corrections for varying frequency components at different stages of the denoising process, we achieve significant improvements in corrections and overall performance. Notably, our method can further enhance the performance of improved models~\cite{ning2023input,ningelucidating,yu2025frequency} for exposure bias, which highlights the significance and superiority of our proposed problem and method. In summary, our contributions are:

% from yum
%To mitigate the SNR-t bias, a natural solution is to align the reverse predicted distribution—which tends to have lower SNR—with the corresponding forward perturbed distribution. Given the complexity of existing DPM frameworks, training or fine-tuning based approaches would incur high costs. Instead, we aim to leverage DPM’s own capabilities to correct the bias. Specifically, at each step of the denoising process, we always obtain a reconstruction estimate, which predicts the original data sample given the predicted sample. By analytically modeling the predicted and reconstructed distributions, we discover that their difference signal contains gradient information that can push the biased predicted distribution toward the ideal perturbed distribution. We inject this differential signal into each denoise step to guide the predicted distribution to align with the perturbed distribution.

%Furthermore, we introduce the differential correction method into the wavelet domain to separately correct different components of samples, which stems from the unique denoising characteristics of DPMs in frequency domain. Specifically, DPMs first focus on the reconstruction of low-frequency contours during reverse process, and then concentrate on restoring high-frequency details of samples. Thus, specialized correction for different components at different stages is a highly promising approach. Meanwhile, we also assign dynamic weight coefficients to the correction operations of different components to further enhance the performance. In summary, our contributions are:
\begin{itemize}
  \item We identify the SNR-t bias in DPMs and provide comprehensive experimental analysis and theoretical proof.
  \item
  We propose a dynamic differential correction method in wavelet domain to effectively alleviate the SNR-t bias.
  \item Our method is training-free and plug-and-play, effectively improving the generation quality of various DPMs. It can also be extended to other bias-correction models with significant gains and negligible computation.
\end{itemize}

% 我们将这种信噪比-时间步在训练和推理之间的不匹配问题视为一个关键发现，并为其提供清晰的见解、严密的证明以及简单有效的解决方案。

% 首先，我们观察到经过绑定信噪比-时间步训练的神经网络具备信噪比水平感知能力，通过在前向过程中向特定时间步的网络输入不同信噪比水平的扰动样本，我们得到一个巧妙的结论：推理阶段的预测样本相较于相同时间步的扰动样本往往具备更低的信噪比；然后，通过对神经网络的偏差进行有证据的严谨建模，我们从理论上推导出推理阶段预测样本与训练的扰动样本之间的解析数学关系，进一步证明先前的更低信噪比的见解；最后，

% 有趣的是，我们的发现和见解还可以为先前大量关于曝光偏差、Unet微调和加速采样的工作提供理论依据和可解释性证据。同时，我们也关注到TimeOUT提出丢弃扩散模型的时间步条件，总之，我们的贡献是：

% 1.我们发现扩散模型推理过程存在信噪比-时间步漂移现象，通过可靠的实验和严密的推导，我们证实了反向过程预测分布的低信噪比漂移规律；

% 2.结合低信噪比漂移结论和扩散模型的去噪特性，我们提出一种即插即用、无需训练的基于频域感知的差分矫正方法，以几乎忽略不计的计算代价显著提升多种扩散模型的生成质量；

% 3.我们的发现和结论可以为大量关于曝光偏差、Unet微调和时间步学习的工作提供理论理论依据，我的方法可以集成到各种偏差矫正的改进模型上，进一步提升矫正效果。
%-------------------------------------------------------------------------

\section{Related Work}
\label{sec:formatting}
This section first reviews the development of DPMs, followed by some recent works on bias analysis in DPMs.

The foundational theory of DPMs is introduced by DPM~\cite{sohl2015deep}, with major advances brought by DDPM~\cite{ho2020denoising}. ADM~\cite{dhariwal2021diffusion} employs classifier guidance to make DPMs outperform GANs~\cite{goodfellow2014generative}, while EDM~\cite{Karras2022edm} systematically explores the training and inference design space to further boost generation quality and efficiency of DPMs. Notably, ODE-based DPMs~\cite{lu2022dpmsolver,zhou2024fast,zhao2024unipc,dockhorn2022genie}, knowledge distillation-based DPMs~\cite{salimans2022progressive,liu2023instaflow,meng2023distillation,luhman2021knowledge}, and consistency models~\cite{song2023consistency,song2024improved,lu2025simplifying,lei2026there} are widely studied. Meanwhile, DPMs have advanced downstream tasks like text-to-image models~\cite{rombach2022high,blackforestlabs2024flux,lan2025flux,chen2026layer}, image editing~\cite{mengsdedit,couairondiffedit,parmar2023zero}, and super-resolution generation~\cite{saharia2022image,li2022srdiff,he2026texts}. Furthermore, USP~\cite{chu2025usp}, SY-TDM~\cite{miao2025shining}, FE2E~\cite{wang2025editor}, S$^2$-Guidance~\cite{chen2026stochastic}, and ADE-COT~\cite{qu2026scale} improve DPMs from different perspectives.

% and A-DPM derives the analytical form of the reverse variance in DPMs to address theoretical gaps

% The theoretical underpinnings of DPMs were laid by DPM~\cite{sohl2015deep}, with major progress later made by DDPM~\cite{ho2020denoising}. The introduction of classifier guidance in ADM~\cite{dhariwal2021diffusion} marked the first time DPMs outperformed GANs~\cite{goodfellow2014generative} in generation quality. SDE~\cite{song2021scorebased} established a unified framework for DPMs using stochastic differential equations, while DDIM~\cite{songdenoising} enabled faster sampling by allowing step skipping. Analytic-DPM~\cite{baoanalytic} derived the optimal reverse process variance theoretically, and EDM~\cite{Karras2022edm} systematized the architectural design space of DPMs, further boosting their performance. Moreover, ODE-driven DPMs~\cite{lu2022dpmsolver,zhou2024fast,zhao2024unipc,zheng2023fast,dockhorn2022genie}, DPMs integrated with knowledge distillation~\cite{salimans2022progressive,liu2023instaflow,meng2023distillation,luhman2021knowledge,zheng2023fast}, and consistency models~\cite{song2023consistency,song2024improved,lu2025simplifying} have been extensively explored, greatly accelerating inference and improving sample fidelity.

Research on exposure bias is closely related to our work. Exposure bias in DPMs refers to the sample mismatch between training and sampling. ADM-IP \cite{ning2023input} re-perturbs training data to imitate the discrepancies in inference, exposing the model to possible prediction errors. MDSS \cite{ren2024multi} interprets exposure bias as deviations between predicted samples and network outputs and adopts a multi-step denoising schedule to reduce it. EP-DDPM \cite{li2023error} derives an upper bound on accumulated errors and incorporates it as a retraining regularizer to lessen the bias. While these models require retraining, TS-DPM \cite{li2024alleviating} and ADM-ES \cite{ningelucidating} offer training-free, plug-and-play alternatives. In addition, MCDO \cite{yao2025manifold}, DPM-AT \cite{zhang2025antiexposure}, DPM-AE \cite{wangimproved}, BMGDM \cite{yu2025bias}, and DPM-FR \cite{yu2025frequency} also analyze and mitigate this bias from different perspectives.

Exposure bias acts across samples, whereas SNR-t bias arises between samples and timesteps.

%In subsequent sections, we will use existing improvement works targeting exposure bias as baseline models, and our methods can further enhance the generation quality.From a geometric perspective, MCDO \cite{yao2025manifold} enforces a manifold constraint to reduce prediction errors, while DPM-AT \cite{zhang2025antiexposure} learns lightweight prompts to correct denoising trajectories.
\section{Background}
\label{sec:3}
In this section, we review the preliminaries of DPMs.
% \subsection{Diffusion Probabilistic Models}

DPMs generally comprise a forward process and a reverse process, with both formulated as Markov chains. Given a target data distribution $q(\x_0)$ and a variance schedule $\beta_t$, the forward process is defined as
\begin{equation}
q(\x_{1:T}|\x_0)=\prod_{t=1}^T q(\x_t|\x_{t-1}),
\label{eq1:forward_markov}
\end{equation}
where $q(\x_t|\x_{t-1})=\mathcal{N}(\x_t;\sqrt{1-\beta_t}\x_{t-1},\beta_t\I)$. 
Utilizing the attributes of the Gaussian distribution, the perturbed sample $\x_t$ is directly expressed in a closed form as the conditional distribution $q(\x_t|\x_0)$:
\begin{equation}
\x_t=\sqrt{\bar{\alpha}_t}\x_0+\sqrt{1-\bar{\alpha}_t}\eps_t,
\label{eq2:forward_onestep}
\end{equation}
where ${\alpha}_t=1-\beta_t$, $\bar{\alpha}_t = \prod_{i=1}^t\alpha_i$, and $\eps_t\sim \mathcal{N}(\bm{0},\I)$.  
Then, by applying Bayes’ theorem, the corresponding posterior distribution can be expressed as:
\begin{equation}
q(\x_{t-1}|\x_t,\x_0)=\mathcal{N}(\tilde{\m}_t(\x_t,\x_0),\tilde{\beta}_t\I),
\label{eq3:real_posterior}
\end{equation}
where
\[
\tilde{\m}_{t}=\frac{\sqrt{\bar{\alpha}_{t-1}}\beta_{t}}{1-\bar{\alpha}_{t}}\x_{0}
+\frac{\sqrt{\alpha_{t}}(1-\bar{\alpha}_{t-1})}{1-\bar{\alpha}_{t}}\x_{t},
\quad
\tilde{\beta}_t=\frac{1-\bar{\alpha}_{t-1}}{1-\bar{\alpha}_t}\beta_t.
\]A neural network $p_{\The}(\x_{t-1}|\x_t)=\mathcal{N}(\x_{t-1};\mu_{\The}(\x_t,t),\sigma_t\I)$ is employed to approximate $q(\x_{t-1}|\x_t,\x_0)$, which aims to minimize $D_{\mathrm{KL}}(q(\x_{t-1}|\x_t,\x_0)\| p_{\The}(\x_{t-1}|\x_t))$.  
Through reparameterization, we are able to obtain:
\begin{equation}
\begin{aligned}
\m_{\The}(\x_t,t)
&= \frac{\sqrt{\bar{\alpha}_{t-1}}\beta_{t}}{1-\bar{\alpha}_{t}}\, \x^0_\The(\x_t,t)
+ \frac{\sqrt{\alpha_{t}}(1-\bar{\alpha}_{t-1})}{1-\bar{\alpha}_{t}}\, \x_{t} \\
&= \frac{1}{\sqrt{\alpha_t}} \left(\x_t - \frac{1 - \alpha_t}{\sqrt{1 - \bar{\alpha}_t}} \eps_\The(\x_t, t) \right),
\end{aligned}
\end{equation}
where $\x^0_\The(\x_t,t)$ represents the reconstruction of $\x_0$ given $\x_t$, and $\eps_\The(\cdot)$ denotes the noise prediction network. Specifically, the relationship between the two is:
\begin{equation}
\x^0_\The(\x_t,t)=\frac{\x_t-\sqrt{1-\bar{\alpha}_t}\eps_\The(\x_t,t)}{\sqrt{\bar{\alpha}_t}}.
\label{eq5:reconstruction}
\end{equation}
Finally, we obtain the concise training objective:
\begin{equation}
\mathcal{L}_{\mathrm{simple}}
=\mathbb{E}_{t,\x_0,\eps_t\sim\mathcal{N}(\mathbf{0},\I)}\left[\|\eps_{\The}(\x_t,t)-\eps_t\|_2^2\right].
\label{eq6:loss_function}
\end{equation}
Once the noise prediction network is trained to convergence, we can start from a standard Gaussian noise, perform step-by-step iterative denoising via $p_{\The}(\x_{t-1}|\x_t)$, and ultimately generate the clean data sample.

% Once the noise prediction network converges, reverse sampling can be performed starting from a standard Gaussian distribution using $p_{\The}(\x_{t-1}|\x_t)$ to generate new data samples.

\section{SNR-t Bias}
\label{sec:4}
In this section, we present the specific definition of SNR-t bias and elaborate on two key findings.

The DPM takes the perturbed sample $\x_t$ and the timestep $t$ as input during training, as shown in Fig.~\ref{fig1a:snr_t_bias}, and the SNR of $\x_t$ is directly determined by the timestep $t$:
\begin{equation}
    \text{SNR}(t) = \bar{\alpha}_t/(1-\bar{\alpha}_t).
\end{equation}
Due to the forced binding between the SNR of samples and timesteps during the training phase, the network $\eps_\The(\cdot,t)$ is proficient in accurately predicting samples with a corresponding $\text{SNR}(t)$. But what happens if the network $\eps_\The(\cdot,s)$ receives a sample $\x_t$ with a mismatched $\text{SNR}(t)$?

To validate this, we design and conduct an experiment to assess the network predictions using samples with the mismatched SNR. Specifically, we select the ADM~\cite{dhariwal2021diffusion} model as our baseline model and utilize 2,000 samples from the CIFAR-10~\cite{krizhevsky2009learning} dataset. We first fix the timestep as $s$ for the network $\eps_\The(\cdot,s)$ and then generate a series of forward perturbed samples $\{\x_0, \x_1,\cdots,\x_t,\cdots,\x_T\}$ using Eq.~\ref{eq2:forward_onestep}. These perturbed samples are subsequently fed into the network $\eps_\The(\cdot,s)$, after which we compute their mean squared norm and present the results in Fig.~\ref{fig1b:network}.

%conduct an analysis using an ingenious sliding window experiment. Specifically, we select ADM as the baseline model, CIFAR10 as the test dataset, 2000 as the sampling batch, and then perform subsequent operations in sequence. Firstly, we fix the timestep component of the network as $s$. Then, we obtain a series of forward noised samples $\{\x_0, \x_1,\cdots,\x_t,\cdots,\x_T\}$ via Eq.~\ref{eq2:forward_onestep}, and feed them all into the network $\eps_\The(\cdot,s)$ with fixed timestep $s$. Finally, we compute their mean squared norms $\{\|\eps_\The(\x_0,s)\|_2^2,\ldots,\|\eps_\The(\x_T,s)\|_2^2\}$, as shown in Fig.~\ref{fig1b:network}.
% \noindent\textbf{Key Finding 1.}
\paragraph{Key Finding 1.} \textit{The network produces significantly inaccurate predictions when processing samples with mismatched SNR and timesteps.} 
As illustrated in Fig.~\ref{fig1b:network}, this bias exhibits a specific pattern: for the fixed timestep $s$, when handling the input sample $\x_t$ with a lower SNR, where $t > s$, the network tends to overestimate the predicted output. In contrast, when dealing with the sample $\x_t$ with a higher SNR, the predicted output is typically underestimated. In summary, samples with lower SNR lead the network to produce larger noise predictions, while those with higher SNR result in smaller noise predictions.

With the Key Finding 1 highlighting the significant performance degradation caused by SNR-t bias in DPMs, a natural subsequent question arises: how does SNR-t bias exactly manifest during the actual denoising process?

%Now, we turn to the practical denoising process and introduce the definition and manifestation of SNR-t bias.

%The inference process of DPMs can be interpreted as a numerical solution process based on SDE or ODE, thus inevitably introducing discretization errors from numerical solving. Meanwhile, the neural network of DPMs always have unavoidable prediction errors. Therefore, due to these two types of errors, the reverse denoising trajectory often deviates from the ideal path, leading to a mismatch between the actual SNR of reverse predicted samples $\hx_t$ and the preset timestep $t$. 

The inference process in DPMs can be understood as a numerical solution to a Stochastic Differential Equation (SDE) or an Ordinary Differential Equation (ODE), which inevitably introduces discretization errors during numerical computations. Additionally, the neural network within DPMs is subject to inherent prediction errors. Consequently, these two types of errors can cause the reverse denoising trajectory to deviate from the ideal path, resulting in a mismatch between the actual SNR of the reverse predicted samples $\hat{\x}_t$ and the designated timestep $t$. Thus, the actual reverse denoising process can be expressed as:
\begin{equation}
\hat{\x}_{t-1} = \frac{1}{\sqrt{\alpha_t}} \left(\hat{\x}_t - \frac{1 - \alpha_t}{\sqrt{1 - \bar{\alpha}_t}} \eps_\The(\hat{\x}_t, t) \right) + \sigma_t \z.
\label{eq7:actual_inference}
\end{equation}

To further investigate the manifestations of SNR-t bias, we adopt the same experimental setup as in Fig.~\ref{fig1b:network} and conduct the following comparative experiment.
%Thus, the SNR-t Bias refers to the mismatch between the actual SNR of predicted samples $\hx_t$ and their assigned timestep $t$ during the inference phase. However, we aim to further explore the manifestations of the SNR-t bias. 
%Thus, we adopt the same baseline model, dataset, and sampling batch as in the first experiment, and conduct the following comparative evaluations. 
(1) We generate perturbed samples $\{\x_1,\x_2,\ldots,\x_T\}$ via Eq.~\ref{eq2:forward_onestep}, and feed $\x_t$ and timestep $t$ into the network to obtain $\eps_\The(\x_t,t)$.  
(2) Then, we initialize 2,000 samples of standard Gaussian noise and perform iterative denoising via Eq.~\ref{eq7:actual_inference} to obtain samples $\{\hx_1,\hx_2,\ldots,\hx_T\}$ and corresponding network outputs $\eps_\The(\hx_t,t)$.  
(3) Finally, we compute and plot the expectation of $\ell_2$ norms $||\eps_\The(\x_t,t)||_2^2$ and $||\eps_\The(\hx_t,t)||_2^2$, as shown in Fig.~\ref{fig1c:sample_bias}. Particularly, similar experiments were also conducted in ADM-ES~\cite{ningelucidating}, and we provide the evidence of the differences, together with more robust analyses in Appendix~\ref{app:a}. Building on this, we derive the second key finding:

\paragraph{Key Finding 2.} \textit{Reverse denoising samples often exhibit lower SNR compared to their corresponding forward samples at the same timestep.} Fig.~\ref{fig1c:sample_bias} shows that for any timestep $t$, the mean $\ell_2$ norm of reverse predictions $\eps_\The(\hx_t,t)$ consistently exceeds that of forward predictions $\eps_\The(\x_t,t)$. The Key Finding 1 shows that the network tends to produce an overestimated output when processing samples with lower SNR. Therefore, we have reason to conclude that the denoising sample $\hx_t$ generally maintains a lower SNR than the forward perturbed sample $\x_t$ at the same timestep, leading to overestimated predictions at each denoising step.

% We first provide an intuitive visualization of SNR-t bias. Diffusion models are trained with timesteps and corresponding perturbed samples of specific signal-to-noise ratio (SNR) levels. For a given timestep, if the input sample’s SNR deviates from that of the training perturbations, this deviation should be reflected in the network output.  

% To explore this, we conduct a \emph{network sliding experiment} to examine the SNR-awareness of $\eps_\The(\cdot,t)$. Using the same setup, we fix a timestep $t=s$ and feed all perturbed samples $\{\x_1,\x_2,\ldots,\x_T\}$ into $\eps_\The(\cdot,s)$, obtaining $\{\eps_\The(\x_1,s),\ldots,\eps_\The(\x_T,s)\}$. We compute their mean squared norms $\{\|\eps_\The(\x_1,s)\|_2^2,\ldots,\|\eps_\The(\x_T,s)\|_2^2\}$ for $s=10$ and $s=16$, as shown in Fig.~\ref{fig1b:network}.

% As seen in Fig.~\ref{fig1b:network}, for $\eps_\The(\cdot,10)$, when $\{\x_t|t<10\}$, we have $\|\eps_\The(\x_t,10)\|<\|\eps_\The(\x_{10},10)\|$, while for $\{\x_t|t>10\}$, $\|\eps_\The(\x_t,10)\|>\|\eps_\The(\x_{10},10)\|$. The same holds for $s=16$.  

% Intuitively, since $\eps_\The(\cdot,t)$ is trained only on inputs with a specific SNR, it can accurately predict samples at that SNR level. When the input SNR deviates, the network exhibits biased predictions: it overestimates noise for lower-SNR inputs and underestimates it for higher-SNR ones.

\section{Method}
In this section, we first analytically model the reverse process of DPMs and derive the analytical form of the SNR-t bias, providing a comprehensive theoretical basis for this bias. Then, based on the theoretical analysis, we propose a simple yet effective differential correction method to mitigate the SNR-t bias, thereby improving the generation quality of DPMs. Finally, by incorporating the denoising laws of DPMs, we introduce differential correction into the wavelet domain and design a specialized weighting strategy to further enhance the correction effect.
\subsection{Theoretical Proof}
\label{sec:4.1}
% Previous bias studies~\cite{ningelucidating,li2024alleviating}in DPMs rely on a strong assumption, which is
% \begin{equation}
%     \x^0_\The(\x_t,t) = \x_0 + \phi_t \eps_t,
% \end{equation}
% where $\x^0_\The(\x_t,t)$ denotes the reconstruction network that predicts $\x_0$ from $\x_t$ (as defined in Eq.~\ref{eq5:reconstruction}), and $\phi_t$ is a scalar coefficient related to the diffusion timestep.  
% However, we challenge this assumption both empirically and theoretically. Specifically, we propose a more reasonable assumption:
For the theoretical analysis of bias in DPMs, prior works have proposed two distinct assumptions. ADM-ES~\cite{ningelucidating} and TS-DPM~\cite{li2024alleviating} propose the following formulation:
\begin{equation}
    \x^0_\The(\x_t,t) = \x_0 + \phi_t \eps_t,
\end{equation}
where $\eps_t \sim \mathcal{N}(\bm{0},\I)$, with $\phi_t$ a scalar coefficient. LA-DPM~\cite{zhang2023lookahead} and DPM-FR~\cite{yu2025frequency} propose another formulation: $\x^0_\The(\x_t,t) = \gamma_t\x_0 + \phi_t \eps_t$, with $\gamma_t$ also a scalar coefficient. Unfortunately, these prior assumptions are overly strong and lack sufficient theoretical grounding and empirical validation. Furthermore, there is a clear discrepancy in the coefficient of $\x_0$ between the two hypotheses. To address this issue, we conduct extensive theoretical and experimental analyses in this work, and ultimately decide to adopt the second hypothesis for our subsequent analysis.
% To explore the intrinsic modeling behavior of $\x^0_\The(\x_t,t)$, we compute $\x^0_\The(\x_t,t)$ and $\x^0_\The(\hat{\x}_t,t)$ using Eq.~\ref{eq5:reconstruction}, corresponding respectively to $\x_t$ and $\hat{\x}_t$ from the experiment described in \S\ref{sec:4.1}.  
% We then perform wavelet decomposition and norm analysis on these reconstructed samples.  
% As illustrated in Figs.~\ref{fig4c:x0_ll} and \ref{fig4d:x0_lh}, neither $\x_t$ nor $\hat{\x}_t$ can perfectly reconstruct $\x_0$ at any timestep.  
% Moreover, across all frequency components and timesteps, the reconstruction $\x^0_\The(\hat{\x}_t,t)$ obtained from predicted noisy samples consistently exhibits lower magnitude than $\x^0_\The(\x_t,t)$ reconstructed from perturbed noisy samples.  
% This key observation motivates a more physically meaningful and generalizable assumption about $\x^0_\The(\x_t,t)$.
% \noindent\textbf{Theorem 4.1 }
\paragraph{Assumption 5.1.} \textit{During both the forward and reverse processes, the reconstruction sample $\x^0_\The(\x_t,t)$ can be expressed in terms of the original data $\x_0$ as follows:}
\begin{equation}
    \x^0_\The(\hat{\x}_t,t) = \gamma_t \x_0 + \phi_t \eps_t,
\label{eq10:assumption}
\end{equation}
\textit{where $0 < \gamma_t \leq 1$, $\phi_t < M$, and $M$ denotes a uniform upper bound constant across all timesteps.}

\textit{Sketch of Proof.} $\x_\theta^0(\x_t, t)$ is the reconstruction output for predicting $\x_0$ given $\x_t$, which is also known as the posterior mean $\mathbb{E}[\x_0|\x_t]$~\cite{umdon}. Based on Tweedie's formula \cite{umdon} and the L2-norm loss function \cite{Karras2022edm}, DPMs tend to predict the mean value of the target data. Thus, $\x_\theta^0$ can be viewed as the mean prediction $\bar{\x}_0$ of $\x_0$. By the variance identity $\mathbb{E}[\| \x_0 \|^2] = \| \bar{\x}_0 \|^2 + \text{Var}(\|\x_0\|)$ and the non-negativity of variance, we get
\begin{equation*}
    \| \bar{\x}_0 \|^2 \leq \mathbb{E}[\| \x_0 \|^2].
\end{equation*}
Since the expectation of a constant is itself, we can obtain:
\begin{equation}
    \mathbb{E}[||{\x}_\theta^0||^2] \leq \mathbb{E}[\| \x_0 \|^2].
    \label{var}
\end{equation}
The assumption $\x_\theta^0 = \x_0 + \phi_t \eps_t$ implies that $\mathbb{E}[||\x_\theta^0||^2] = \mathbb{E}[||\x_0||^2] + \phi_t^2$. Obviously, this conflicts with Eq.~\ref{var}. Thus, a more accurate formulation is given in Eq.~\ref{eq10:assumption}, where $\gamma_t<1$  denotes energy and information loss during the reconstruction of $\x_0$. In particular, ASBGM~\cite{miglani2025analysing} also provides indirect evidence for this view. Furthermore, more theoretical and experimental evidence is provided in Appendix \ref{app:b}.

Based on Assumption 5.1, we can derive the analytical form of the SNR for $\hx_t$ in the reverse process:
% \paragraph{Theorem 4.2}

% \noindent\textbf{Theorem 4.2 }
\paragraph{Theorem 5.1.} \textit{For a specific timestep $t$ in the reverse denoising process of DPMs, the SNR of the biased denoising sample $\hx_t$ is given by:}
\begin{equation}
    \text{SNR}(t)=\hat{\gamma}^2_{t}{\bar{\alpha}_{t}}/\big(1 - \bar{\alpha}_{t} +
        (
            \frac{\sqrt{\bar{\alpha}_{t}}\beta_{t+1}}{1-\bar{\alpha}_{t+1}}\phi_{t+1})^2
        \big),
\label{eq12:actural_snr}
\end{equation}
\textit{where $0 < \hat{\gamma}_t \leq 1$ and $\phi_{t+1}$ is derived from the reconstruction model $\x^0_\The(\hat{\x}_{t+1},t+1)$ in Eq.~\ref{eq10:assumption}.}

\textit{Sketch of Proof.} For the sake of brevity of the formula, we present the denoising process from $\hx_t$ to $\hx_{t-1}$. By substituting the reconstruction model in Eq.~\ref{eq5:reconstruction} into the inverse denoising Eq.~\ref{eq7:actual_inference}, we can obtain:
\begin{equation}
    \hat{\x}_{t-1} = 
    \frac{\sqrt{\overline{\alpha}_{t-1}}\beta_{t}}{1-\overline{\alpha}_{t}} \x^0_\The(\x_t,t)
    + \frac{\sqrt{\alpha_{t}}(1-\overline{\alpha}_{t-1})}{1-\overline{\alpha}_{t}}\x_{t}
    + \sqrt{\tilde{\beta}_t}\eps_1,
\label{eq13.re_actual_reverse}
\end{equation}
where $\eps_1 \sim \mathcal{N}(\mathbf{0},\I)$.  
Substituting Eqs.~\ref{eq10:assumption} and \ref{eq2:forward_onestep} into Eq.~\ref{eq13.re_actual_reverse} yields the analytical form of $\hat{\x}_{t-1}$:
\begin{equation}
    \hat{\x}_{t-1} =
    \hat{\gamma}_{t-1}\sqrt{\bar{\alpha}_{t-1}}\x_0 +
    \sqrt{
        1 - \bar{\alpha}_{t-1} +
        \left(
            \frac{\sqrt{\bar{\alpha}_{t-1}}\beta_t}{1-\bar{\alpha}_t}\phi_t
        \right)^2
    }\eps_2,
\label{eq13: predicted_result}
\end{equation}
where $\eps_2 \sim \mathcal{N}(\mathbf{0},\I)$. By substituting timestep $t+1$ into Eq.~\ref{eq13: predicted_result}, we can calculate the actual SNR of the predicted sample $\hx_t$, thereby completing the proof of Theorem 5.1. With the aid of the forward noising Eq.~\ref{eq2:forward_onestep}, a more concise expression form is obtained:
\begin{equation}
    \begin{aligned}
        &\hat{\x}_{t-1}=\hat{\gamma}_{t-1}\x_{t-1}+\psi_{t-1}\eps_3.
    \end{aligned}
    \label{eq14: re_predicted_result}
\end{equation}
where $\eps_3 \sim \mathcal{N}(\mathbf{0},\I)$, with more details in Appendix~\ref{app:c}.

\begin{table}[t]
\centering

\begin{tabular}{lcc}
\toprule
  & Type & SNR \\
\midrule
$\x_t$ & Forward & $\bar{\alpha}_t/(1-\bar{\alpha}_t)$ \\
$\hx_t$ & Reverse & $\hat{\gamma}^2_{t}{\bar{\alpha}_{t}}/\big(1 - \bar{\alpha}_{t} +
        (
            \frac{\sqrt{\bar{\alpha}_{t}}\beta_{t+1}}{1-\bar{\alpha}_{t+1}}\phi_{t+1})^2
        \big)$ \\ 
\bottomrule
\end{tabular}

\caption{The actual SNR of $\x_t$ and $\hx_t$.}
\label{tab1:lower_snr}
\end{table}

Tab.~\ref{tab1:lower_snr} and Eq.~\ref{eq14: re_predicted_result} clearly show that the actual SNR of the predicted samples $\hx_t$ in the reverse process is always lower than that of the perturbed sample $\x_t$ in the forward process, thus there is always a SNR-t bias where the SNR of predicted samples does not match the timestep $t$ during the inference phase, which provides solid theoretical evidence for the experimental conclusions in Sec.~\ref{sec:4}.

\subsection{Differential Correction in Pixel Space} 

\begin{figure}[!t]
\centering
   \includegraphics[width=0.95\linewidth]{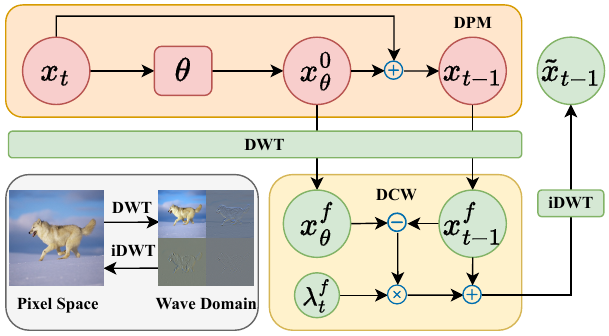} 
\caption{The overall framework of Differential Correction in Wavelet domain (DCW). At each denoising step, DPMs always generate the reconstructed sample $\x_\The^0$ for predicting $\x_0$ based on $\x_t$. After each denoising is completed, DCW maps $\x_\The^0$ and $\x_{t-1}$ to the wavelet domain via DWT to obtain $\x_\The^f$ and $\x_{t-1}^f$, where $f \in \{ll, lh, hl, hh\}$. Then, DCW corrects the different frequency components of $\x_{t-1}$ using Eq.~\ref{dcw}. Finally, DCW maps the corrected $\title{\x}_{t-1}^f$ back to the pixel space via iDWT.}
\label{fig2.method}
\end{figure}

% 在去噪的每一步，扩散模型总可以获得给定xt去预测x0的重建样本xy。因此，在去噪时间步t结束时，DCW利用离散小波变换将xt,x0和x_{t-1}分解映射到小波域进而得到z。随后，我们在小波域中对x_{t-1}的不同频率分量进行基于公式8的矫正。最后，我们将经过矫正的xt通过离散小波逆变换映射回像素空间。

In Sec.~\ref{sec:4} and Sec.~\ref{sec:4.1}, we clarify the SNR-t bias of DPMs and its specific manifestations from both empirical and theoretical perspectives. Meanwhile, we find that the actual SNR of $\hx_t$ in the inverse process is always lower than that of $\x_t$ at the same timestep $t$ in the forward process. Thus, we can infer that if we move the predicted sample toward the perturbed sample, the SNR-t bias can be alleviated. Interestingly, this gradient information pushing $\hx_t$ toward $\x_t$ is implicitly contained in each step of the denoising process. Now, we focus on the differential signal between the predicted sample $\hx_{t-1}$ and the reconstructed sample $\x_{\The}^0(\hx_t,t)$ in Eq.~\ref{eq13: predicted_result}. Based on Eq.~\ref{eq14: re_predicted_result} for $\hx_{t-1}$ and Eq.~\ref{eq10:assumption} for $\x_\The^0(\hx_t,t)$, the differential signal is expressed as:
\begin{equation}
    \hx_{t-1} - \x_{\The}^0(\hx_{t},t)=\hat{\gamma}_{t-1} (\x_{t-1} - \frac{\gamma_t}{\hat{\gamma}_{t-1}} \x_0) + \eta_t \eps_t
    \label{eq16:diff}
\end{equation}
where $\eta_t = \sqrt{\phi_t^2+\psi_{t-1}^2}$. Obviously, the differential signal based on Eq.~\ref{eq16:diff} contains directional information pointing to $\x_{t-1}$. Inspired by various directional information guidance strategies~\cite{vahdat2021score,zhang2023lookahead}, we integrate this gradient information into each step of denoising to guide the predicted samples $\hx_{t-1}$ to move toward the ideal perturbed samples $\x_{t-1}$:
\begin{equation}
    \hx_{t-1} = \hx_{t-1} + \lambda_t (\hx_{t-1} - \x_{\The}^0(\hx_t,t)),
    \label{eq17:dc_xt}
\end{equation}
where $\lambda_t$ is a scalar guidance factor that adjusts the magnitude of the effect of the differential signal. More specifically, the difference guidance shifts the predicted sample toward the noisy direction targeting $\x_{t-1}$. When the parameter is properly selected, it will improve the accuracy of the predicted sample to mitigate the SNR-t bias.

% Meanwhile, we also find that the differential gradient based on Eq.~\ref{eq:diff} can be reused across adjacent time steps. Leveraging the conditional distribution $q(\x_t \vert \x_{t-1}) = \N\left(\x_t| \sqrt{\alpha}\x_{t-1}, (1-\alpha_t) \I\right)$ of adjacent time steps in the forward process of DPMs, we can reasonably transform Eq.~\ref{eq:diff} as follows:
% \begin{equation}
%     \hx_t - \x_{\The}^0(\hx_t,t)=\hat{\gamma}_t \sqrt{\alpha_t} (\x_{t-1} - \frac{\gamma_t}{\hat{\gamma}_t} \x_0) + \hat{\eta}_t \eps_t,
%     \label{eq:diff}
% \end{equation}
% where $\hat{\eta}_t=\sqrt{\phi_t^2+\psi_t^2+\beta_t\hat{\gamma_t}^2}$. Given the multiplexing characteristic of differential signals, we switch to performing differential correction on $\hx_{t-1}$:
% \begin{equation}
%     \hx_{t-1} = \hx_{t-1} + \lambda_t (\hx_t - \x_{\The}^0(\hx_t,t)).
% \end{equation}
We emphasize that correcting $\hx_{t-1}$ is more advantageous than correcting $\hx_t$, as it not only enhances the quality of generation more effectively but also incurs less computational overhead.
Specifically, Eq.~\ref{eq13.re_actual_reverse} shows that the denoising result $\hx_{t-1}$ of the current step $t$ is jointly influenced by $\hx_t$ and $\x_\The^0(\hx_t,t)$ (or $\eps_\The(\hx_t,t)$). Meanwhile, the acquisition of $\x_\The^0(\hx_t,t)$ indicates the network has completed prediction. Thus, without increasing Neural Function Evaluations (NFE), Eq.~\ref{eq17:dc_xt} can correct $\hx_{t-1}$ and has no effect on the network output. Additionally, correcting the denoising result $\hx_{t-1}$ will bring gains to both the predicted sample and the network output in the next denoising process.

\subsection{Differential Correction in Wavelet Domain} 

In this subsection, we introduce the \textbf{D}ifferential \textbf{C}orrection method into the \textbf{W}avelet domain \textbf{(DCW)}, as shown in Fig.~\ref{fig2.method}, which stems from two key motivations: (1) During inference, DPMs first focus on reconstructing the low-frequency contours of images and then concentrate on the high-frequency details~\cite{yi2024towards}. Thus, our method should align with this important characteristic of DPMs; (2) The direction indicated by the differential signal based on Eq.~\ref{eq16:diff} is disturbed by Gaussian noise $\eta_t \eps_t$, thus performing correction in the time-frequency domain helps reduce noise interference.

Specifically, during the denoising process, DCW employs Discrete Wavelet Transform (DWT)~\cite{graps1995introduction} to decompose $\hx_t$ and $\x_{\The}^0(\hx_t,t)$ into four frequency subbands. For a given image sample $\x$ in the pixel space, after DWT is applied to $\x$, the following are obtained: $\x^{ll}$, $\x^{lh}$, $\x^{hl}$, and $\x^{hh}$, where the size of all four subbands is $ \mathbb{R}^{H/2\times W/2}$. $\x^{ll}$ represents the low-frequency subband, which characterizes the low-frequency information of the image, such as the shape of a human face or a house. $\x^{lh}, \x^{hl}$, and $\x^{hh}$ correspond to the high-frequency subbands in different directions, which characterize the high-frequency information of the image, such as the wrinkles of an elderly person or the veins of leaves. Subsequently, we separately perform differential correction on each type of frequency subband:
\begin{equation}
    \hx^f_{t-1} = \hx^f_{t-1} + \lambda^f_t (\hx^f_{t-1} - \x^f_{\The}(\hx_t,t)),
    \label{dcw}
\end{equation}
where $f\in\{ll,lh,hl,hh\}$, $\lambda^f_t$ is an adjustment coefficient related to both timesteps and frequency components. Then, we utilize the inverse discrete wavelet transform (iDWT)~\cite{graps1995introduction} to map the samples back to the pixel space, thereby forming a complete DCW operation:
\begin{equation}
    \tilde{\x}_{t-1} = \textrm{iDWT}(\hat{\x}_{t-1}^f|f\in\{ll,lh,hl,hh\})
\end{equation}

Next, we discuss the adjustment strategy for $\lambda_t^f$. For the low-frequency component, we propose a time-dependent weighting strategy that follows a decaying schedule as the denoising advances. Conversely, a decreasing strategy is adopted for the high-frequency components. Specifically, in early denoising steps, we assign a relatively large coefficient to the low-frequency correction term to prioritize the generation of low-frequency components, which also effectively mitigates the interference of high-frequency noise errors during the initial denoising phase. In the later denoising stages, we assign a larger coefficient to the high-frequency correction to focus on the restoration of high-frequency details, which helps suppress the over-expression of low-frequency components towards the end of the process.

Notably, the reverse process variance $\sigma_t$ in DPMs serves as a robust indicator of the denoising progress and has been widely adopted for dynamic modulation in various sampling techniques~\cite{dhariwal2021diffusion,um2024self,yu2025frequency}. Consequently, we leverage this reverse variance to implement our dynamic correction. The low-frequency component coefficient is formulated as:
\begin{equation}
    \lambda_t^l = \lambda_l \cdot \sigma_t,
\label{eq17:low}
\end{equation}
where $\lambda_l$ denotes a scalar coefficient. Similarly, the high-frequency component coefficient is defined as:
\begin{equation}
    \lambda_t^h = (1-\lambda_h) \sigma_t,
\label{eq18:high}
\end{equation}
where $\lambda_h$ also denotes a scalar coefficient. Furthermore, inspired by SG-Minority~\cite{um2024self}, more weight design strategies are provided in Appendix~\ref{app:d}.

\section{Experiments}
\label{sec:5}
In this section, we conduct extensive experiments on numerous datasets and DPMs to show the effectiveness, generality, superiority, and robustness of our method.

We evaluate it on multiple representative DPM frameworks and samplers, including IDDPM~\cite{nichol2021improved}, ADM~\cite{dhariwal2021diffusion}, DDIM~\cite{songdenoising}, A-DPM~\cite{baoanalytic}, EA-DPM~\cite{bao2022estimating}, EDM~\cite{Karras2022edm}, DiT~\cite{peebles2023scalable}, PFGM++~\cite{xu2023pfgm++}, FLUX~\cite{blackforestlabs2024flux}, and Qwen-Image~\cite{wu2025qwenimagetechnicalreport}. Then, we choose DPM-AE~\cite{wangimproved} (ICLR 2025) and DPM-AT~\cite{zhang2025antiexposure} (ICLR 2025) as comparative models. Furthermore, we also integrate our method into the open-source bias-corrected models ADM-IP~\cite{ning2023input} (ICML 2023), ADM-ES~\cite{ningelucidating} (ICLR 2024), and DPM-FR~\cite{yu2025frequency} (ACM MM 2025) to further demonstrate the superiority of our approach. Meanwhile, experiments are conducted across datasets of varying resolutions, including CIFAR-10~\cite{krizhevsky2009learning}, CelebA 64$\times$64~\cite{liu2015deep}, ImageNet 128$\times$128~\cite{chrabaszcz2017downsampled}, and LSUN Bedroom 256$\times$256~\cite{yu2015lsun}. 

Overall, we categorize our evaluations into two main types: stochastic generation~\cite{ho2020denoising} and deterministic generation~\cite{songdenoising}. To comprehensively assess generation quality, we employ standard metrics including Fréchet Inception Distance (FID)~\cite{heusel2017gans} and Recall~\cite{heusel2017gans}, where FID serves as the primary metric. All quantitative results are computed over 50K generated samples with the full training set as the reference distribution. For qualitative evaluation, we visualize text-to-image results to intuitively demonstrate the effectiveness of our method.  
% To validate the advancement of our approach, we compare against several state-of-the-art bias-correction methods, including ADM-IP \cite{ning2023input}, ADM-ES \cite{ningelucidating}, DPM-FR \cite{yu2025frequency}, DPM-AE (ICLR2025) andDPM-AE (ICLR2025).

\subsection{Results on Classic Diffusion Models}
To verify the effectiveness and generality of the proposed method, we select several classic diffusion models, namely IDDPM, ADM, and ADM-IP. Additionally, we choose datasets with different resolutions, including CIFAR-10~\cite{krizhevsky2009learning} $32\times32$, CelebA $64\times64$~\cite{liu2015deep}, ImageNet $128\times128$~\cite{chrabaszcz2017downsampled}, and LSUN Bedroom $256\times256$~\cite{yu2015lsun}. Meanwhile, we select FID and Recall as evaluation metrics to assess fidelity and diversity, and use 20 and 50 as sampling steps.

Tab.~\ref{tab2:datasets} clearly shows that our method comprehensively improves the generation quality of the baseline models across all models and datasets. For example, on the CIFAR-10 dataset, DCW helps IDDPM reduce the FID score by $42.6\%$ and $25\%$ in the 20-step and 50-step tasks, respectively.

For a fair comparison with recent methods on exposure bias, we follow previous works and use the same baselines, namely DDIM~\cite{songdenoising} sampler applied to A-DPM and ADM. Tab.~\ref{tab:ICLR25} clearly shows our method consistently outperforms DPM-AE~\cite{wangimproved} (ICLR 2025) and DPM-AT~\cite{zhang2025antiexposure} (ICLR 2025) across all generation results, further validating its superiority.

% \begin{table}[t]
% \centering
% \caption{FID $\downarrow$ on PFGM using different deterministic sampling methods.}
% \begin{tabularx}{0.47\textwidth}{@{}l *{6}{c}@{}}  % 移除type列
% \toprule
% & \multicolumn{3}{c}{T=20} & \multicolumn{3}{c}{T=50} \\
% \cmidrule(lr){2-4} \cmidrule(lr){5-7}  % 调整分隔线范围
% model & FID$\downarrow$ & IS$\uparrow$ & Prec$\uparrow$ & FID$\downarrow$ & IS$\uparrow$ & Prec$\uparrow$ \\
% \midrule
% IDDPM      & 13.19 & 0.49 & 5.55 & 0.56 & 6.53 & 3.88 \\
% \textbf{+Ours}  & \textbf{7.32} & \textbf{3.94} & \textbf{2.69} & \textbf{7.32} & \textbf{3.94} & \textbf{2.69} \\
% ADM & 8.79 & 4.54 & 2.91 & 8.79 & 4.54 & 2.91 \\
% \textbf{+Ours} & \textbf{8.00} & \textbf{4.41} & \textbf{2.84} & \textbf{6.18} & \textbf{4.41} & \textbf{2.84} \\
% ADM-IP & 6.62 & 3.67 & 2.53 & 6.62 & 3.67 & 2.53 \\
% \textbf{+Ours} & \textbf{6.18} & \textbf{3.46} & \textbf{2.48} & \textbf{6.18} & \textbf{3.46} & \textbf{2.48} \\
% \bottomrule
% \end{tabularx}
% \label{tab1:pfgm}
% \end{table}

\begin{table}[t]
\centering

\begin{tabularx}{0.49\textwidth}{@{}l l *{4}{c}@{}}  % 添加一个 'l' 列（dataset）
\toprule
& & \multicolumn{2}{c}{$T=20$} & \multicolumn{2}{c}{$T=50$} \\
\cmidrule(lr){3-4} \cmidrule(lr){5-6}
Model & Dataset & FID$\downarrow$ & Rec$\uparrow$ & FID$\downarrow$ & Rec$\uparrow$ \\  % 插入 dataset 列
\midrule
IDDPM      & CIFAR-10 32 & 13.19 & 0.50 & 5.55 & 0.56 \\
\textbf{+Ours}  & CIFAR-10 32 & \textbf{7.57} & \textbf{0.56} & \textbf{4.16} & \textbf{0.58} \\
\midrule
ADM-IP & CelebA 64  & 11.95 & 0.42 & 4.52 & 0.55 \\
\textbf{+Ours} & CelebA 64 & \textbf{10.41} & \textbf{0.47} & \textbf{4.34} & \textbf{0.57} \\
\midrule
ADM & ImageNet 128& 12.28 & 0.52 & 5.18 & 0.58 \\
\textbf{+Ours} & ImageNet 128 & \textbf{10.34} & \textbf{0.54} & \textbf{4.52} & \textbf{0.58} \\
\midrule
IDDPM & LSUN 256& 18.69 & 0.27 & 8.42 & 0.41 \\
\textbf{+Ours} & LSUN 256 & \textbf{11.03} & \textbf{0.36} & \textbf{5.24} & \textbf{0.45} \\
\bottomrule
\end{tabularx}

\caption{FID and Recall (Rec)  on datasets with different resolutions.}
\label{tab2:datasets}
\end{table}

\begin{table}[t]
\centering

\begin{tabularx}{0.49\textwidth}{@{}l *{6}{c}@{}}  % 移除type列
\toprule
& \multicolumn{3}{c}{DDIM} & \multicolumn{3}{c}{ADM} \\
\cmidrule(lr){2-4} \cmidrule(lr){5-7}  % 调整分隔线范围
Model & 10 & 20 & 50 & 10 & 20 & 50 \\
\midrule
Base      & 14.40 & 6.87 & 4.15 & 22.62 & 10.52 & 4.55 \\
Base-AE & 13.98 & 6.76 & 4.10 & - & - & - \\
Base-AT & - & - & - & 15.88 & 6.60 & 3.34 \\
\midrule
\textbf{Base+Ours} & \textbf{9.36} & \textbf{4.64} & \textbf{3.33} & \textbf{13.01} & \textbf{5.59} & \textbf{2.95} \\
\bottomrule
\end{tabularx}

\caption{FID $\downarrow$ on CIFAR-10 using ADM and DDIM.}
\label{tab:ICLR25}
\end{table}

\subsection{Results on Bias-Corrected Diffusion Models}
\label{sec6.2:biad_corr}
To verify the generality and advancement of our method, we select several improved models for exposure bias as comparative models and integrate DCW into them, namely ADM-ES~\cite{ningelucidating} and DPM-FR~\cite{yu2025frequency}. Notably, DPM-FR is the SOTA model for exposure bias. To be consistent with them, we divide the generation task into two categories: stochastic sampling and deterministic sampling. In stochastic sampling, we select A-DPM~\cite{baoanalytic} and NPR-DM in EA-DPM~\cite{bao2022estimating} as the baseline models. In deterministic sampling, we use EDM~\cite{Karras2022edm} and PFGM++~\cite{xu2023pfgm++} as baseline models and measure the sampling cost by Neural Function Evaluations (NFE)~\cite{vahdat2021score}.

% vahdat2021score

% , and the former elaborates on the design space of DPMs in training and inference, and the latter is an improved Poisson flow-based DPM.

% , and the former derives the analytical form of the reverse variance of DPMs, while the latter estimates more accurate reverse variance using neural networks

Tab.~\ref{tab3:a_dpm} shows that in stochastic sampling, DCW comprehensively improves the generation quality of baseline models. For different models, noise scheduling strategies, and time-step settings, DCW consistently achieves a significant reduction in the FID scores. For the corrected models, even though they have already achieved extremely low FID scores, DCW can still further reduce the FID results, which demonstrates the advancement of our method.

\begin{table}[t] 
\centering
\setlength{\tabcolsep}{4.3pt}

% \resizebox{0.46\textwidth}{!}{
% \begin{tabular}{@{\extracolsep{\fill}}lcccccc@{}}
\begin{tabular}{lcccccc}
\toprule
& \multicolumn{3}{c}{CIFAR-10 (LS)} & \multicolumn{3}{c}{CIFAR-10 (CS)} \\
\cmidrule(lr){2-4} \cmidrule(lr){5-7}
Model & 10 & 25 & 50 & 10 & 25 & 50 \\
% \midrule
% DDPM & 44.45 & 21.83 & 15.21 & 34.76 & 16.18 & 11.11 \\
% DDPM, $\beta_n$ & 233.41 & 125.05 & 66.28 & 205.31 & 84.71 & 37.35 \\
\midrule
A-DPM & 34.26 & 11.60 & 7.25 & 22.94 & 8.50 & 5.50 \\
\textbf{+Ours} & \textbf{17.56} & \textbf{8.81} & \textbf{5.38} & \textbf{12.44} & \textbf{5.99} & \textbf{4.06} \\
\midrule
A-DPM-FR & 12.38 & 6.63 & 4.52 & 11.61 & 4.40 & 3.62 \\
\textbf{+Ours} & \textbf{10.91} & \textbf{6.03} & \textbf{4.44} & \textbf{9.80} & \textbf{4.33} & \textbf{3.56} \\
\midrule
NPR-DM & 32.35 & 10.55 & 6.18 & 19.94 & 7.99 & 5.31 \\
% +LA & 25.59 & 8.84 & 5.28 & - & - & - \\
\textbf{+Ours} & \textbf{16.60} & \textbf{8.64} & \textbf{5.40} & \textbf{11.44} & \textbf{6.38} & \textbf{4.80} \\
\midrule
NPR-DM-FR & 10.86 & 5.76 & 4.19 & 10.18 & 4.07 & 3.44 \\
% +LA & 25.59 & 8.84 & 5.28 & - & - & - \\
\textbf{+Ours} & \textbf{9.81} & \textbf{5.30} & \textbf{4.11} & \textbf{8.46} & \textbf{3.96} & \textbf{3.33} \\
% \midrule
% SN-DPM & 24.06 & 6.91 & 4.63 & 16.33 & 6.05 & 4.17 \\
% LA-SN-DPM & 19.75 & 5.92 & 4.31 & - & - & - \\
% \textbf{SN-DPM-W++} & \textbf{11.73} & \textbf{4.73} & \textbf{3.78} & \textbf{12.53} & \textbf{4.51} & \textbf{3.47} \\
\bottomrule
\end{tabular}
% }

\caption{FID $\downarrow$ on CIFAR-10 using A-DPM and EA-DPM.}
\label{tab3:a_dpm}
\end{table}

\begin{table}[t]
\centering
\setlength{\tabcolsep}{6.8pt}

\begin{tabularx}{0.47\textwidth}{@{}l *{6}{c}@{}}  % 移除type列
\toprule
& \multicolumn{3}{c}{EDM} & \multicolumn{3}{c}{PFGM++} \\
\cmidrule(lr){2-4} \cmidrule(lr){5-7}  % 调整分隔线范围
Model & 13 & 21 & 35 & 13 & 21 & 35 \\
\midrule
Base      & 10.66 & 5.91 & 3.74 & 12.92 & 6.53 & 3.88 \\
\textbf{+Ours}  & \textbf{5.67} & \textbf{3.37} & \textbf{2.41} & \textbf{6.98} & \textbf{3.83} & \textbf{2.64} \\
\midrule
Base-ES & 6.59 & 3.74 & 2.59 & 8.79 & 4.54 & 2.91 \\
\textbf{+Ours} & \textbf{6.13} & \textbf{3.57} & \textbf{2.50} & \textbf{8.00} & \textbf{4.41} & \textbf{2.84} \\
\midrule
Base-FR & 4.68 & 2.84 & 2.13 & 6.62 & 3.67 & 2.53 \\
\textbf{+Ours} & \textbf{4.57} & \textbf{2.79} & \textbf{2.12} & \textbf{6.18} & \textbf{3.46} & \textbf{2.48} \\
\bottomrule
\end{tabularx}

\caption{FID $\downarrow$ on CIFAR-10 using different fast samplers.}
\label{tab4:edm_pfgm}
\end{table}

Tab.~\ref{tab4:edm_pfgm} shows that in deterministic sampling, DCW can not only improve the generation quality of baseline models but also further reduce the FID of corrected models. For EDM, DCW reduces the FID by $47.1\%$, $47.4\%$, and $36.4\%$ in the 13, 21, and 35 NFE generation tasks, respectively. Although ADM-ES and ADM-FR have already improved generation performance by alleviating exposure bias, DCW can still further improve the corrected models. For EDM-ES the reductions of FID under the three NFE tasks are $7.0\%$, $5.3\%$, and $3.5\%$, respectively. For PFGM-FR, the corresponding reductions are $6.6\%$, $5.7\%$, and $2.0\%$, respectively. Meanwhile, we also provide the DiT~\cite{peebles2023scalable} experiments on the ImageNet $256 \times 256$ dataset in Appendix~\ref{app:e}.

\subsection{Qualitative Comparison}
\begin{figure}[t] 
    \centering
    \begin{subfigure}[b]{0.15\textwidth} 
        \includegraphics[width=\linewidth]{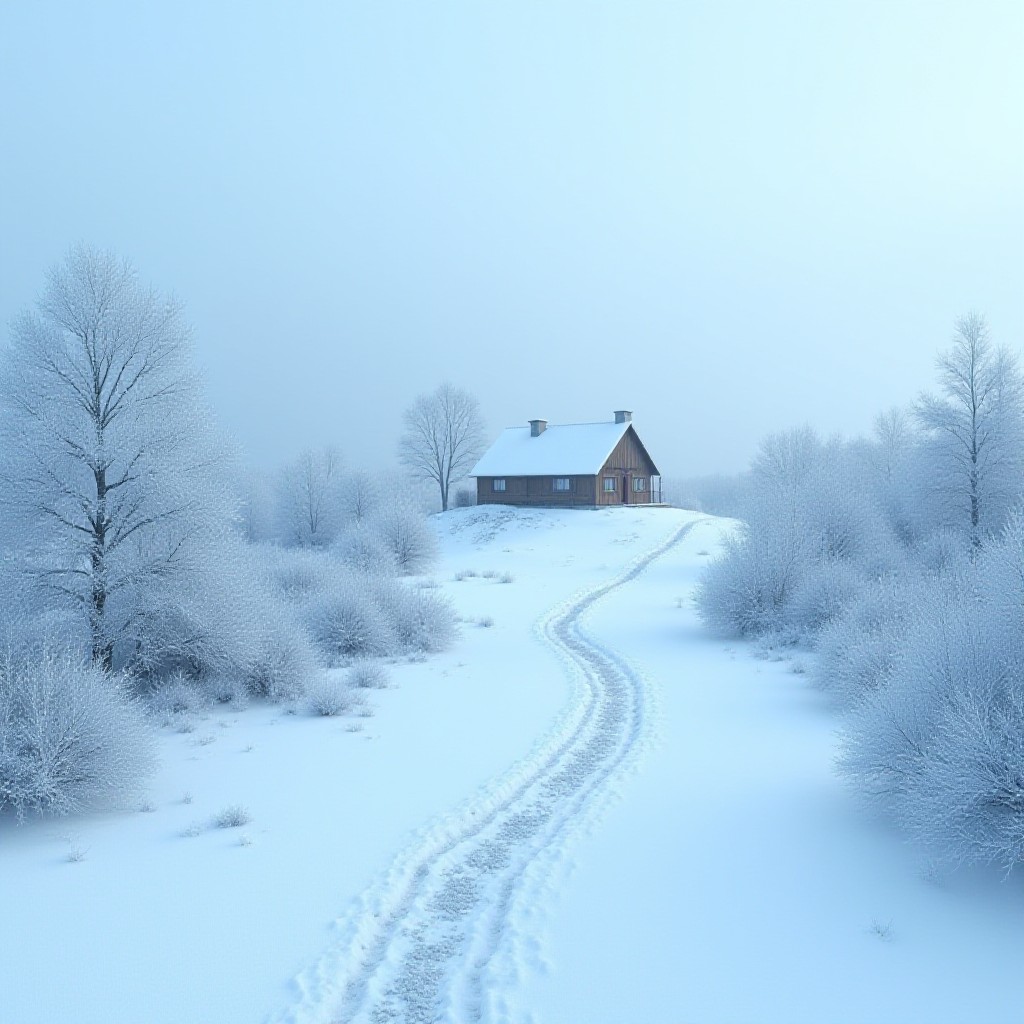}
    \end{subfigure}
    \begin{subfigure}[b]{0.15\textwidth}
        \includegraphics[width=\linewidth]{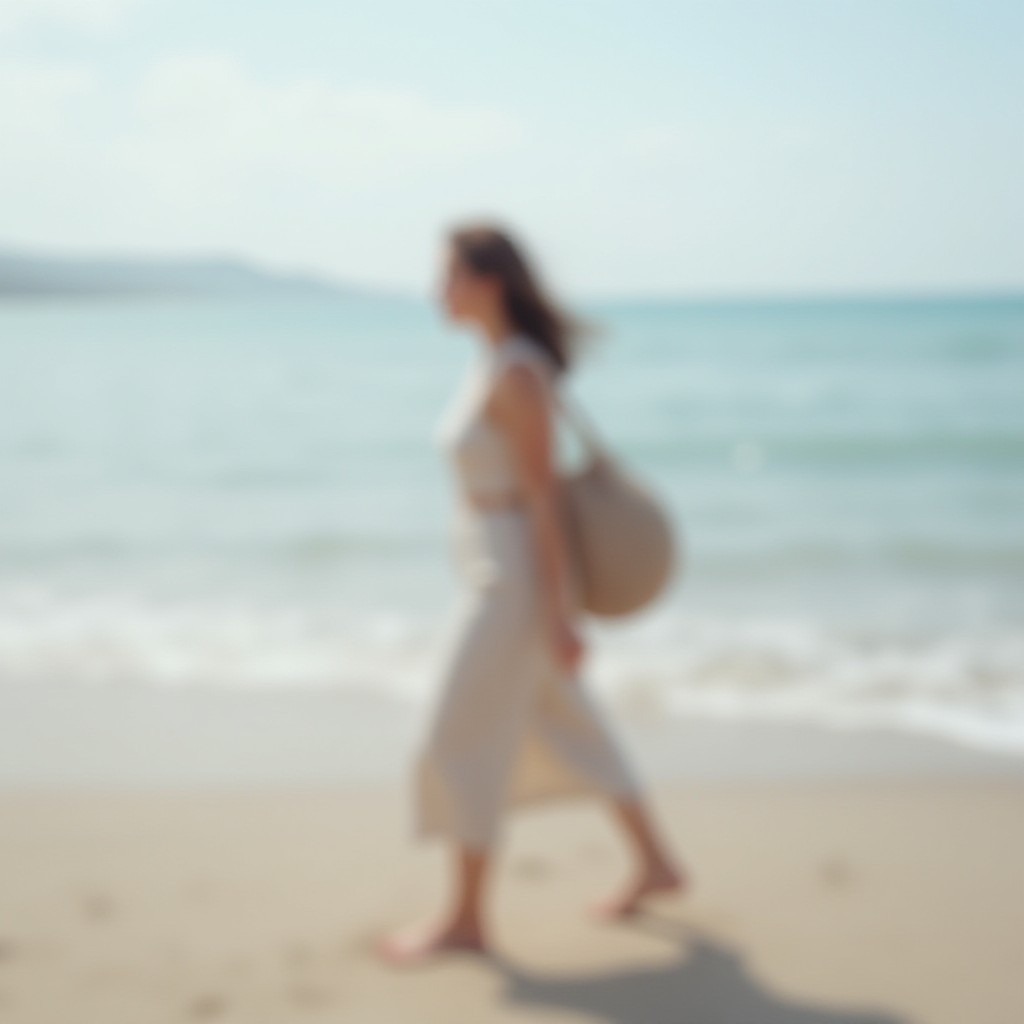}
    \end{subfigure}
    \begin{subfigure}[b]{0.15\textwidth}
        \includegraphics[width=\linewidth]{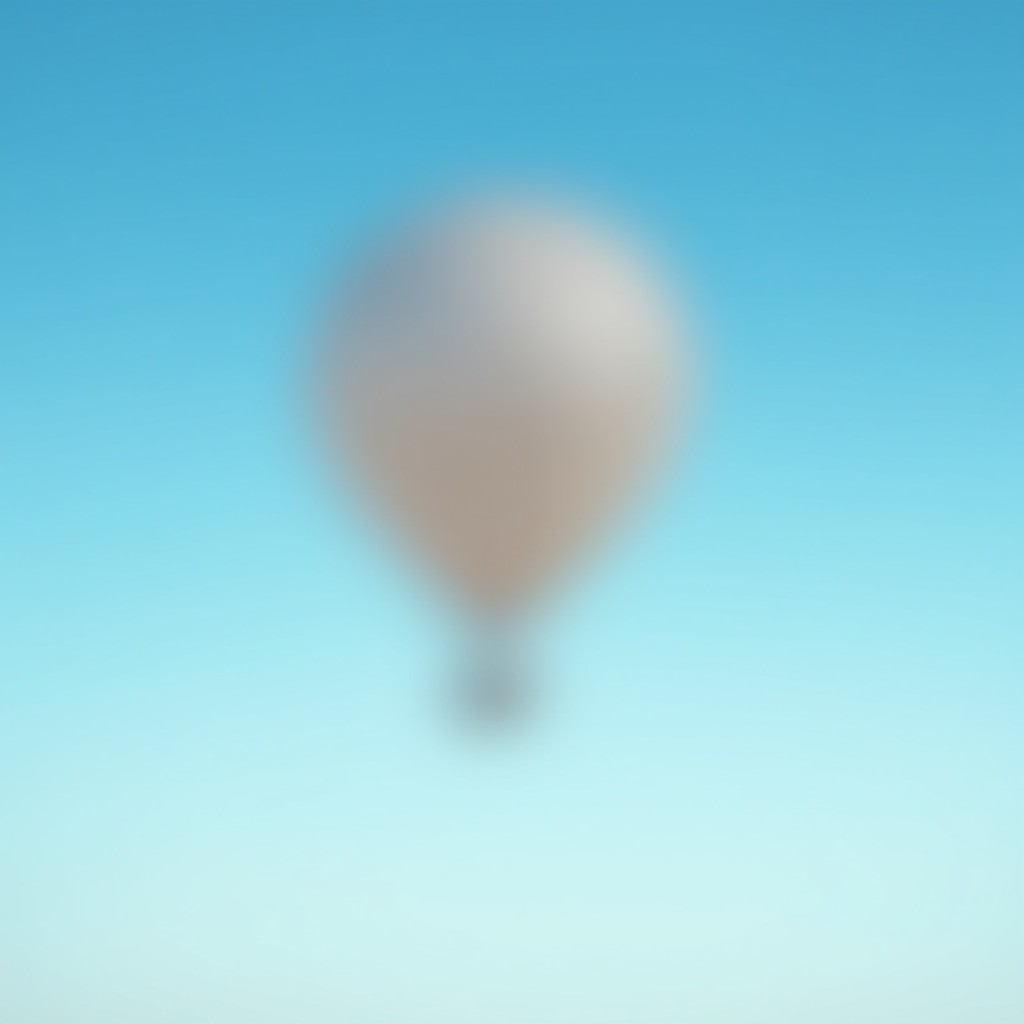}
    \end{subfigure}
    % \begin{subfigure}[b]{0.15\textwidth}
    %     \includegraphics[width=\linewidth]{figures/show/b1.jpg}
    % \end{subfigure}
    \begin{subfigure}[b]{0.15\textwidth}
        \includegraphics[width=\linewidth]{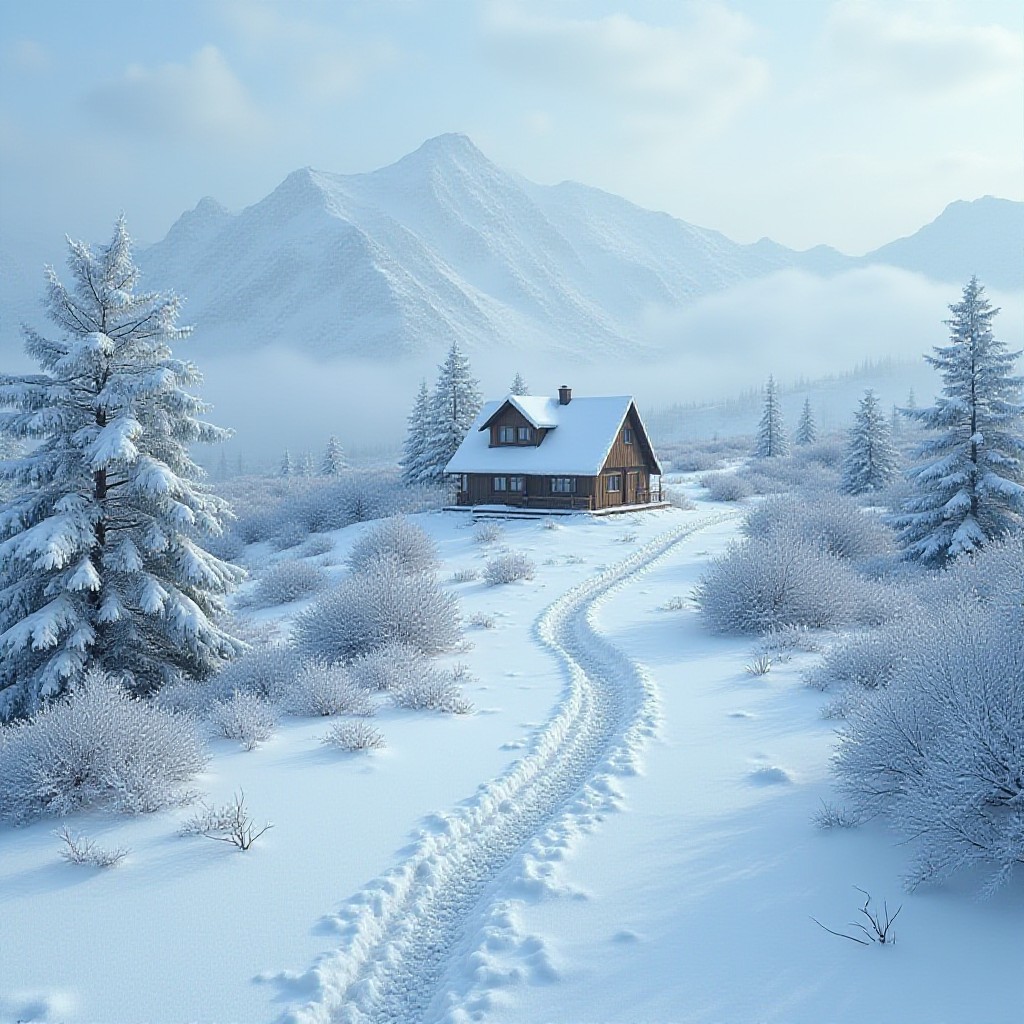}
    \end{subfigure}
    \begin{subfigure}[b]{0.15\textwidth}
        \includegraphics[width=\linewidth]{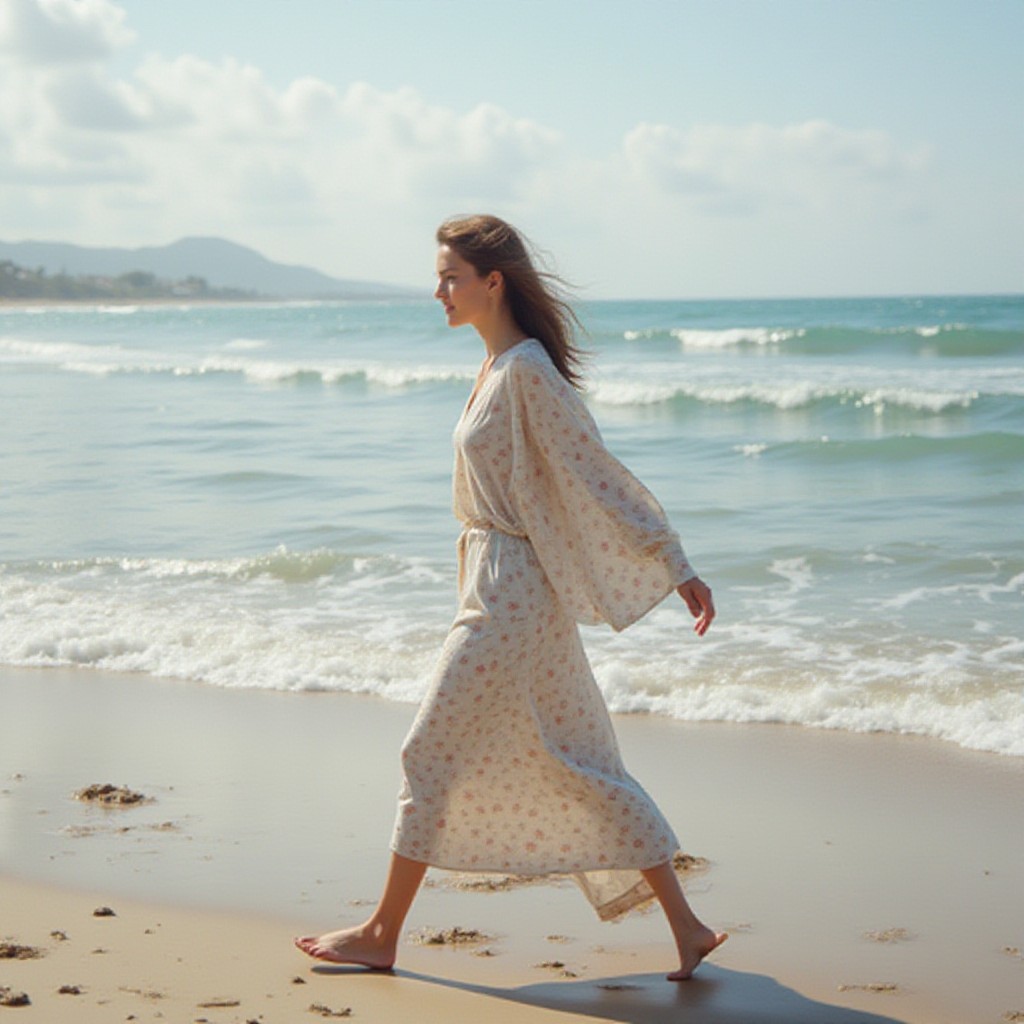}
    \end{subfigure}
    \begin{subfigure}[b]{0.15\textwidth}
        \includegraphics[width=\linewidth]{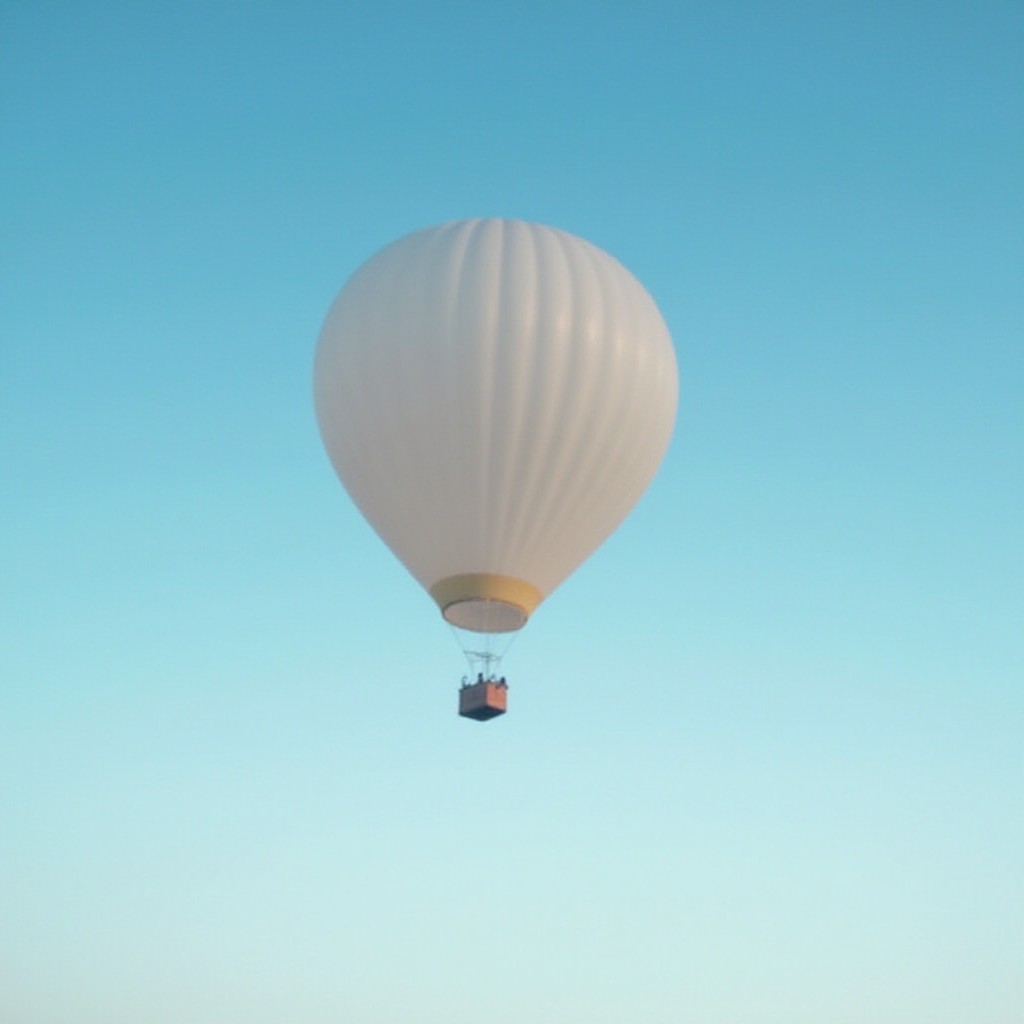}
    \end{subfigure}
    % \begin{subfigure}[b]{0.15\textwidth}
    %     \includegraphics[width=\linewidth]{figures/show/d1.jpg}
    % \end{subfigure}
    \caption{Qualitative comparison between FLUX (first row) and FLUX-DCW (second row) using 10 steps.}
    \label{fig3:sdxl_show}
\end{figure}

To intuitively demonstrate the impact of DCW on the generation quality, we set the same random seed and sampling steps for the baseline models and improved models during inference,  ensuring that they follow similar denoising trajectories. Specifically, we adopt FLUX~\cite{blackforestlabs2024flux} as the baseline model and use 10 sampling steps. As shown in Fig.~\ref{fig3:sdxl_show}, images generated by FLUX suffer from distortion issues such as over-smoothing and overexposure. In contrast, DCW significantly mitigates these problems, substantially enhancing the aesthetic quality and visual appeal of the generated images. More qualitative results are provided in Appendix~\ref{app:f}, including the qualitative experiments on Qwen-Image~\cite{wu2025qwenimagetechnicalreport}.

% wu2025qwenimagetechnicalreport
\subsection{Ablation Study}
% 1.diff-Fre, 2.low jian, high zeng, 3.参数敏感度， 4.时间计算，5.小波基。

In this subsection, we conduct detailed ablation experiments to examine the role of each component in DCW. We primarily use CIFAR-10 as the test dataset.

\textbf{Effect of the Wavelet Domain.} First, we investigate the impact of each component in DCW on generation quality via four comparative variants. Differential correction applied solely in the pixel space is denoted as ``DC". Then, we denote differential correction applied only to high frequency or low frequency wavelet components as ``DH" and ``DL", respectively. Finally, our complete framework involves applying differential correction to both the high frequency and low frequency components, denoted as ``DCW". Tab.~\ref{tab6:ablation_study} clearly shows that the differential correction method is effective in both the pixel and wavelet space, resulting in noticeable improvements in the generation quality. Furthermore, the simultaneous integration of differential correction into both high-frequency and low-frequency components enhances performance even further, underscoring the necessity and advantages of applying the method within the wavelet domain.

\begin{table}[t]
\centering

\begin{tabularx}{0.47\textwidth}{@{}l c *{3}{c}@{}}  % 新增c列用于type
\toprule  
Model & Type & 10 & 25 & 50  \\  % 新增type表头
\midrule
A-DPM   & Baseline & 22.94 & 8.50 & 5.50  \\  % 填充low
\midrule
A-DPM-DC & Pixel Space & 15.71 & 6.38 & 4.31  \\  % 填充low
A-DPM-DH & High Frequency & 16.72 & 6.05 & 4.06  \\  % 填充low
A-DPM-DL & Low Frequency & 13.21 & 7.00 & 5.10  \\  % 填充low
\midrule
\textbf{A-DPM-DCW} & \textbf{High $\&$ Low} & \textbf{12.46} & \textbf{5.99}  &\textbf{4.06} \\  % 填充low
\bottomrule
\end{tabularx}

\caption{Ablation study (FID $\downarrow$) of different frequency components.}
\label{tab6:ablation_study}
\end{table}

% \begin{table}[t]
% \centering
% \caption{Ablation study of DCW.}
% \begin{tabular*}{0.46\textwidth}{@{\extracolsep{\fill}}lcccc@{}}  
% \toprule
% Models & FID$\downarrow$ & IS$\uparrow$ & Precision$\uparrow$ & Recall$\uparrow$ \\  
% \midrule
% IDDPM & 13.03 & 8.65 & 0.63 & 0.49 \\ 
% IDDPM-DC-T & 7.87 & 8.88 & 0.66 & 0.56 \\ 
% IDDPM-DC-H & 7.81 & 9.11 & 0.67 & 0.57\\ 
% IDDPM-DC-L & 7.58 & 9.67 & 0.66 & 0.57 \\
% \textbf{IDDPM-DCW} & \textbf{6.62} & \textbf{9.74} & \textbf{0.67} & \textbf{0.60} \\ 
% \bottomrule
% \end{tabular*}
% \label{tab5:ablation_study}
% \end{table}

% \begin{figure}[t]
% \centering
%     \includegraphics[width=\linewidth]{figures/par.jpg}
%     \caption{Search experiments of $\lambda$\,.}
%     \label{fig3a:s}
% \caption{Hyperparameters search experiments on CIFAR-10 (LS) and CIFAR-10 (CS) using A-DPM with 25 as the sampling number.}
% \label{fig8:parameter}
% \end{figure}

\begin{figure}[t]
\centering
\begin{subfigure}[b]{0.495\linewidth}
    \centering
    \includegraphics[width=\linewidth]{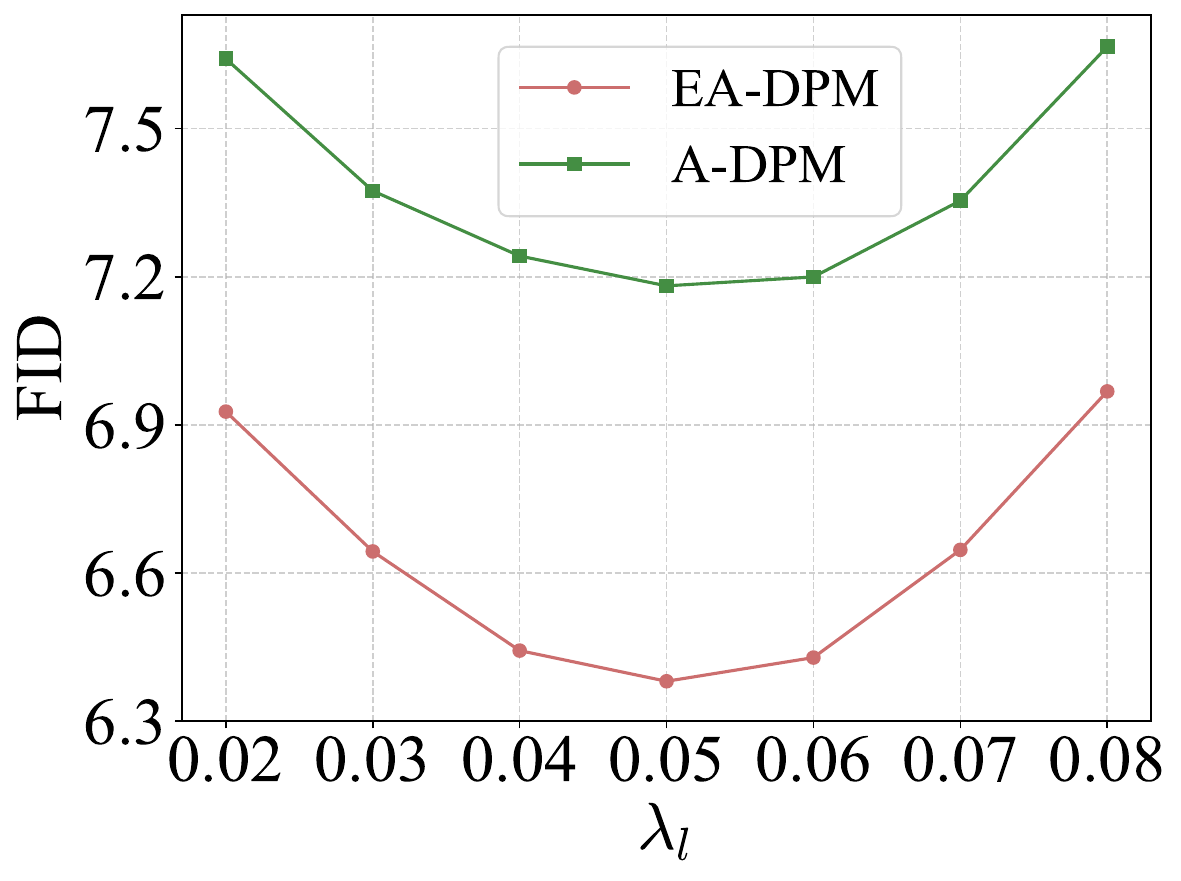}
    \caption{Search experiments of $\lambda_l$\,.}
    \label{fig3a:l}
\end{subfigure}%
\begin{subfigure}[b]{0.495\linewidth}
    \centering
    \includegraphics[width=\linewidth]{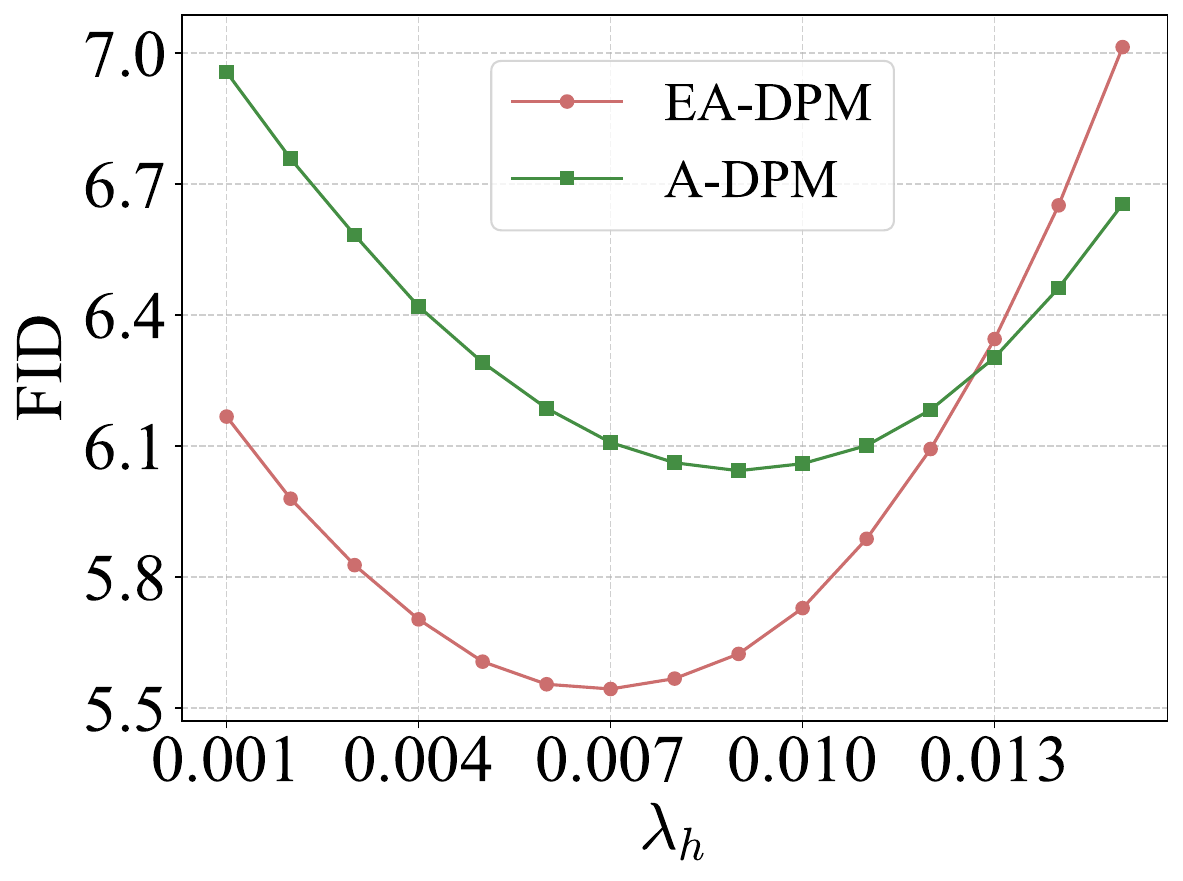}
    \caption{Search experiments of $\lambda_h$\,.}
    \label{fig3b:h}
\end{subfigure}%
\caption{Hyperparameter search experiments on CIFAR-10 (CS) using A-DPM and EA-DPM with $T=25$.}
\label{fig4:parameter}
\end{figure}

\textbf{Sensitivity of Hyperparameter $\lambda_f$.} Next, we examine the sensitivity of DCW to hyperparameters. DCW is robust to variations in hyperparameters: for both low-frequency and high-frequency adjustment factors, the intensity of differential correction gradually increases with the growth of the adjustment parameters, and the FID of the final generated results exhibits a trend of first decreasing and then increasing, as shown in Fig.~\ref{fig4:parameter}. Therefore, we can quickly determine the optimal values of the hyperparameters through a simple two-stage search method, as presented in Appendix~\ref{app:g}.

% \begin{table}[t]
% \centering
% \caption{Time for a batch generation on various datasets with one Nvidia H20 GPU.}
% \small
% \setlength{\tabcolsep}{3pt} 
% \begin{tabular}{@{}l l c c c@{}} % 新增Dataset列的列格式
% \toprule
% % 第一个数据集（对应两行）
% CelebA $64 \times 64$ & Models & ADM & ADM-DCW & Overhead \\
% & Time & 1.4136 & 1.4139 & 0.02\% \\
% \midrule
% % 第二个数据集（对应两行）
% ImageNet $128 \times 128$ & Models & ADM & ADM-DCW & Overhead \\
% & Time & 1.4136 & 1.4139 & 0.02\% \\
% \midrule
% % 第二个数据集（对应两行）
% Bedroom $256 \times 256$ & Models & ADM & ADM-DCW & Overhead \\
% & Time & 1.4136 & 1.4139 & 0.02\% \\
% \bottomrule
% \end{tabular}
% \label{tab8:time}
% \end{table}

\begin{table}[t]
\setlength{\tabcolsep}{4.5pt}
\centering

\begin{tabularx}{0.47\textwidth}{@{}l l *{3}{c}@{}}  % 添加一个 'l' 列（dataset）
\toprule
Model & Dataset & Time & DCW Time & Overhead \\  % 插入 dataset 列
\midrule
ADM-IP & CelebA 64  & 4.25 & 4.27 & $0.47\%$  \\
ADM & ImageNet 128& 12.59 & 12.60 & $0.08\%$  \\
IDDPM & LSUN 256& 15.57 & 15.61 & $0.26\%$  \\

\bottomrule
\end{tabularx}

\caption{Batch generation time on a single NVIDIA A6000 GPU.}
\label{tab6:overhead}
\end{table}

\textbf{Impact of DCW on computational overhead.} Finally, we evaluate the impact of DCW on computational overhead. Without loss of generality, we fix the random seed, the number of timesteps, and batch size, then conduct extensive experiments on datasets of varying resolutions: CelebA $64\times64$, ImageNet $128\times128$, and LSUN Bedroom $256\times256$. To address statistical bias, each experiment is repeated 100 times, and the average runtime is reported. Tab.~\ref{tab6:overhead} demonstrates that the computational cost incurred by DCW for DPMs is negligible, introducing virtually no generation latency. Specifically, DCW adds an additional time overhead of approximately 0.47\%, 0.08\%, and 0.26\% for the three generation tasks, which is clearly minimal. These experimental results regarding time overhead further reinforce the practicality and superiority of DCW.

\section{Conclusion}
% In conclusion, we identify and address the SNR-t bias in DPMs, highlighting a critical issue in which the misalignment between SNR and timestep during inference significantly degrades generation quality. Through rigorous theoretical analysis and empirical validation, we propose a differential correction method that effectively mitigates this bias by leveraging frequency-component decomposition, aligning with the inherent low-to-high frequency denoising order of diffusion models. Extensive experiments conducted across a diverse array of DPMs—including classic variants such as IDDPM, ADM, DDIM, EDM, DiT, and PFGM++, as well as the more recent SDXL—demonstrate that our approach substantially enhances generation quality with negligible computational overhead. This work not only offers a novel perspective on the robustness of DPMs during inference but also presents a lightweight yet powerful technique to improve their performance, which can be readily integrated into various diffusion frameworks. The implications of this method are promising for advancing generative tasks in computer vision.

In conclusion, we find that DPMs often suffer from a signal-to-noise ratio–timestep (SNR-t) bias. This bias refers to the mismatch between the SNR of a denoising sample and its associated timestep during inference. During training, the SNR of a sample is a deterministic function of its timestep, but this coupling is broken at inference due to accumulated prediction and discretization errors, which leads to error accumulation and degraded generation quality.
We provide empirical evidence and theoretical analysis for this phenomenon and propose a simple differential correction method to mitigate the SNR-t bias. Since diffusion models tend to reconstruct low-frequency components before refining high-frequency details in the reverse process, we decompose samples into multiple frequency components and apply differential correction to each component separately. Extensive experiments show that our approach improves the generation quality of various diffusion models on datasets with different resolutions, while incurring negligible computational overhead.

%In conclusion, we identify and address the SNR-t bias in DPMs, a critical issue that the misalignment between SNR and timestep during inference severely degrades generation quality. Through rigorous theoretical analysis and empirical validation, we propose a differential correction method that, by leveraging frequency-component decomposition (aligning with diffusion models’ inherent low-to-high frequency denoising order), effectively mitigates this bias. Extensive experiments across a diverse range of DPMs, from classic variants (IDDPM, ADM, DDIM, EDM, DiT, PFGM++) to SDXL, shows that our approach substantially enhances generation quality with negligible computational overhead. This work not only provides a novel perspective on the robustness of DPMs during inference but also offers a lightweight yet powerful technique to boost their performance, which can be readily integrated into various diffusion frameworks and has promising implications for advancing generative tasks in computer vision.

{
    \small
    \bibliographystyle{ieeenat_fullname}
    \bibliography{main}

@String(CVPR= {IEEE Conf. Comput. Vis. Pattern Recog.})

@String(ICCV= {Int. Conf. Comput. Vis.})

@String(ECCV= {Eur. Conf. Comput. Vis.})

@String(ICLR = {Int. Conf. Learn. Represent.})

@String(AAAI = {AAAI})

@String(CVPR  = {CVPR})

@String(ICCV  = {ICCV})

@String(ECCV  = {ECCV})

@String(ICLR  = {ICLR})

@InProceedings{dhariwal2021diffusion,
  title={Diffusion models beat {GANs} on image synthesis},
  author={Dhariwal, Prafulla and Nichol, Alexander},
  booktitle={NeurIPS},
  year={2021}
}

@inproceedings{rombach2022high,
  title={High-resolution image synthesis with latent diffusion models},
  author={Rombach, Robin and Blattmann, Andreas and Lorenz, Dominik and Esser, Patrick and Ommer, Bj{\"o}rn},
  booktitle={CVPR},
  year={2022}
}

@InProceedings{kongdiffwave,
  title={{DiffWave}: A Versatile Diffusion Model for Audio Synthesis},
  author={Kong, Zhifeng and Ping, Wei and Huang, Jiaji and Zhao, Kexin and Catanzaro, Bryan},
  booktitle={ICLR},
  year={2021}
}

@inproceedings{
chen2021wavegrad,
title={WaveGrad: Estimating Gradients for Waveform Generation},
author={Nanxin Chen and Yu Zhang and Heiga Zen and Ron J Weiss and Mohammad Norouzi and William Chan},
booktitle={ICLR},
year={2021}
}

@inproceedings{songdenoising,
  title={Denoising diffusion implicit models},
  author={Song, Jiaming and Meng, Chenlin and Ermon, Stefano},
  booktitle={ICLR},
  year={2021}
}

@inproceedings{zhou2024fast,
  title={Fast {ODE}-based sampling for diffusion models in around 5 steps},
  author={Zhou, Zhenyu and Chen, Defang and Wang, Can and Chen, Chun},
  booktitle={CVPR},
  year={2024}
}

@inproceedings{nichol2021improved,
  title={Improved denoising diffusion probabilistic models},
  author={Nichol, Alexander Quinn and Dhariwal, Prafulla},
  booktitle={ICLR},
  year={2021}
}

@inproceedings{ning2023input,
  title={Input Perturbation Reduces Exposure Bias in Diffusion Models},
  author={Ning, Mang and Sangineto, Enver and Porrello, Angelo and Calderara, Simone and Cucchiara, Rita},
  booktitle={ICML},
  year={2023}
}

@Inproceedings{ren2024multi,
  title={Multi-step denoising scheduled sampling: Towards alleviating exposure bias for diffusion models},
  author={Ren, Zhiyao and Zhan, Yibing and Ding, Liang and Wang, Gaoang and Wang, Chaoyue and Fan, Zhongyi and Tao, Dacheng},
  booktitle={AAAI},
  year={2024}
}

@InProceedings{ningelucidating,
  title={Elucidating the exposure bias in diffusion models},
  author={Ning, Mang and Li, Mingxiao and Su, Jianlin and Salah, Albert Ali and Ertugrul, Itir Onal},
  booktitle=  {ICLR},
  year      = {2024}
}

@Inproceedings{li2024alleviating,
  title={Alleviating exposure bias in diffusion models through sampling with shifted time steps},
  author={Mingxiao Li and Tingyu Qu and Ruicong Yao and Wei Sun and Marie-Francine Moens},
  booktitle={ICLR},
  year={2024}
}

@inproceedings{li2023error,
  title={On error propagation of diffusion models},
  author={Li, Yangming and van der Schaar, Mihaela},
  booktitle={ICLR},
  year={2024}
}

@inproceedings{sohl2015deep,
  title={Deep unsupervised learning using nonequilibrium thermodynamics},
  author={Sohl-Dickstein, Jascha and Weiss, Eric and Maheswaranathan, Niru and Ganguli, Surya},
  booktitle={ICML},
  year={2015}
}

@Inproceedings{ho2020denoising,
  title={Denoising diffusion probabilistic models},
  author={Jonathan Ho and Ajay Jain and Pieter Abbeel},
  booktitle={NeurIPS},
  year={2020}
}

@Inproceedings{song2021scorebased,
  title={Score-based generative modeling through stochastic differential equations},
  author={Yang Song and Jascha Sohl-Dickstein and Diederik P Kingma and Abhishek Kumar and Stefano Ermon and Ben Poole},
  booktitle={ICLR},
  year={2021}
}

@inproceedings{baoanalytic,
  title={{Analytic-DPM}: an analytic estimate of the optimal reverse variance in diffusion probabilistic models},
  author={Bao, Fan and Li, Chongxuan and Zhu, Jun and Zhang, Bo},
  booktitle={ICLR},
  year={2022}
}

@inproceedings{kim2023refining,
  title={Refining Generative Process with Discriminator Guidance in Score-based Diffusion Models},
  author={Kim, Dongjun and Kim, Yeongmin and Kwon, Se Jung and Kang, Wanmo and Moon, Il-Chul},
  booktitle={ICML},
  year={2023}
}

@inproceedings{zhao2024unipc,
  title={Unipc: A unified predictor-corrector framework for fast sampling of diffusion models},
  author={Zhao, Wenliang and Bai, Lujia and Rao, Yongming and Zhou, Jie and Lu, Jiwen},
  booktitle={NeurIPS},
  year={2024}
}

@inproceedings{xu2023pfgm++,
  title={{PFGM++}: Unlocking the potential of physics-inspired generative models},
  author={Xu, Yilun and Liu, Ziming and Tian, Yonglong and Tong, Shangyuan and Tegmark, Max and Jaakkola, Tommi},
  booktitle={ICML},
  year={2023},
}

@inproceedings{salimans2022progressive,
title={Progressive Distillation for Fast Sampling of Diffusion Models},
author={Tim Salimans and Jonathan Ho},
booktitle={ICLR},
year={2022}
}

@inproceedings{liu2023instaflow,
  title={Instaflow: One step is enough for high-quality diffusion-based text-to-image generation},
  author={Liu, Xingchao and Zhang, Xiwen and Ma, Jianzhu and Peng, Jian and others},
  booktitle={ICLR},
  year={2023}
}

@inproceedings{heusel2017gans,
  title={{GANs} trained by a two time-scale update rule converge to a local nash equilibrium},
  author={Heusel, Martin and Ramsauer, Hubert and Unterthiner, Thomas and Nessler, Bernhard and Hochreiter, Sepp},
  booktitle={NeurIPS},
  year={2017}
}

@article{krizhevsky2009learning,
  title={Learning multiple layers of features from tiny images},
  author={Krizhevsky, Alex and Hinton, Geoffrey and others},
  year={2009}
}

@article{chrabaszcz2017downsampled,
  title={A downsampled variant of {ImageNet} as an alternative to the {CIFAR} datasets},
  author={Chrabaszcz, Patryk and Loshchilov, Ilya and Hutter, Frank},
  journal={arXiv preprint arXiv:1707.08819},
  year={2017}
}

@inproceedings{bao2022estimating,
  title={Estimating the optimal covariance with imperfect mean in diffusion probabilistic models},
  author={Bao, Fan and Li, Chongxuan and Sun, Jiacheng and Zhu, Jun and Zhang, Bo},
  booktitle={ICML},
  year={2022}
}

@Inproceedings{lu2022dpmsolver,
  title={{DPM}-solver: A fast {ODE} solver for diffusion probabilistic model sampling in around 10 steps},
  author={Cheng Lu and Yuhao Zhou and Fan Bao and Jianfei Chen and Chongxuan Li and Jun Zhu},
  booktitle={NeurIPS},
  year={2022}
}

@Inproceedings{Karras2022edm,
  author    = {Tero Karras and Miika Aittala and Timo Aila and Samuli Laine},
  title     = {Elucidating the design space of diffusion-based generative models},
  booktitle = {NeurIPS},
  year      = {2022}
}

@Inproceedings{goodfellow2014generative,
  title={Generative adversarial nets},
  author={Goodfellow, Ian and Pouget-Abadie, Jean and Mirza, Mehdi and Xu, Bing and Warde-Farley, David and Ozair, Sherjil and Courville, Aaron and Bengio, Yoshua},
  booktitle={NeurIPS},
  year={2014}
}

@Inproceedings{zhang2023lookahead,
  title={Lookahead diffusion probabilistic models for refining mean estimation},
  author={Zhang, Guoqiang and Niwa, Kenta and Kleijn, W Bastiaan},
  booktitle={CVPR},
  year={2023}
}

@inproceedings{
zhang2025antiexposure,
title={Anti-Exposure Bias in Diffusion Models},
author={Junyu Zhang and Daochang Liu and Eunbyung Park and Shichao Zhang and Chang Xu},
booktitle={ICLR},
year={2025}
}

@inproceedings{
yao2025manifold,
title={Manifold constraint reduces exposure bias in accelerated diffusion sampling},
author={Yuzhe YAO and Jun Chen and Zeyi Huang and Haonan Lin and Mengmeng Wang and Guang Dai and Jingdong Wang},
booktitle={ICLR},
year={2025}
}

@article{graps1995introduction,
  title={An introduction to wavelets},
  author={Graps, Amara},
  journal={IEEE Computational Science and Engineering},
  year={1995},
  publisher={IEEE}
}

@inproceedings{vahdat2021score,
  title={Score-based generative modeling in latent space},
  author={Vahdat, Arash and Kreis, Karsten and Kautz, Jan},
  booktitle={NeurIPS},
  year={2021}
}

@inproceedings{
yu2025bias,
title={Bias Mitigation in Graph Diffusion Models},
author={Meng Yu and Kun Zhan},
booktitle={ICLR},
year={2025}
}

@article{luhman2021knowledge,
  title={Knowledge distillation in iterative generative models for improved sampling speed},
  author={Luhman, Eric and Luhman, Troy},
  journal={arXiv preprint arXiv:2101.02388},
  year={2021}
}

@inproceedings{meng2023distillation,
  title={On distillation of guided diffusion models},
  author={Meng, Chenlin and Rombach, Robin and Gao, Ruiqi and Kingma, Diederik and Ermon, Stefano and Ho, Jonathan and Salimans, Tim},
  booktitle={CVPR},
  year={2023}
}

@inproceedings{song2023consistency,
  title={Consistency models},
  author={Song, Yang and Dhariwal, Prafulla and Chen, Mark and Sutskever, Ilya},
  booktitle={ICML},
  year={2023}
}

@inproceedings{
    lu2025simplifying,
    title={Simplifying, stabilizing and scaling continuous-time consistency models},
    author={Cheng Lu and Yang Song},
    booktitle={ICLR},
    year={2025}
}

@inproceedings{song2024improved,
    title={Improved techniques for training consistency models},
    author={Yang Song and Prafulla Dhariwal},
    booktitle={ICLR},
    year={2024}
}

@inproceedings{dockhorn2022genie,
  title={Genie: Higher-order denoising diffusion solvers},
  author={Dockhorn, Tim and Vahdat, Arash and Kreis, Karsten},
  booktitle={NeurIPS},
  year={2022}
}

@article{yu2015lsun,
  title={Lsun: Construction of a large-scale image dataset using deep learning with humans in the loop},
  author={Yu, Fisher and Seff, Ari and Zhang, Yinda and Song, Shuran and Funkhouser, Thomas and Xiao, Jianxiong},
  journal={arXiv preprint arXiv:1506.03365},
  year={2015}
}

@inproceedings{yi2024towards,
  title={Towards understanding the working mechanism of text-to-image diffusion model},
  author={Yi, Mingyang and Li, Aoxue and Xin, Yi and Li, Zhenguo},
  booktitle={NeurIPS},
  year={2024}
}

@inproceedings{yu2025frequency,
  title={Frequency Regulation for Exposure Bias Mitigation in Diffusion Models},
  author={Meng Yu and Kun Zhan},
  booktitle={ACM MM},
  year={2025}
}

@inproceedings{umdon,
  title={Don't Play Favorites: Minority Guidance for Diffusion Models},
  author={Um, Soobin and Lee, Suhyeon and Ye, Jong Chul},
  booktitle={ICLR},
  year={2024}

}

@inproceedings{
miglani2025analysing,
title={Analysing The Spectral Biases in Generative Models},
author={Amitoj Singh Miglani and Shweta Singh and Vidit Aggarwal},
booktitle={The Fourth Blogpost Track at ICLR 2025},
year={2025}
}

@inproceedings{liu2015deep,
  title={Deep learning face attributes in the wild},
  author={Liu, Ziwei and Luo, Ping and Wang, Xiaogang and Tang, Xiaoou},
  booktitle={ICCV},
  year={2015}
}

@inproceedings{mengsdedit,
  title={{SDEdit}: Guided Image Synthesis and Editing with Stochastic Differential Equations},
  author={Meng, Chenlin and He, Yutong and Song, Yang and Song, Jiaming and Wu, Jiajun and Zhu, Jun-Yan and Ermon, Stefano},
  booktitle={ICLR},
  year={2022}
}

@inproceedings{couairondiffedit,
  title={{DiffEdit}: Diffusion-based semantic image editing with mask guidance},
  author={Couairon, Guillaume and Verbeek, Jakob and Schwenk, Holger and Cord, Matthieu},
  booktitle={ICLR},
  year={2023}
}

@inproceedings{parmar2023zero,
  title={Zero-shot image-to-image translation},
  author={Parmar, Gaurav and Kumar Singh, Krishna and Zhang, Richard and Li, Yijun and Lu, Jingwan and Zhu, Jun-Yan},
  booktitle={ACM SIGGRAPH},
  year={2023}
}

@article{saharia2022image,
  title={Image super-resolution via iterative refinement},
  author={Saharia, Chitwan and Ho, Jonathan and Chan, William and Salimans, Tim and Fleet, David J and Norouzi, Mohammad},
  journal={TPAMI},
  year={2022}
}

@article{li2022srdiff,
  title={{Srdiff}: Single image super-resolution with diffusion probabilistic models},
  author={Li, Haoying and Yang, Yifan and Chang, Meng and Chen, Shiqi and Feng, Huajun and Xu, Zhihai and Li, Qi and Chen, Yueting},
  journal={Neurocomputing},
  year={2022}
}

@inproceedings{wangimproved,
  title={Improved Diffusion-based Generative Model with Better Adversarial Robustness},
  author={Wang, Zekun and Yi, Mingyang and Xue, Shuchen and Li, Zhenguo and Liu, Ming and Qin, Bing and Ma, Zhi-Ming},
  booktitle={ICLR},
  year={2025}
}

@inproceedings{peebles2023scalable,
  title={Scalable diffusion models with transformers},
  author={Peebles, William and Xie, Saining},
  booktitle={CVPR},
  year={2023}
}

@inproceedings{qian2024boosting,
  title={Boosting diffusion models with moving average sampling in frequency domain},
  author={Qian, Yurui and Cai, Qi and Pan, Yingwei and Li, Yehao and Yao, Ting and Sun, Qibin and Mei, Tao},
  booktitle={CVPR},
  year={2024}
}

@article{zheng2024open,
  title={{Open-sora}: Democratizing efficient video production for all},
  author={Zheng, Zangwei and Peng, Xiangyu and Yang, Tianji and Shen, Chenhui and Li, Shenggui and Liu, Hongxin and Zhou, Yukun and Li, Tianyi and You, Yang},
  journal={arXiv preprint arXiv:2412.20404},
  year={2024}
}

@inproceedings{blattmann2023align,
  title={Align your latents: High-resolution video synthesis with latent diffusion models},
  author={Blattmann, Andreas and Rombach, Robin and Ling, Huan and Dockhorn, Tim and Kim, Seung Wook and Fidler, Sanja and Kreis, Karsten},
  booktitle={CVPR},
  year={2023}
}

@inproceedings{khachatryan2023text2video,
  title={{Text2video-zero}: Text-to-image diffusion models are zero-shot video generators},
  author={Khachatryan, Levon and Movsisyan, Andranik and Tadevosyan, Vahram and Henschel, Roberto and Wang, Zhangyang and Navasardyan, Shant and Shi, Humphrey},
  booktitle={ICCV},
  year={2023}
}

@misc{blackforestlabs2024flux,
  title={Flux},
  author={{Black Forest Labs}},
  howpublished={\url{https://github.com/black-forest-labs/flux}},
  year={2024}
}

@inproceedings{um2024self,
  title={Self-guided generation of minority samples using diffusion models},
  author={Um, Soobin and Ye, Jong Chul},
  booktitle={ECCV},
  year={2024}
}

@misc{wu2025qwenimagetechnicalreport,
      title={Qwen-Image Technical Report}, 
      author={Chenfei Wu and Jiahao Li and Jingren Zhou and Junyang Lin and Kaiyuan Gao and Kun Yan and Sheng-ming Yin and Shuai Bai and Xiao Xu and Yilei Chen and Yuxiang Chen and Zecheng Tang and Zekai Zhang and Zhengyi Wang and An Yang and Bowen Yu and Chen Cheng and Dayiheng Liu and Deqing Li and Hang Zhang and Hao Meng and Hu Wei and Jingyuan Ni and Kai Chen and Kuan Cao and Liang Peng and Lin Qu and Minggang Wu and Peng Wang and Shuting Yu and Tingkun Wen and Wensen Feng and Xiaoxiao Xu and Yi Wang and Yichang Zhang and Yongqiang Zhu and Yujia Wu and Yuxuan Cai and Zenan Liu},
      year={2025},
      eprint={2508.02324},
      archivePrefix={arXiv},
      primaryClass={cs.CV},
      url={https://arxiv.org/abs/2508.02324}, 
}

@article{lan2025flux,
  title={Flux-text: A simple and advanced diffusion transformer baseline for scene text editing},
  author={Lan, Rui and Bai, Yancheng and Duan, Xu and Li, Mingxing and Jin, Dongyang and Xu, Ryan and Sun, Lei and Chu, Xiangxiang},
  journal={arXiv preprint arXiv:2505.03329},
  year={2025}
}

@article{he2026texts,
  title={TEXTS-Diff: TEXTS-Aware Diffusion Model for Real-World Text Image Super-Resolution},
  author={He, Haodong and Zhan, Xin and Bai, Yancheng and Lan, Rui and Sun, Lei and Chu, Xiangxiang},
  journal={arXiv preprint arXiv:2601.17340},
  year={2026}
}

@article{wang2025editor,
  title={From Editor to Dense Geometry Estimator},
  author={Wang, JiYuan and Lin, Chunyu and Sun, Lei and Liu, Rongying and Nie, Lang and Li, Mingxing and Liao, Kang and Chu, Xiangxiang and Zhao, Yao},
  journal={arXiv preprint arXiv:2509.04338},
  year={2025}
}

@inproceedings{miao2025shining,
  title={Shining yourself: High-fidelity ornaments virtual try-on with diffusion model},
      author={Miao, Yingmao and Huang, Zhanpeng and Han, Rui and Wang, Zibin and Lin, Chenhao and Shen, Chao},
      booktitle={CVPR},
      year={2025}
}

@article{qu2026scale,
  title={From Scale to Speed: Adaptive Test-Time Scaling for Image Editing},
  author={Qu, Xiangyan and Yuan, Zhenlong and Tang, Jing and Chen, Rui and Tang, Datao and Yu, Meng and Sun, Lei and Bai, Yancheng and Chu, Xiangxiang and Gou, Gaopeng and others},
  journal={arXiv preprint arXiv:2603.00141},
  year={2026}
}

@inproceedings{chu2025usp,
  title={Usp: Unified self-supervised pretraining for image generation and understanding},
  author={Chu, Xiangxiang and Li, Renda and Wang, Yong},
  booktitle={ICCV},
  year={2025}
}

@inproceedings{
chen2026stochastic,
title={Stochastic Self-Guidance for Training-Free Enhancement of Diffusion Models},
author={Chubin Chen and Jiashu Zhu and Xiaokun Feng and Nisha Huang and Chen Zhu and Meiqi Wu and Fangyuan Mao and Jiahong Wu and Xiangxiang Chu and Xiu Li},
booktitle={ICLR},
year={2026}
}

@article{chen2026layer,
  title={Layer-wise instance binding for regional and occlusion control in text-to-image diffusion transformers},
  author={Chen, Ruidong and Bai, Yancheng and Zhang, Xuanpu and Zeng, Jianhao and Wang, Lanjun and Song, Dan and Sun, Lei and Chu, Xiangxiang and Liu, Anan},
  journal={arXiv preprint arXiv:2603.05769},
  year={2026}
}

@inproceedings{
lei2026there,
title={There is No {VAE}: End-to-End Pixel-Space Generative Modeling via Self-Supervised Pre-Training},
author={Jiachen Lei and Keli Liu and Julius Berner and Y HoiM and Hongkai Zheng and Jiahong Wu and Xiangxiang Chu},
booktitle={ICLR},
year={2026}
}
}
\appendix
% \clearpage
\setcounter{page}{1}
\maketitlesupplementary

\section{Difference from Prior Works}
\label{app:a}
\begin{figure*}[t]
\centering
\begin{subfigure}[b]{0.33\linewidth}
    \centering
    \includegraphics[width=\linewidth]{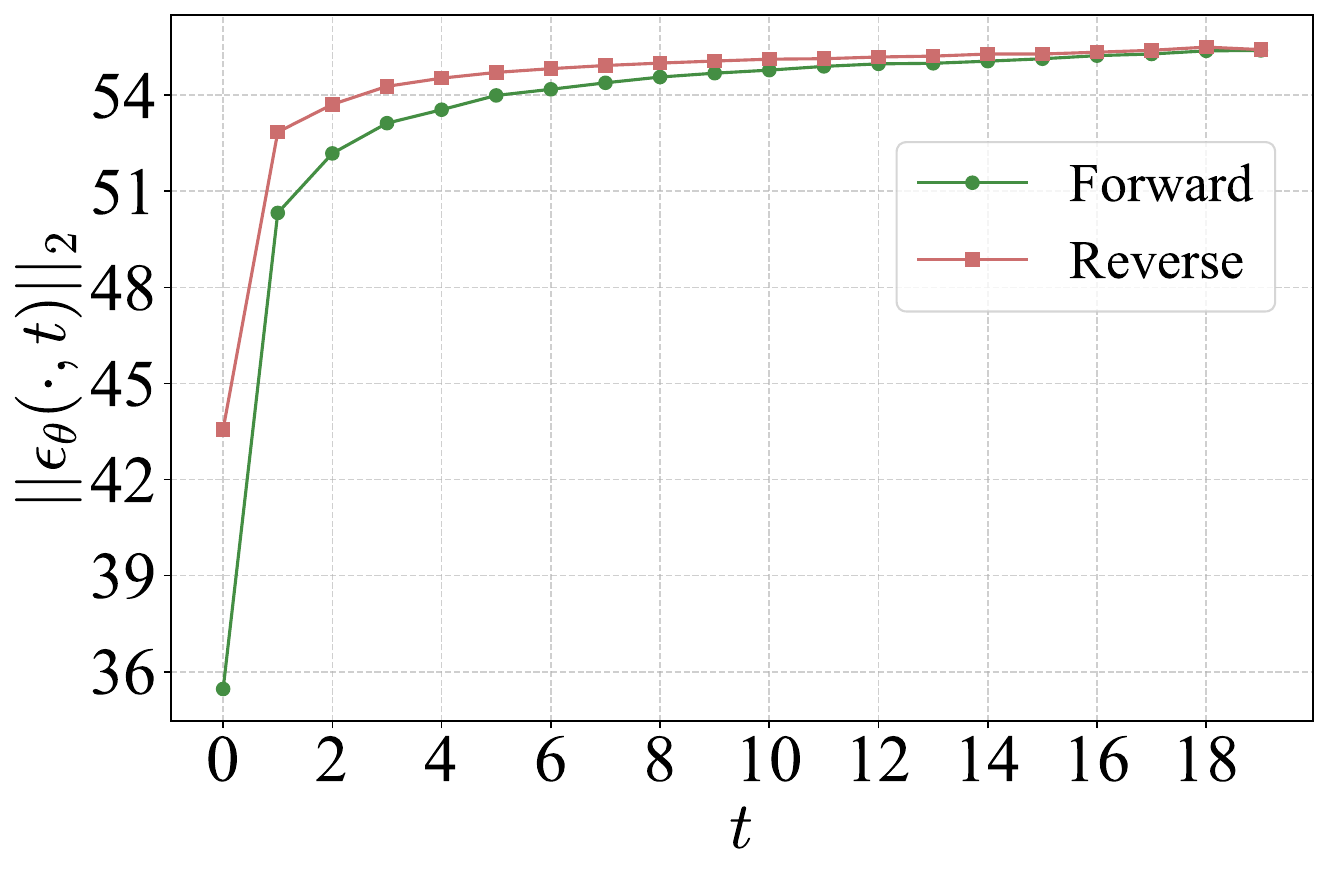}
    \caption{Seed=16, Batch Size=2000}
    \label{fig5a:s16}
\end{subfigure}%
\begin{subfigure}[b]{0.33\linewidth}
    \centering
    \includegraphics[width=\linewidth]{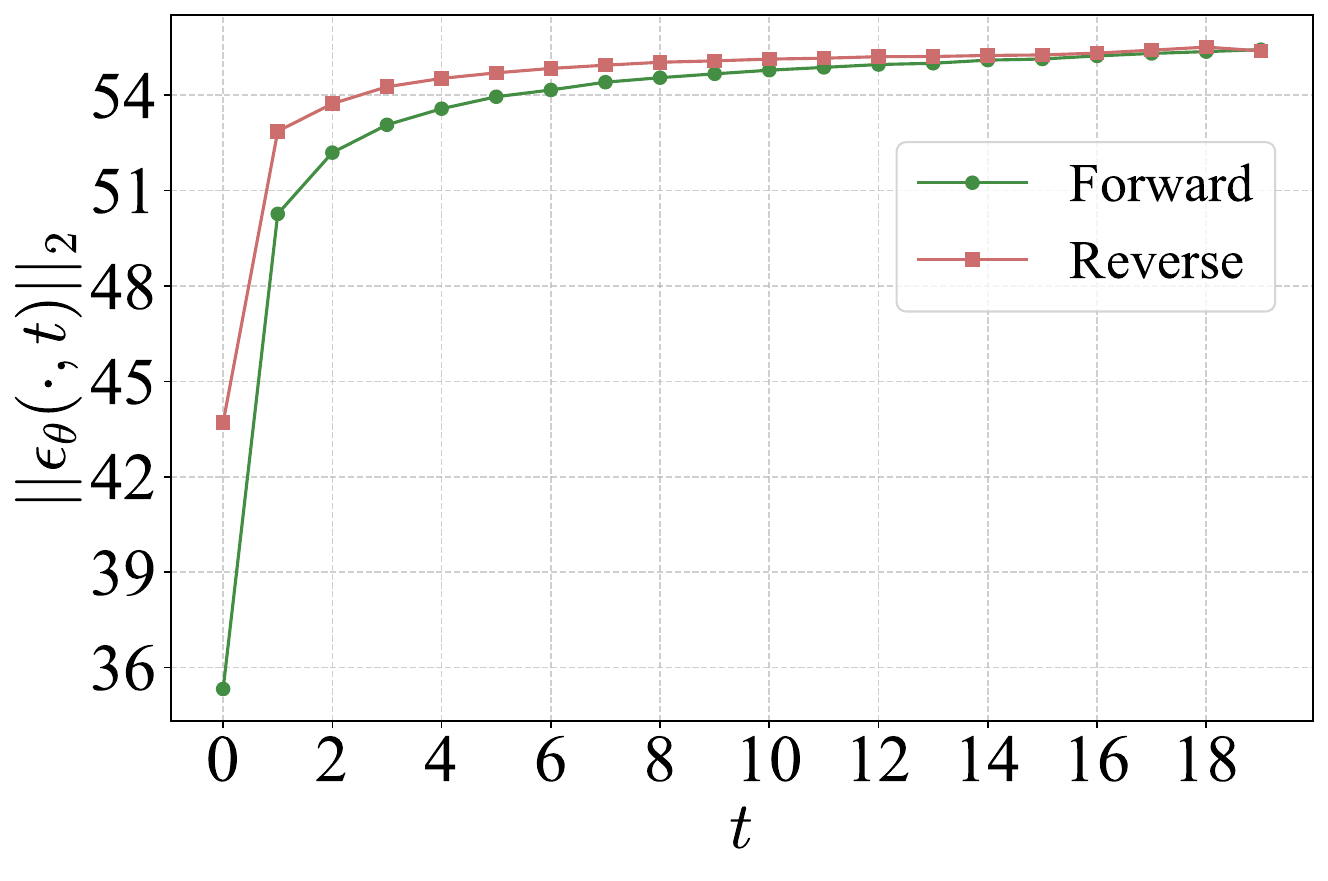}
    \caption{Seed=42, Batch Size=2000}
    \label{fig5b:s42}
\end{subfigure}%
\begin{subfigure}[b]{0.33\linewidth}
    \centering
    \includegraphics[width=\linewidth]{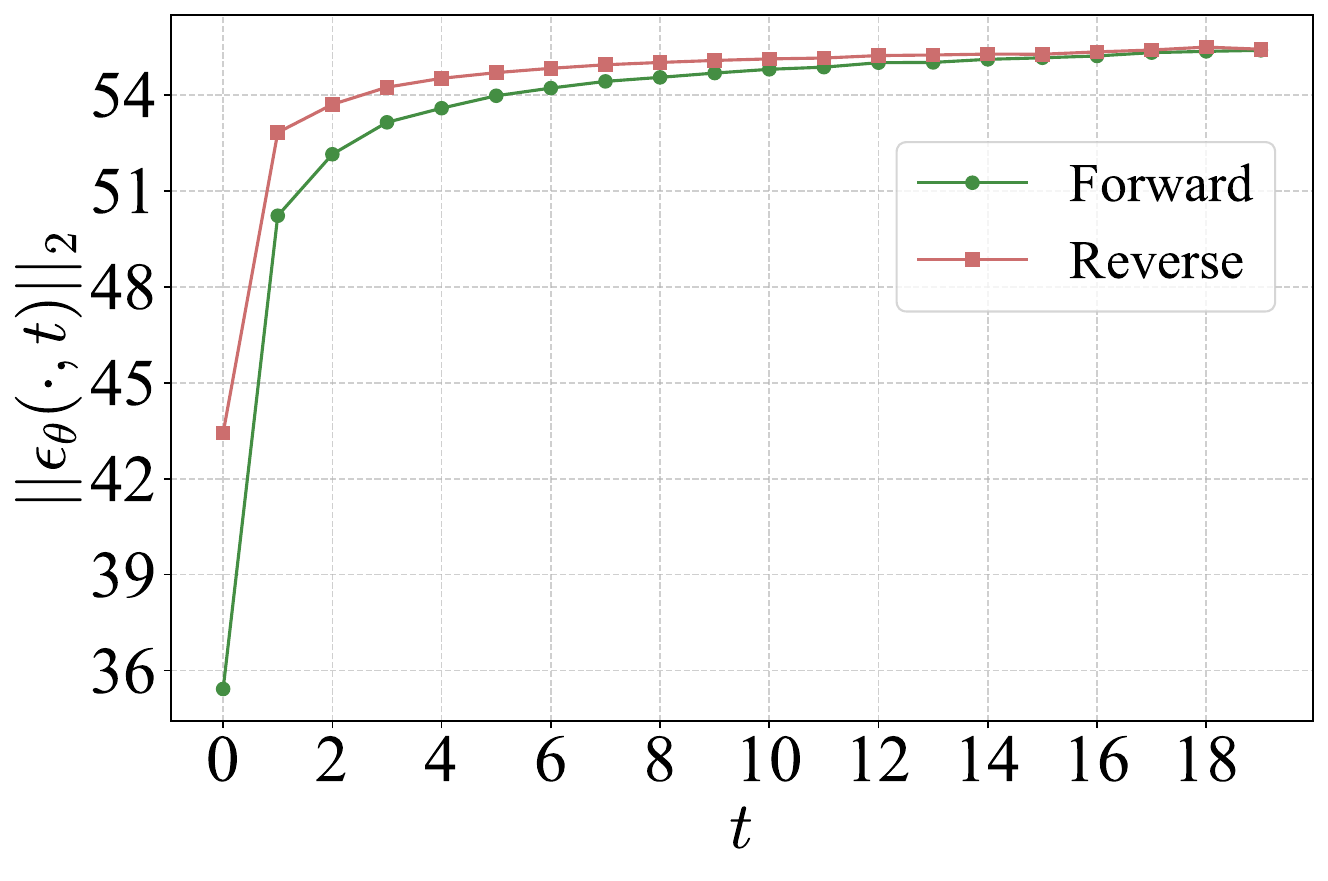}
    \caption{Seed=99, Batch Size=2000}
    \label{fig5c:s99}
\end{subfigure}
\begin{subfigure}[b]{0.33\linewidth}
    \centering
    \includegraphics[width=\linewidth]{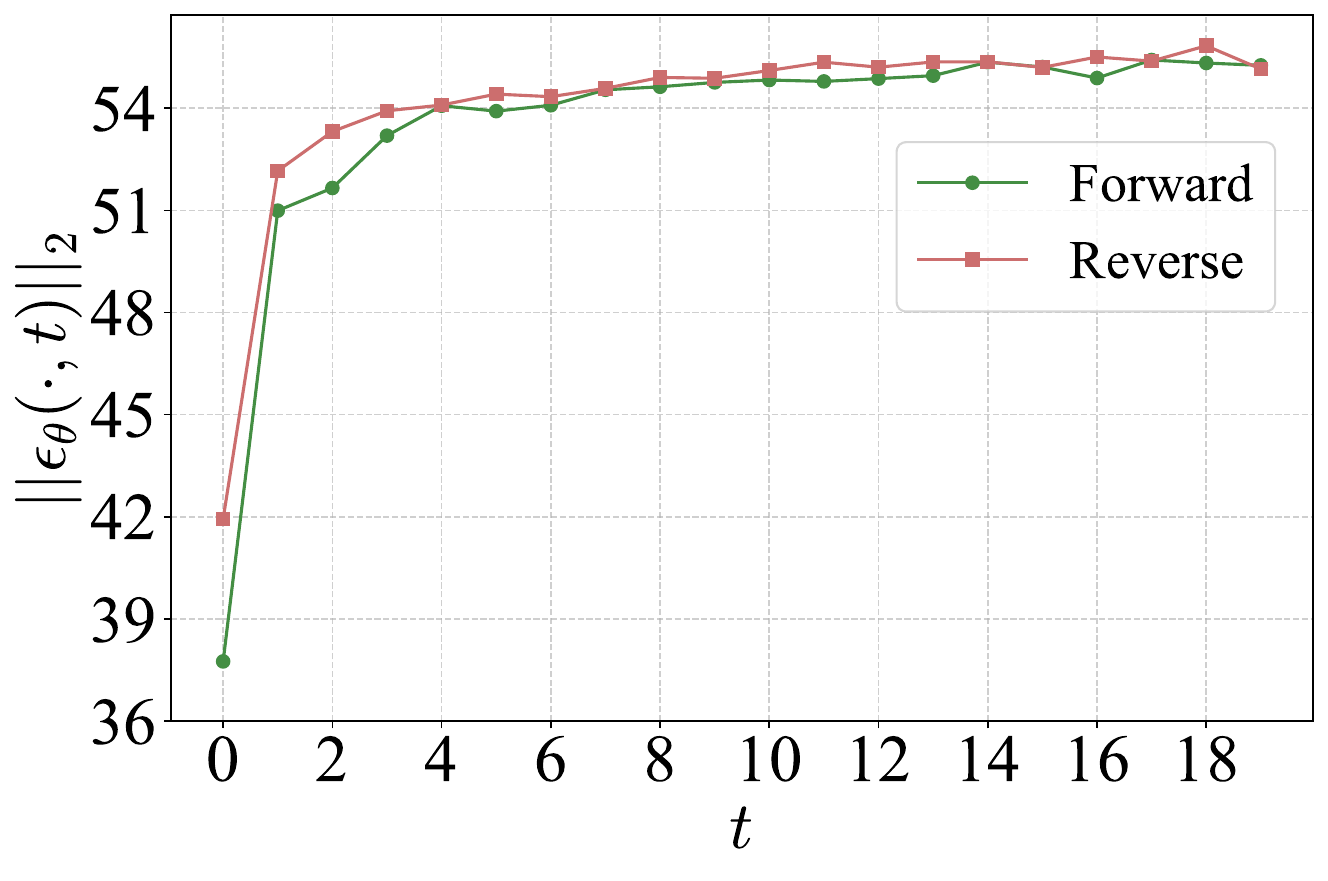}
    \caption{Seed=42, Batch Size=10}
    \label{fig5d:b10}
\end{subfigure}%
\begin{subfigure}[b]{0.33\linewidth}
    \centering
    \includegraphics[width=\linewidth]{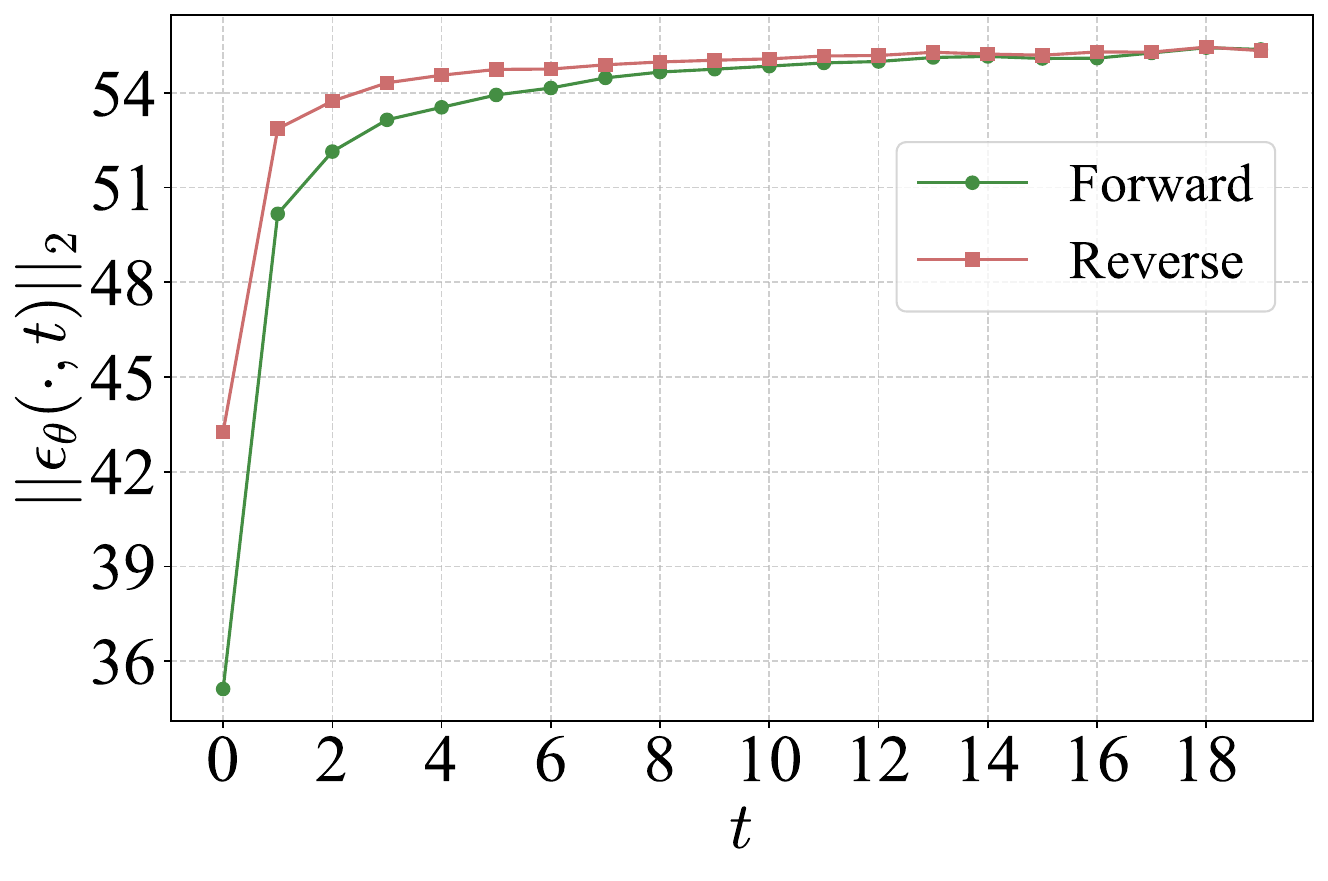}
    \caption{Seed=42, Batch Size=100}
    \label{fig5e:b100}
\end{subfigure}%
\begin{subfigure}[b]{0.33\linewidth}
    \centering
    \includegraphics[width=\linewidth]{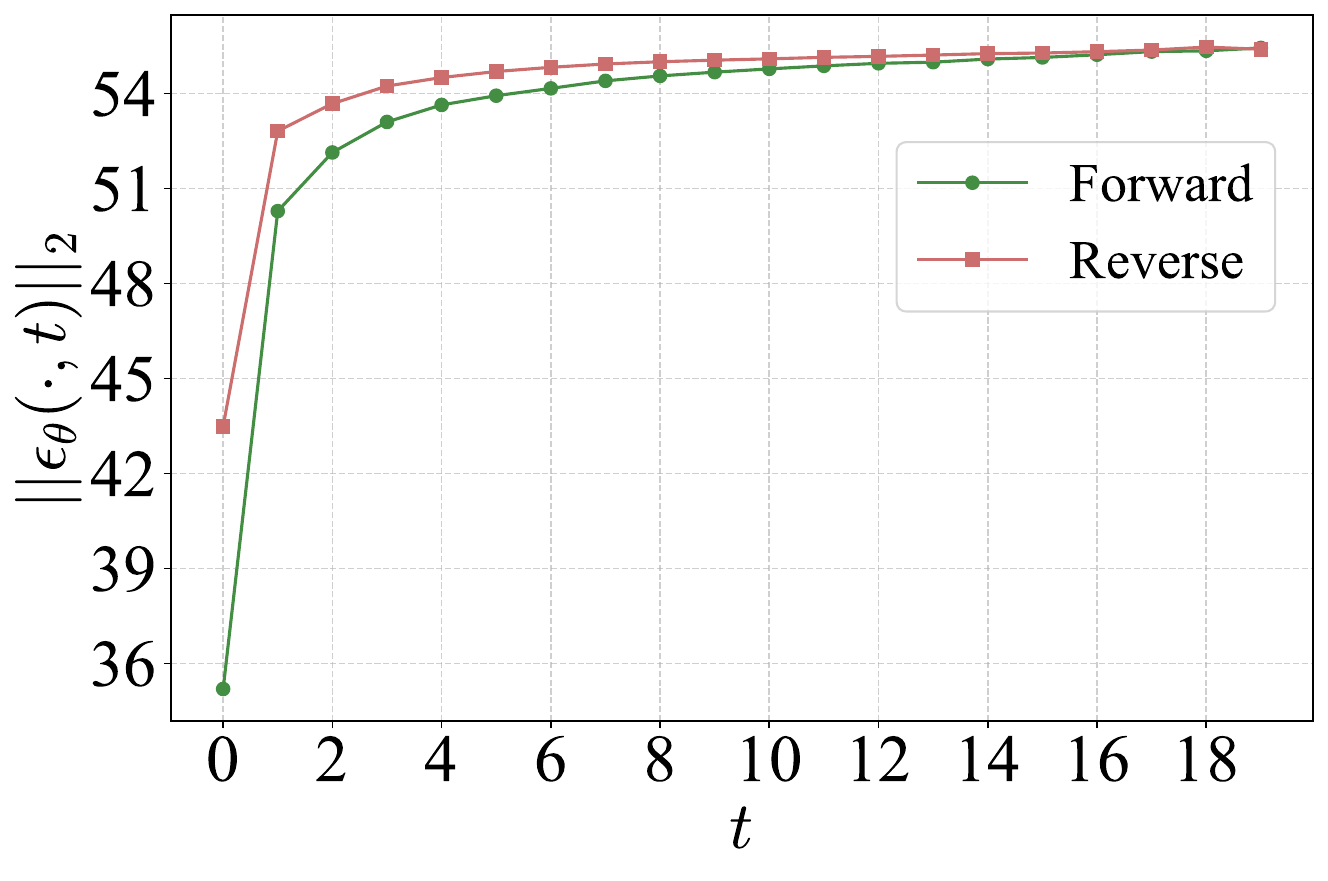}
    \caption{Seed=99, Batch Size=1000}
    \label{fig5f:b1000}
\end{subfigure}
\caption{Robust experimental results for Fig.~\ref{fig1c:sample_bias} with varied random number seeds and sampling batch sizes. These figures show the network output $||\eps_\The(\cdot,t)||_2$ using forward samples $\x_t$ via Eq.~\ref{eq2:forward_onestep} and reverse predicted samples $\hx_t$ via Eq.~\ref{eq7:actual_inference}, respectively. $||\eps_\The(\hx_t,t)||_2$ is always larger than $||\eps_\The(\x_t,t)||_2$ in every figure.} 

\label{fig5:bias}
\end{figure*}

In this section, we outline the differences between the second experiment (Fig.~\ref{fig1c:sample_bias}) in Sec~\ref{sec:4} of this paper and prior work ~\cite{ning2023input,ningelucidating}. We emphasize that ADM-ES~\cite{ningelucidating} only provides a phenomenological conclusion and does not delve into the underlying causes of the phenomenon. In contrast, the SNR-t bias discovered in this paper, along with the sliding window experiments on neural networks based on Fig.~\ref{fig1b:network}, provide in-depth explanations and evidence for this phenomenon. Additionally, this section offers more robust experimental analyses for the phenomenon.

(1) The SNR-t bias is the underlying cause of exposure bias proposed by ADM-IP~\cite{ning2023input} and ADM-ES~\cite{ningelucidating}. ADM-IP and ADM-ES define the exposure bias as an intuitively inter-sample bias between the perturbed sample $\x_t$ and the predicted sample $\hx_t$. Meanwhile, ADM-ES also claims that exposure bias leads to the accumulation of errors, yet it fails to provide fundamental evidence for such error accumulation. In contrast, we explicitly demonstrate when the SNR of the input sample mismatches the timestep, the network’s predictive output exhibits significant errors, as shown in the Key Finding 1 (Fig.~\ref{fig1b:network}). Furthermore, since the SNR of reverse-process samples is consistently lower than the ideal level, as shown in the Key Finding 2 (Fig.~\ref{fig1c:sample_bias}), the network’s predictions during the reverse process are invariably erroneous, specifically manifesting as overestimated outputs. In summary, the SNR-t bias stems primarily from the forced coupling of sample SNR and timestep during training. 

(2) Unlike ADM-ES, this paper focuses on drawing deeper conclusions and uncovering the underlying patterns. Specifically, Figure 2 in ADM-ES concludes that the $L_2$-norm of $\eps_\The(\hx_t,t)$ in the reverse process is always larger than that of $\eps_\The(\x_t,t)$ in the forward process. However, ADM-ES does not explore the deep-seated reasons for this overestimation phenomenon. In this paper, we derive Finding 1 through the sliding window experiments in Sec.~\ref{sec:4}: for the fixed timestep $s$, when handling the sample $\x_t$ with a lower SNR, where $t > s$, the network tends to overestimate the predicted output. Conversely, when dealing with the sample $\x_t$ with a higher SNR, the predicted output is typically underestimated. Therefore, combining the findings of ADM-ES and Finding 1 of this paper, we arrive at Finding 2: Reverse denoising samples often exhibit lower SNR compared to their corresponding forward samples at the same timestep.

(3) Unlike exposure bias, an inter-sample bias, the SNR-t bias is a more specific SNR-timestep bias. Meanwhile, our method based on the SNR-t bias can be naturally integrated into state-of-the-art models for correcting exposure bias, such as ADM-IP, ADM-ES, and DPM-FR, further improving the generation quality of these correction models as shown in Sec.~\ref{sec6.2:biad_corr}. Additionally, our method can significantly enhance the generation quality in the latest text-to-image models, as shown in Appendix E. Thus, these experiments further illustrate the differences between SNR-t bias and exposure bias, as well as the necessity of researching SNR-t bias.

Furthermore, we also provide more robust experimental evidence for Fig.~\ref{fig1c:sample_bias} to eliminate interference caused by random seeds and sampling batch sizes. Specifically, we fix the sampling batch size at 2000 and then select different random number seeds (16, 42, and 99) to obtain distinct sampling trajectories, as illustrated in Figs.~\ref{fig5a:s16}, \ref{fig5b:s42}, and \ref{fig5c:s99}, respectively. Subsequently, we fix the random number seed and vary the sampling batch sizes (10, 100, and 1000), as shown in Figs.~\ref{fig5d:b10}, \ref{fig5e:b100}, and \ref{fig5f:b1000}, respectively. Fig.~\ref{fig5:bias} clearly demonstrates that regardless of the random number seed and sampling batch size, the network output of the reverse process is consistently larger than that of the forward process, which provides more robust evidence for our analysis.

\section{Theoretical evidence of Assumption 5.1}
\label{app:b}
\paragraph{Assumption 5.1.} \textit{During both the forward and reverse processes, the reconstruction sample $\x^0_\The(\x_t,t)$ can be expressed in terms of the original data $\x_0$ as follows:}
\begin{equation}
    \x^0_\The(\hat{\x}_t,t) = \gamma_t \x_0 + \phi_t \eps_t,
\label{eqapp:assumption}
\end{equation}
\textit{where $0 < \gamma_t \leq 1$, $\phi_t < M$, and $M$ denotes a uniform upper bound constant across all timesteps.}

Specifically, we emphasize that both the forward reconstructed sample $\x^0_\The({\x}_t,t)$ and the reverse reconstructed sample $\x^0_\The(\hat{\x}_t,t)$ adhere to the form specified in Eq.~\ref{eqapp:assumption}.

In this section, we present the detailed proof of Assumption 5.1. As stated in the main text, previous work proposed two distinct linear assumptions but lacked supporting evidence. However, we provide both experimental evidence and theoretical proofs to support our findings. Under Gaussian perturbation $\q_\sigma(\y|\x)$, the Tweedie’s formula is 
\begin{equation}
    \mathbb{E}[\x|\y] = \y+\sigma^2  \nabla_\y \text{log} \q_\sigma(\y),
    \label{eq:tweedie}
\end{equation}
where $\q_\sigma(\y):= \int \q(\y|\x)\q(\x)\D\x$. Now, by substituting the forward perturbation distribution $\q(\x_t|\x_0)$ of DPMs into Eq.~\ref{eq:tweedie}, we can obtain:
 \begin{equation}
\mathbb{E}[\x_0|\x_t]=\frac{\x_t+({1-\bar{\alpha}_t})\nabla_{\x_t} \text{log} \q(\x_t)}{\sqrt{\bar{\alpha}_t}}.
\label{eq:posterior_mean}
\end{equation}
Based on the relationship between the score and the noise $\s_\The(\x_t,t)=-\frac{\eps_\The(\x_t,t)}{\sqrt{1-\bar{\alpha}_t}}$, we further derive:
\begin{equation}
\mathbb{E}[\x_0|\x_t]=\frac{\x_t-\sqrt{1-\bar{\alpha}_t}\eps_\The(\x_t,t)}{\sqrt{\bar{\alpha}_t}}=\x^0_\The(\x_t,t),
\label{eq:tweedie_x0}
\end{equation}
which clearly demonstrates that the reconstructed sample $\x^0_\The(\x_t,t)$ is essentially the posterior mean based on the Tweedie formula. Furthermore, the score network trained with the L2 norm-MSE loss function always have a theoretical analytical solution~\cite{umdon}, which is also the posterior mean:
\begin{equation}
    \s_\The(\x_t,t) = \mathbb{E}_{q(\x_0|\x_t)} \left[ \nabla_{\x_t} \log q(\x_t \mid \x_0) \right].
\end{equation}
Based on the equivalence between the score and noise, the optimal solution for noise prediction is also the same posterior mean. Therefore, based on the mean tendency of denoising operations and network predictions, we can regard $\x^0_\The(\x_t,t)$ as the mean estimate $\bar{\x}_0$ of $\x_0$.

The variance formula is expressed as:
\begin{equation}
    \mathbb{E}[\| \x_0 \|^2] = \| \bar{\x}_0 \|^2 + \text{Var}(\|\x_0\|).
\end{equation}

Based on the non-negativity of the variance, we obtain:
\begin{equation*}
    \| \bar{\x}_0 \|^2 \leq \mathbb{E}[\| \x_0 \|^2].
\end{equation*}
We substitute $\x^0_\The(\x_t,t)$ for $\bar{\x}_0$, then given that the expectation of a constant is the constant itself, we can take the expectation of both sides of the above equation to obtain:

\begin{equation}
  \mathbb{E}[\| \x^0_\The(\x_t,t) \|^2 ]\leq \mathbb{E}[\| \x_0 \|^2].
  \label{eq:xiaoyu}
\end{equation}
Eq.~\ref{eq:xiaoyu} clearly demonstrates that the L2 norm of reconstructed samples is always smaller than that of real samples, which indicates that the reconstruction operation is always accompanied by information loss. 

However, previous work~\cite{ningelucidating,li2024alleviating} argues that reconstructed samples should be modeled as:
\begin{equation}
    \x^0_\The(\x_t,t) = \x_0 + \phi_t \eps_t,
\end{equation}
which is clearly inconsistent with Eq.~\ref{eq:xiaoyu}. Thus, We use the form in Eq.~\ref{eqapp:assumption}, consistent with the assumption of LA-DPM~\cite{zhang2023lookahead} and DPM-FR~\cite{yu2025frequency}.

In addition, we also provide experimental evidence for the above proof. Following the experimental setup described in Sec.~\ref{sec:4}, we perform the following operations sequentially:
(1) We generate perturbed samples $\{\x_1,\x_2,\ldots,\x_T\}$ via Eq.~\ref{eq2:forward_onestep}, and feed $\x_t$ and timestep $t$ into the network to obtain $\eps_\The(\x_t,t)$ to compute $\x_\The^0(\x_t,t)$ via Eq.~\ref{eq5:reconstruction}.  
(2) Then, we initialize 2,000 standard Gaussian noise and iteratively denoise operation via Eq.~\ref{eq7:actual_inference} to obtain samples $\{\hx_1,\hx_2,\ldots,\hx_T\}$ and corresponding network outputs $\eps_\The(\hx_t,t)$ to compute $\x_\The^0(\hx_t,t)$ via Eq.~\ref{eq5:reconstruction}.  
(3) Finally, we compute and plot the expectation of $||\x_\The^0(\x_t,t)||_2^2$, $||\x_\The^0(\hx_t,t)||_2^2$, and $||\x_0||_2^2$. 

\begin{figure}[!t]
\centering
   \includegraphics[width=0.90\linewidth]{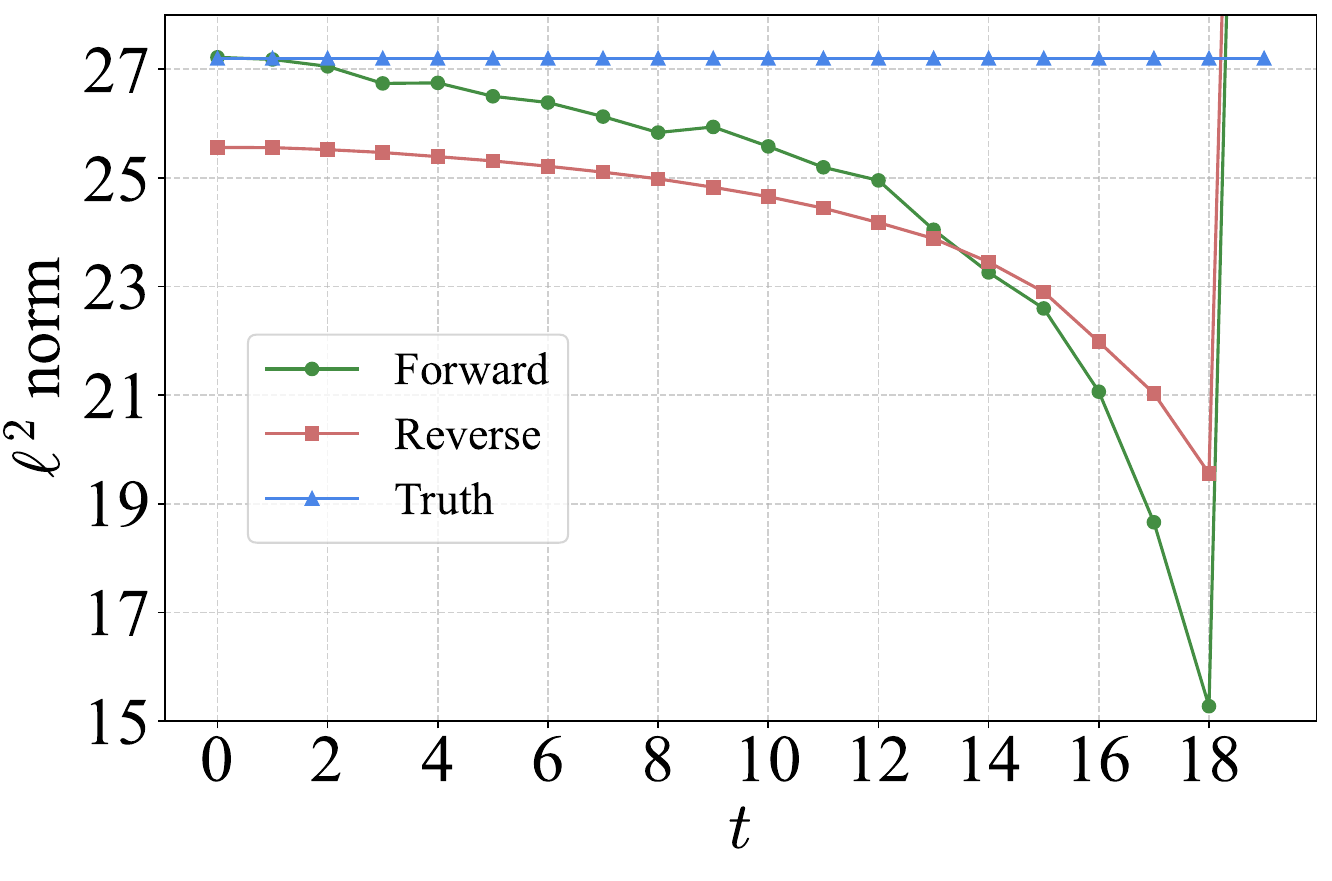} 
\caption{The expectation of $||\x_\The^0(\x_t,t)||_2^2$, $||\x_\The^0(\hx_t,t)||_2^2$, together with the ground-truth norm of $\x_0$.}
\label{fig.x0}
\end{figure}

Fig.~\ref{fig.x0} clearly demonstrates that DPMs fail to fully reconstruct real data $\x_0$, both in the forward and reverse processes. This further indicates that the reconstruction operation incurs information loss. Notably, similar experiments are also reported in DPM-FR~\cite{yu2025frequency}. However, it focuses on the differences between the forward and reverse processes, whereas our work places greater emphasis on whether DPMs can fully reconstruct real data. Additionally, we argue that conducting experiments in the data space is more persuasive. This experiment further demonstrates that $\x^0_\The({\x}_t,t)$ and $\x^0_\The(\hat{\x}_t,t)$ adhere to the form specified in Eq.~\ref{eqapp:assumption}.

\section{Proofs of Theorem 5.1 and Eq.~\ref{eq14: re_predicted_result}}
\label{app:c}
In this section, we present the detailed proofs of Theorem 5.1 and Eq.~\ref{eq14: re_predicted_result}. Our derivation process is mainly based on DPM-FR~\cite{yu2025frequency}. However, we provide a more rigorous derivation process, particularly for $\gamma_t$ and $\hat{\gamma}_t$. Specifically, we focus on SNR, the core theme of this work.

\paragraph{Theorem 5.1.} \textit{For a specific timestep $t$ in the reverse denoising process of DPMs, the SNR of the biased denoising sample $\hx_t$ is given by:}
\begin{equation}
    \text{SNR}(t)=\hat{\gamma}^2_{t}{\bar{\alpha}_{t}}/\big(1 - \bar{\alpha}_{t} +
        (
            \frac{\sqrt{\bar{\alpha}_{t}}\beta_{t+1}}{1-\bar{\alpha}_{t+1}}\phi_{t+1})^2
        \big),
\label{eq12:actural_snr}
\end{equation}
\textit{where $0 < \hat{\gamma}_t \leq 1$ and $\phi_{t+1}$ is derived from the reconstruction model $\x^0_\The(\hat{\x}_{t+1},t+1)$ in Eq.~\ref{eq10:assumption}.}

Firstly, we emphasize that all subsequent noise terms $\eps$ follow the standard Gaussian distribution. We rewrite the fundamental formula of DPMs and the forward noising process is expressed as:
\begin{equation}
    \x_t=\sqrt{\bar{\alpha}_t}\x_0+\sqrt{1-\bar{\alpha}_t}\eps_0.
\label{eqapp:forward_one}
\end{equation}
We assume the current predicted sample is ideal. Thus, the reverse denoising process is expressed as:
\begin{equation}
{\hx}_{t-1} = \frac{1}{\sqrt{\alpha_t}} \left({\x}_t - \frac{1 - \alpha_t}{\sqrt{1 - \bar{\alpha}_t}} \eps_\The({\x}_t, t) \right) + \sigma_t \eps_1.
\label{eqapp:reverse_one}
\end{equation}
Then, substituting Eq.~\ref{eq:tweedie_x0} into Eq.~\ref{eqapp:reverse_one}, we can obtain an equivalent form of the reverse denoising process:
\begin{equation}
    \hat{\x}_{t-1} = \frac{\sqrt{\bar{\alpha}_{t - 1}}\beta_{t}}{1 - \bar{\alpha}_{t}}\x^0_\The(\x_t,t) + \frac{\sqrt{\alpha_{t}}(1 - \bar{\alpha}_{t - 1})}{1 - \bar{\alpha}_{t}}\x_{t} + \sqrt{\tilde{\beta}_t}\eps_{1},
\label{eqapp:reverse_x0}
\end{equation}
By substituting Eqs.~\ref{eqapp:forward_one} and \ref{eqapp:assumption} into Eq.~\ref{eqapp:reverse_x0} to replace $\x_\The^0(\x_t,t)$ and $\x_t$, we can obtain:
\begin{equation}
   \begin{aligned}
 &\hat{\x}_{t-1} = \frac{\sqrt{\bar{\alpha}_{t - 1}}\beta_{t}}{1 - \bar{\alpha}_{t}}(\gamma_t\x_0 + \phi_t\eps_t) + \\
 &\frac{\sqrt{\alpha_{t}}(1 - \bar{\alpha}_{t - 1})}{1 - \bar{\alpha}_{t}}(\sqrt{\bar{\alpha}_t}\x_0 + \sqrt{1 - \bar{\alpha}_t}\eps_0) + \sqrt{\tilde{\beta}_t}\eps_{1}\\ &= \left(\frac{\sqrt{\bar{\alpha}_{t - 1}}\beta_{t}\gamma_t}{1 - \bar{\alpha}_{t}} + \frac{\sqrt{\alpha_{t}}(1 - \bar{\alpha}_{t - 1})\sqrt{\bar{\alpha}_t}}{1 - \bar{\alpha}_{t}}\right)\x_0 + \sqrt{\tilde{\beta}_t}\eps_{1} \\ &+ \frac{\sqrt{\bar{\alpha}_{t - 1}}\beta_{t}\phi_t}{1 - \bar{\alpha}_{t}}\eps_t + \frac{\sqrt{\alpha_{t}}(1 - \bar{\alpha}_{t - 1})\sqrt{1 - \bar{\alpha}_t}}{1 - \bar{\alpha}_{t}}\eps_0\\
 % &=\hat{\gamma}_{t-1} \x_0 + \psi_{t-1} \eps_{t-1}
 \label{eq23:tui}
  \end{aligned} 
 \end{equation}
 
For Eq.~\ref{eq23:tui}, we first focus on the coefficient of $\x_0$:
\begin{equation}
    \begin{aligned}
&\frac{\sqrt{\bar{\alpha}_{t - 1}}\beta_{t}\gamma_t}{1 - \bar{\alpha}_{t}} + \frac{\sqrt{\alpha_{t}}(1 - \bar{\alpha}_{t - 1})\sqrt{\bar{\alpha}_t}}{1 - \bar{\alpha}_{t}}  \\
&=\frac{\sqrt{\bar{\alpha}_{t - 1}}\big((1-{\alpha}_{t})\gamma_t+{\alpha_{t}}(1 - \bar{\alpha}_{t - 1})\big)}{1 - \bar{\alpha}_{t}}.
\label{eq24}
\end{aligned}
\end{equation}

Given that $\gamma_t \leq 1$, we use the scaling method to amplify it to 1, yielding the following inequality:
\begin{equation}
    \begin{aligned}
&\frac{\sqrt{\bar{\alpha}_{t - 1}}\big((1-{\alpha}_{t})\gamma_t+{\alpha_{t}}(1 - \bar{\alpha}_{t - 1})\big)}{1 - \bar{\alpha}_{t}}\\ &\leq\frac{\sqrt{\bar{\alpha}_{t - 1}}\big((1-{\alpha}_{t})+{\alpha_{t}}(1 - \bar{\alpha}_{t - 1})\big)}{1 - \bar{\alpha}_{t}}\\
&=\sqrt{\bar{\alpha}_{t - 1}}
\label{eq25}
    \end{aligned}
\end{equation}

Given that $1-\alpha_{t}>0, \gamma_t\leq1$, We may rigorously define a novel coefficient $\hat{\gamma}_{t-1}\leq1$ for $\hx_{t-1}$ where
\begin{equation}
    \begin{aligned}
        \hat{\gamma}_{t-1}\sqrt{\bar{\alpha}_{t - 1}} = \frac{\sqrt{\bar{\alpha}_{t - 1}}\big((1-{\alpha}_{t})\gamma_t+{\alpha_{t}}(1 - \bar{\alpha}_{t - 1})\big)}{1 - \bar{\alpha}_{t}}.
    \end{aligned}
    \label{eq26}
\end{equation}

For the standard Gaussian noise component in Eq.~\ref{eq23:tui}, based on the properties of the Gaussian distribution, we define a new coefficient $\hat{\psi}_{t-1}$ such that:

\begin{equation}
    \begin{aligned} &\hat{\psi}_{t-1}=(\frac{\sqrt{\bar{\alpha}_{t-1}}\beta_{t}}{1-\bar{\alpha}_{t}}\phi_{t})^{2}+(\frac{\sqrt{\alpha_{t}}(1-\bar{\alpha}_{t-1})}{1-\bar{\alpha}_{t}}\sqrt{1-\bar{\alpha}_{t-1}})^{2}+\tilde{\beta}_{t}\\  &=(\frac{\sqrt{\bar{\alpha}_{t-1}}\beta_{t}}{1-\bar{\alpha}_{t}}\phi_{t})^{2}+\frac{\alpha_{t}(1-\bar{\alpha}_{t-1})^{2}}{1-\bar{\alpha}_{t}}+\frac{(1-\bar{\alpha}_{t-1})(1-\alpha_{t})}{1-\bar{\alpha}_{t}}\\ &=(\frac{\sqrt{\bar{\alpha}_{t-1}}\beta_{t}}{1-\bar{\alpha}_{t}}\phi_{t})^{2}+\frac{\alpha_{t}(1-\bar{\alpha}_{t-1})^{2}+(1-\bar{\alpha}_{t-1})(1-\alpha_{t})}{1-\bar{\alpha}_{t}}\\&=(\frac{\sqrt{\bar{\alpha}_{t-1}}\beta_{t}}{1-\bar{\alpha}_{t}}\phi_{t})^{2}+1-\bar{\alpha}_{t-1}.
    \label{eq27}
    \end{aligned}
\end{equation}
Based on Eqs.~\ref{eq26} and ~\eqref{eq27}, we can obtain 
\begin{equation}
    \begin{aligned}
        \hat{\x}_{t-1} =\hat{\gamma}_{t-1}\sqrt{\bar{\alpha}_{t - 1}}\x_0+\sqrt{1-\bar{\alpha}_{t-1}+(\frac{\sqrt{\bar{\alpha}_{t-1}}\beta_{t}}{1-\bar{\alpha}_{t}}\phi_{t})^{2}}\eps_{t-1}
    \end{aligned}
\label{eqapp:x_t-1}
\end{equation}
Ultimately, based on Eq.~\ref{eqapp:x_t-1}, we obtain the SNR of $\x_{t-1}$ as:
\begin{equation}
    \text{SNR}(t-1)=\hat{\gamma}^2_{t-1}{\bar{\alpha}_{t-1}}/\big(1 - \bar{\alpha}_{t-1} +
        (
            \frac{\sqrt{\bar{\alpha}_{t-1}}\beta_{t}}{1-\bar{\alpha}_{t}}\phi_{t})^2
        \big)
\label{eqapp:actural_snr}
\end{equation}
By replacing the timestep in Eq.~\ref{eqapp:actural_snr}, we ultimately obtain the actual SNR of $\hx_t$ to complet the proof.

% \begin{equation}
%     \begin{aligned}
%         \hat{\x}_{t-1} &=\gamma_{t-1}\sqrt{\bar{\alpha}_{t - 1}}\x_0+\\ &\sqrt{\gamma_t^2(1-\bar{\alpha}_{t-1})  +(\frac{\sqrt{\bar{\alpha}_{t-1}}\beta_{t}}{1-\bar{\alpha}_{t}}\phi_{t})^{2}+(1-\gamma_t^2)(1-\bar{\alpha}_{t-1})}\eps_{t-1}\\
%       &=\gamma_{t-1}\sqrt{\bar{\alpha}_{t - 1}}\x_0+\sqrt{1-\bar{\alpha}_{t-1}+(\frac{\sqrt{\bar{\alpha}_{t-1}}\beta_{t}}{1-\bar{\alpha}_{t}}\phi_{t})^{2}}\eps_{t-1}
%     \end{aligned}
% \end{equation}
To obtain a more concise and intuitive form, we use the piecing-together method to derive:
\begin{equation*}
    \begin{aligned}
        \hat{\psi}_{t-1}^2&=(\frac{\sqrt{\bar{\alpha}_{t-1}}\beta_{t}}{1-\bar{\alpha}_{t}}\phi_{t})^{2}+(1-\hat{\gamma}_{t-1}^2)(1-\bar{\alpha}_{t-1}) \\ &+\hat{\gamma}_{t-1}^2(1-\bar{\alpha}_{t-1}) 
    \end{aligned}
\end{equation*}
In conclusion, we have obtained the biased mean and variance of the reverse process:
\begin{equation}
    \begin{aligned}
        &\hat{\x}_{t-1} =\hat{\gamma}_{t-1}\sqrt{\bar{\alpha}_{t - 1}}\x_0++\hat{\gamma}_{t-1}\sqrt{(1-\bar{\alpha}_{t-1})}\hat{\eps}_3\\ &+\sqrt{(\frac{\sqrt{\bar{\alpha}_{t-1}}\beta_{t}}{1-\bar{\alpha}_{t}}\phi_{t})^{2}+(1-\hat{\gamma}_{t-1}^2)(1-\bar{\alpha}_{t-1})}\tilde{\eps}_3\\
    &=\hat{\gamma}_{t-1}\x_t +\psi_{t-1}\eps_3,
    \end{aligned}
\label{eqapp:snr_t-1_new}
\end{equation}
where $\psi_{t-1}=\sqrt{(\frac{\sqrt{\bar{\alpha}_{t-1}}\beta_{t}}{1-\bar{\alpha}_{t}}\phi_{t})^{2}+(1-\hat{\gamma}_{t-1}^2)(1-\bar{\alpha}_{t-1})}$. Thus, we have completed the proof of Eq.~\ref{eq14: re_predicted_result}. Finally, we emphasize again that $\gamma_t$ is the coefficient of the reconstruction sample $\x^0_\The(\x_t,t)$ in Eq.~\ref{eqapp:assumption}, and $\hat{\gamma}_{t-1}$ is the coefficient of the predicted sample $\hx_{t-1}$ in Eqs.~\ref{eqapp:x_t-1} and \ref{eqapp:snr_t-1_new}.

\section{Weight Strategy Design}
\label{app:d}
The denoising process of DPM inherently follows a coarse-to-fine paradigm: the early stages primarily generate low-frequency global structures, while the later stages progressively recover high-frequency details. To this end, our proposed differential correction method is designed to align with this intrinsic property, prioritizing low-frequency correction in the initial phases and shifting focus to high-frequency correction in the later stages. 

Based on the above reasoning, we assign larger correction coefficients to low-frequency components in the early stage of denoising and higher weighting coefficients to high-frequency components in the later stage of denoising. On this basis, we propose three weighting scheduling strategies.

Firstly, considering that the variance $\sigma_t$ in the reverse process of DPM can dynamically characterize the denoising progress, we adopt the weighting forms shown in Eqs.~\ref{eq17:low} and \ref{eq18:high} in the main text. Second, we design a piecewise weighting strategy. For the timestep $t$ $(0 \leq t < T)$ and threshold $t_s$, based on empirical experience, we classify $t > t_{\text{s}}$ as the early stage of denoising and $t \leq t_s$ as the later stage of denoising. Accordingly, the piecewise weight for low-frequency components can be defined as:
\begin{equation}
    w_t^l = w_l \cdot \mathbb{I}\{t \geq t_s\},
\end{equation}
where $\mathbb{I}(\cdot)$ denotes the indicator function. In a similar vein, the piecewise weight for high-frequency components is naturally defined as:
\begin{equation}
    w_t^h = w_h \cdot \mathbb{I}\{t < t_s\}.
\end{equation}
Furthermore, to simplify the implementation, we also design a constant weighting strategy, where the weights remain unchanged throughout the entire denoising process.

In particular, we emphasize that all three aforementioned weighting strategies are effective after extensive experimental evaluations, as shown in Sec.\ref{sec:5}. Specifically, the variance-based scheduling strategy and the piecewise weighting strategy achieve superior generation quality, which further demonstrates the necessity of aligning the weight design with the denoising dynamics of DPMs.

\section{Additional Results}
\label{app:e}

\begin{figure*}[h] 
    \centering
    \begin{subfigure}[b]{0.28\textwidth} 
        \includegraphics[width=\linewidth]{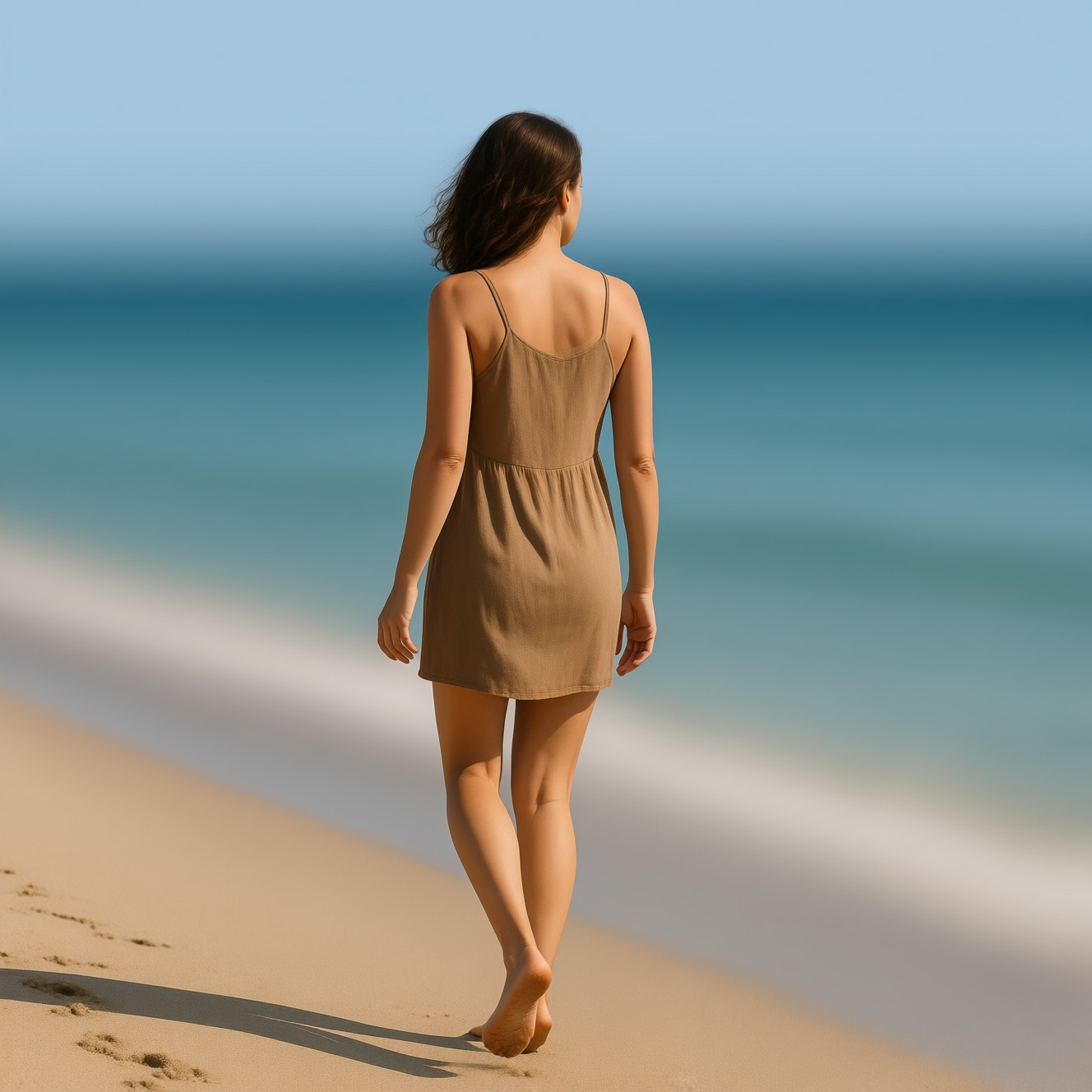}
    \end{subfigure}
    \begin{subfigure}[b]{0.28\textwidth} 
        \includegraphics[width=\linewidth]{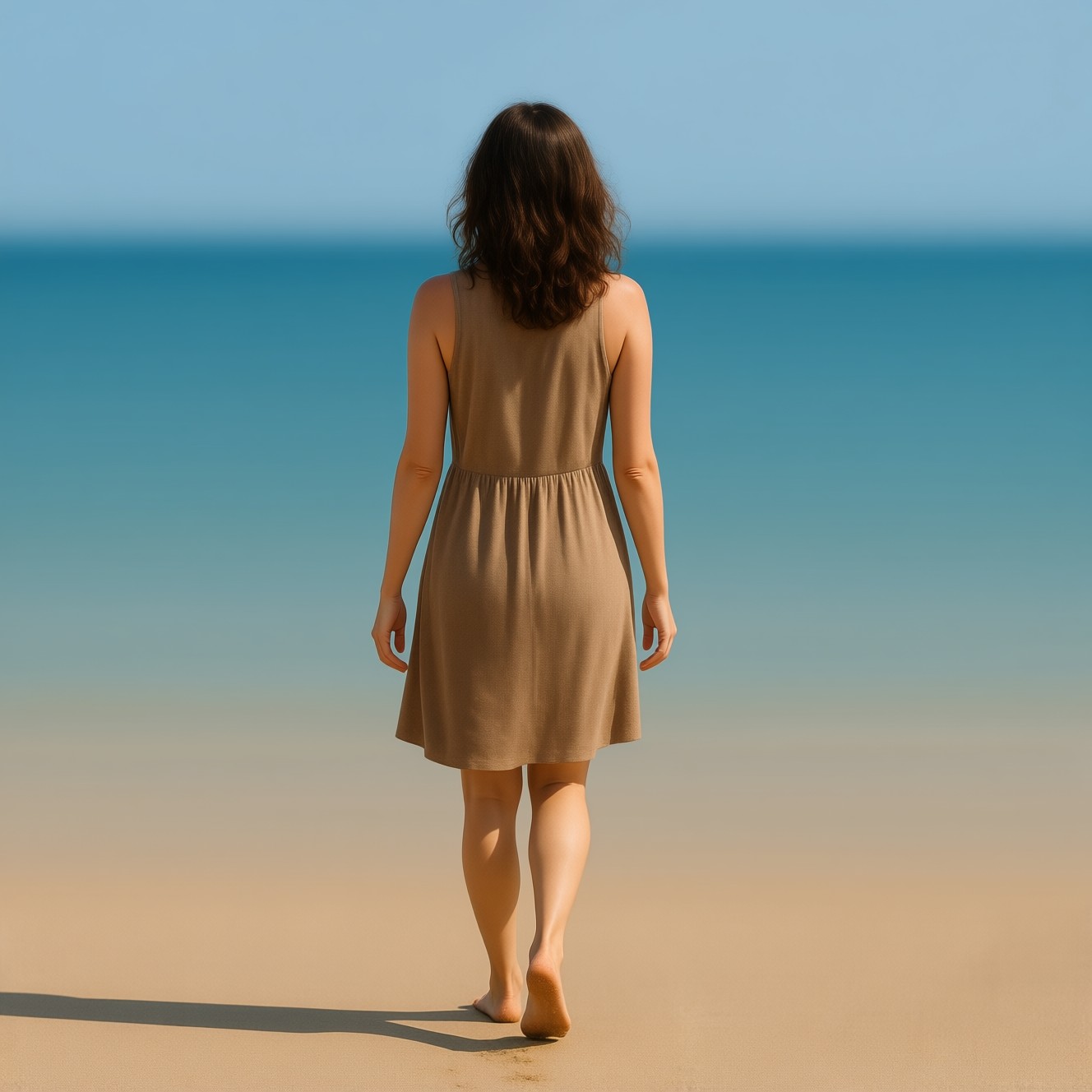}
    \end{subfigure}
    \begin{subfigure}[b]{0.28\textwidth} 
        \includegraphics[width=\linewidth]{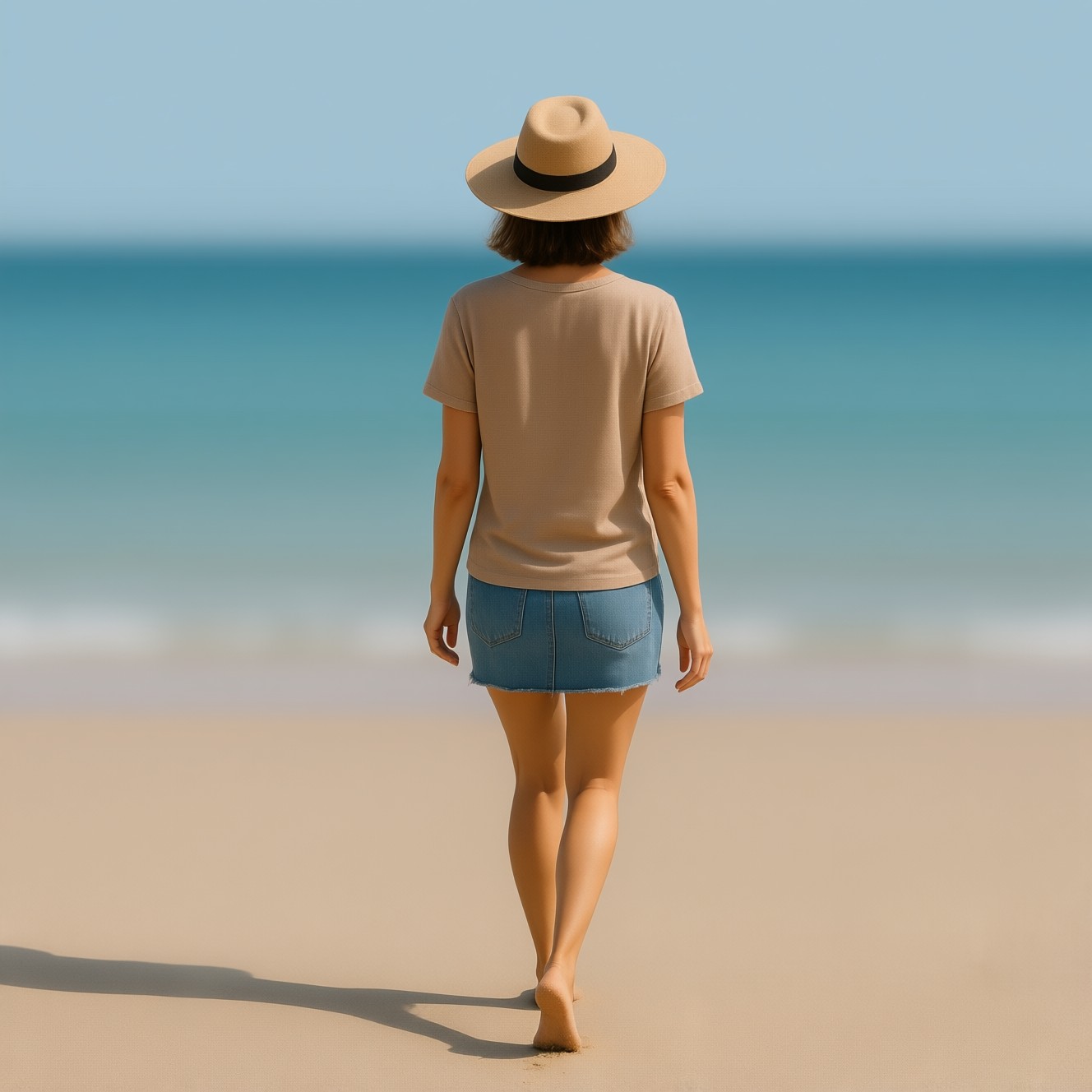}
    \end{subfigure}
    \begin{subfigure}[b]{0.28\textwidth}
        \includegraphics[width=\linewidth]{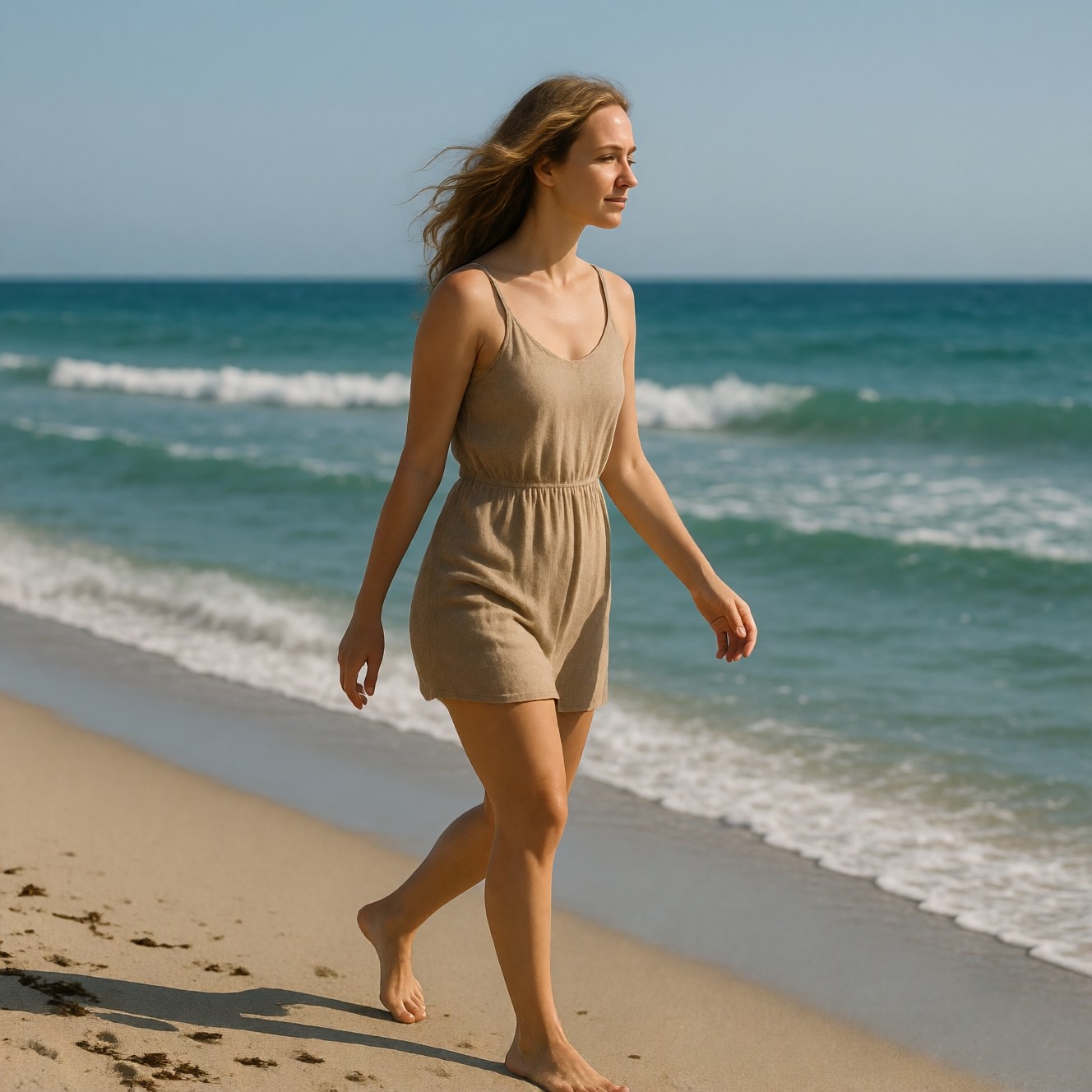}
    \end{subfigure}
    \begin{subfigure}[b]{0.28\textwidth}
        \includegraphics[width=\linewidth]{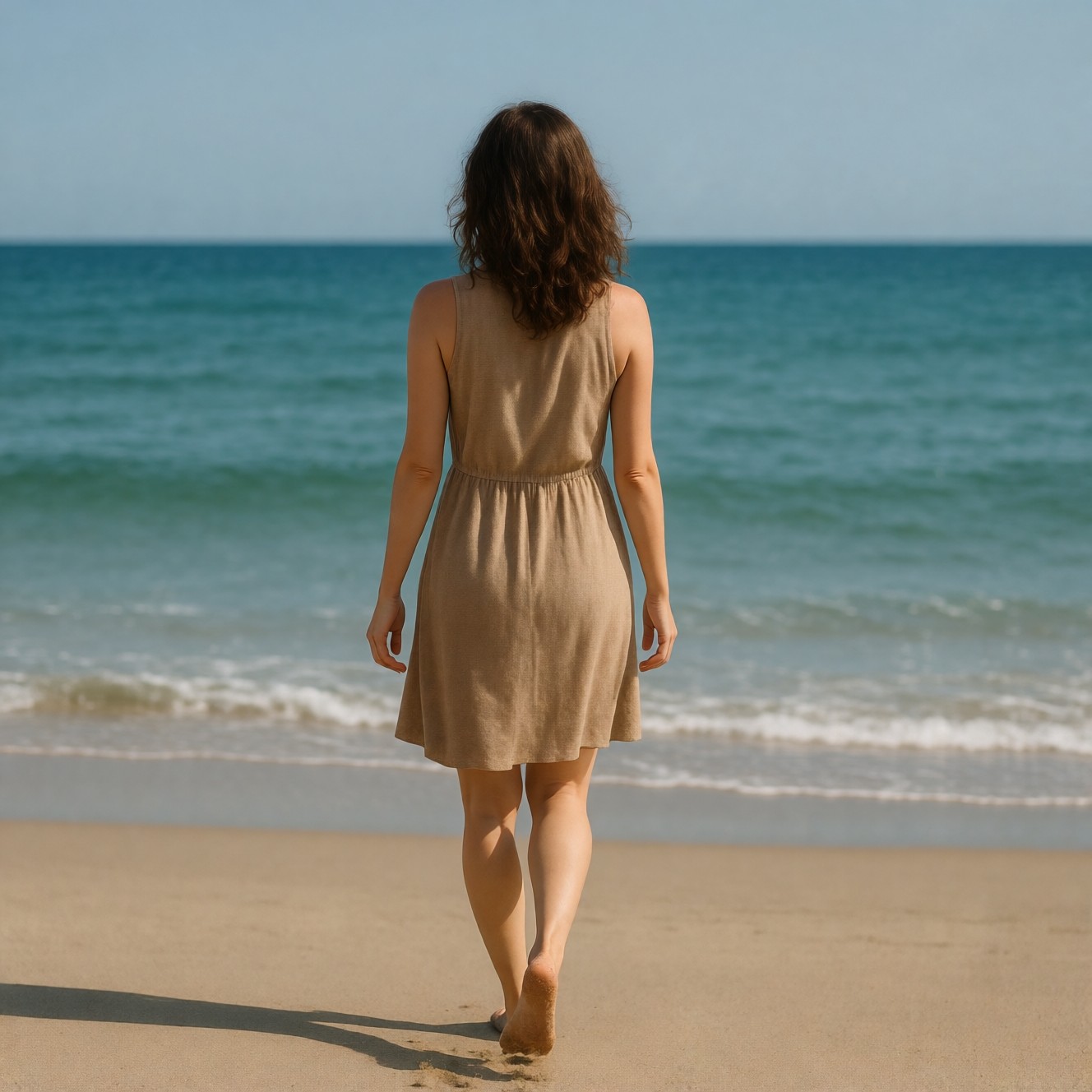}
    \end{subfigure}
    \begin{subfigure}[b]{0.28\textwidth}
        \includegraphics[width=\linewidth]{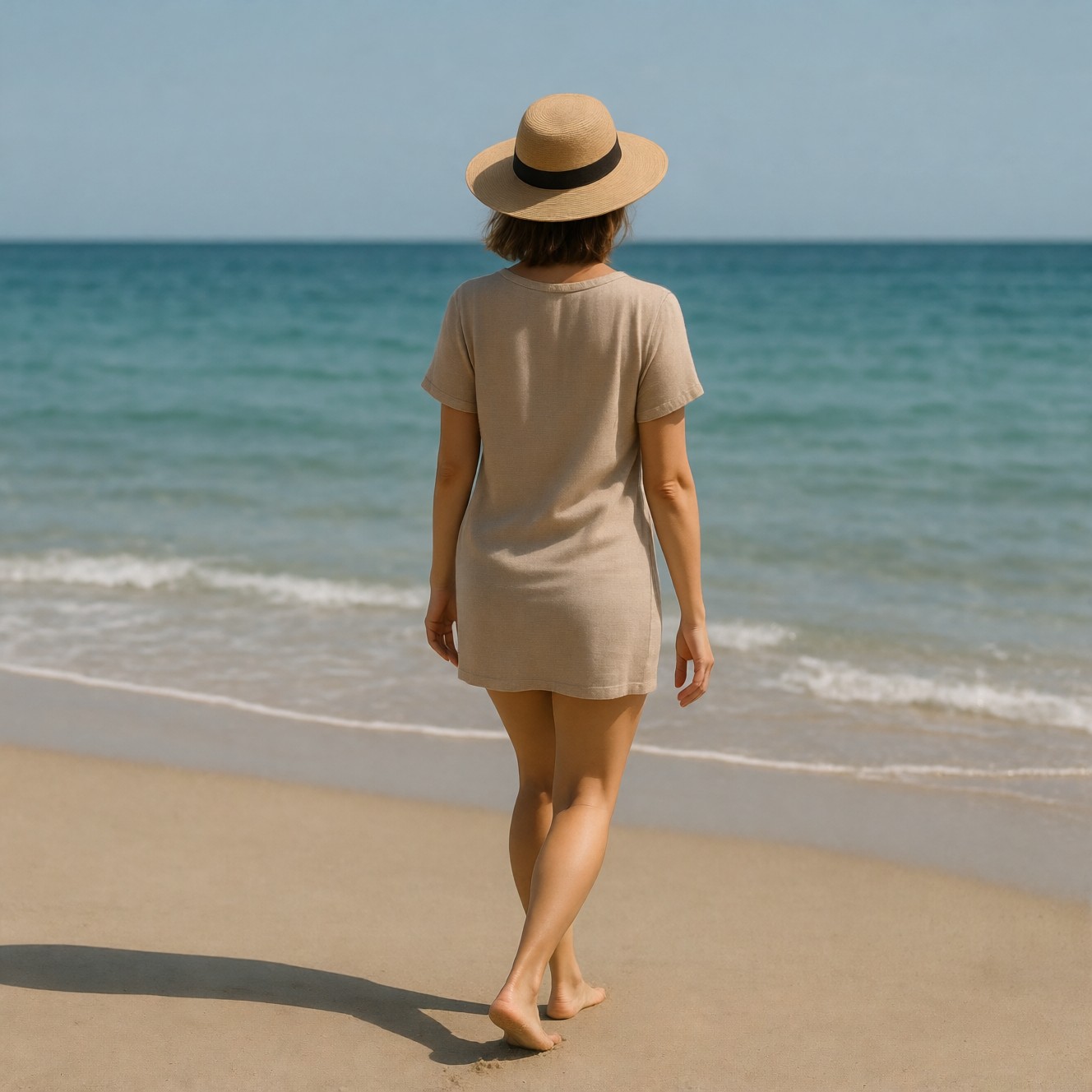}
    \end{subfigure}
   
    \caption{Qualitative comparison between \textbf{Qwen-Image} (first row) and \textbf{Qwen-Image-DCW} (second row) using \textbf{10 steps}, where the prompt is ``A woman is walking on the beach by the sea".}
    \label{fig:qwen_show1}
\end{figure*}

% \begin{table}[t]
% \centering
% \setlength{\tabcolsep}{3.3pt}
% \caption{Settings on CIFAR-10 using EDM and PFGM++.}
% \begin{tabular}{lccccccc}
% % {@{\hspace{0.62em}}l@{\hspace{0.62em}}c@{\hspace{0.62em}}c@{\hspace{0.62em}}c@{\hspace{0.62em}}c@{\hspace{0.62em}}c@{\hspace{0.62em}}c@{\hspace{0.62em}}c@{\hspace{0.62em}}}
% \toprule
% &  & \multicolumn{3}{c}{EDM} & \multicolumn{3}{c}{PFGM++} \\
% \cmidrule(lr){3-5} \cmidrule(lr){6-8}
% $T'$ & $\la$ & 13 & 21 & 35 & 13 & 21 & 35 \\
% \midrule
% Base & $\la_l$ & \underline{0.022}& \underline{0.012} & \underline{0.006}& \underline{0.022}& \underline{0.012} & \underline{0.006} \\
% Base &$\la_h$ & \underline{0.002}& \underline{0.005} & \underline{0.0007}& \underline{0.007}& \underline{0.004} & \underline{0.002}\\
% \midrule
% Ba-ES & $\la_l$ & \underline{0.008}  & \underline{0.005} & \underline{0.002} & \underline{0.016} & \underline{0.005} & \underline{0.002}\\
% Ba-ES &$\la_h$ & \underline{0.000} & \underline{0.000} & \underline{0.000} & \underline{0.000} & \underline{0.000} & \underline{0.000}\\
% \midrule
% Ba-FR &$\la_l$ & \underline{0.00}& \underline{0.000} & \underline{0.000} & \underline{0.000} & \underline{0.000} & \underline{0.000}\\
% Ba-FR &$\la_h$ & \underline{0.03}& \underline{0.024} & \underline{0.005} & \underline{0.018} & \underline{0.047} & \underline{0.006}\\
% \bottomrule
% \end{tabular}
% \label{tab:edm_pa}
% \end{table}

\begin{table}[t]
\centering
\setlength{\tabcolsep}{4.3pt}
\caption{FID and Recall (Rec)  on DiT.}
\label{tab8:dit}
\begin{tabular}{lccccccc}
\toprule
& & \multicolumn{2}{c}{$T=20$} & \multicolumn{2}{c}{$T=50$} \\
\cmidrule(lr){3-4} \cmidrule(lr){5-6}
Model & Dataset & FID$\downarrow$ & Rec$\uparrow$ & FID$\downarrow$ & Rec$\uparrow$ \\  % 插入 dataset 列
\midrule
DiT & ImageNet 256& 12.83 & 0.54 & 3.78 & 0.58 \\
DiT-ES & ImageNet 256& 10.00 & - & 3.30
 & - \\
\textbf{DiT+Ours} & \textbf{ImageNet} 256 & \textbf{7.99} & \textbf{0.51} & \textbf{3.09} & \textbf{0.56} \\
\bottomrule
\end{tabular}

\end{table}

Given the extensive influence of transformer-based diffusion models, we select DiT~\cite{peebles2023scalable} as the baseline model, ADM-ES~\cite{ningelucidating} as the comparative model. Subsequently, we adopt Fr\'echet Inception Distance (FID)~\cite{heusel2017gans} and  Recall~\cite{heusel2017gans} as evaluation metrics, and select ImageNet $256\times256$ as the test dataset for our experiments. 

Tab.~\ref{tab8:dit} clearly demonstrates that our method achieves a comprehensive reduction in the FID scores of DiT and outperforms the comparative models significantly. In the subsequent appendix, we also provide the evaluation results of two text-to-image models, which are also based on the DiT architecture.

% Our method has achieved widespread success across various UNet-based architectures. 

\begin{figure*}[t] 
    \centering
    \begin{subfigure}[b]{0.28\textwidth} 
        \includegraphics[width=\linewidth]{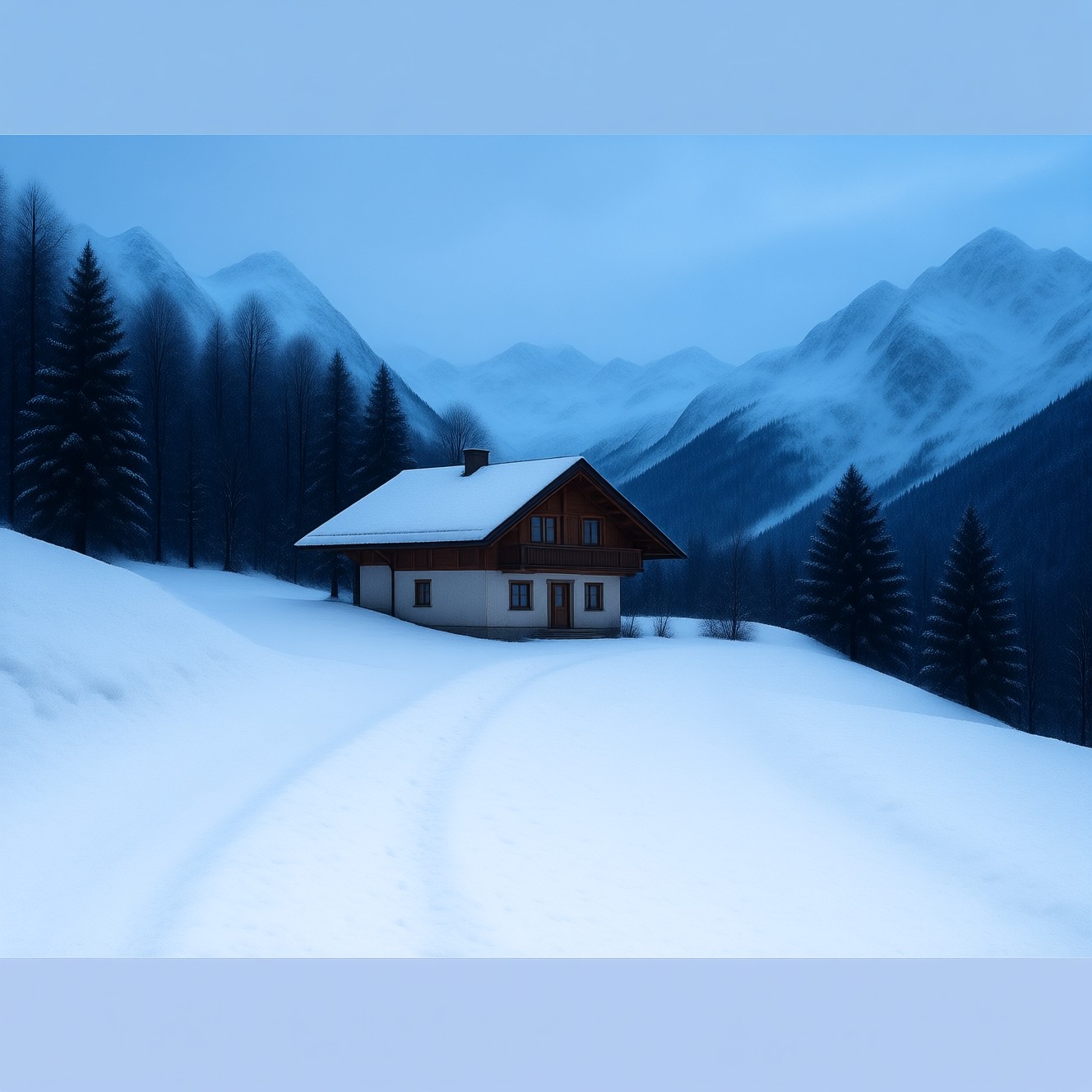}
    \end{subfigure}
    \begin{subfigure}[b]{0.28\textwidth} 
        \includegraphics[width=\linewidth]{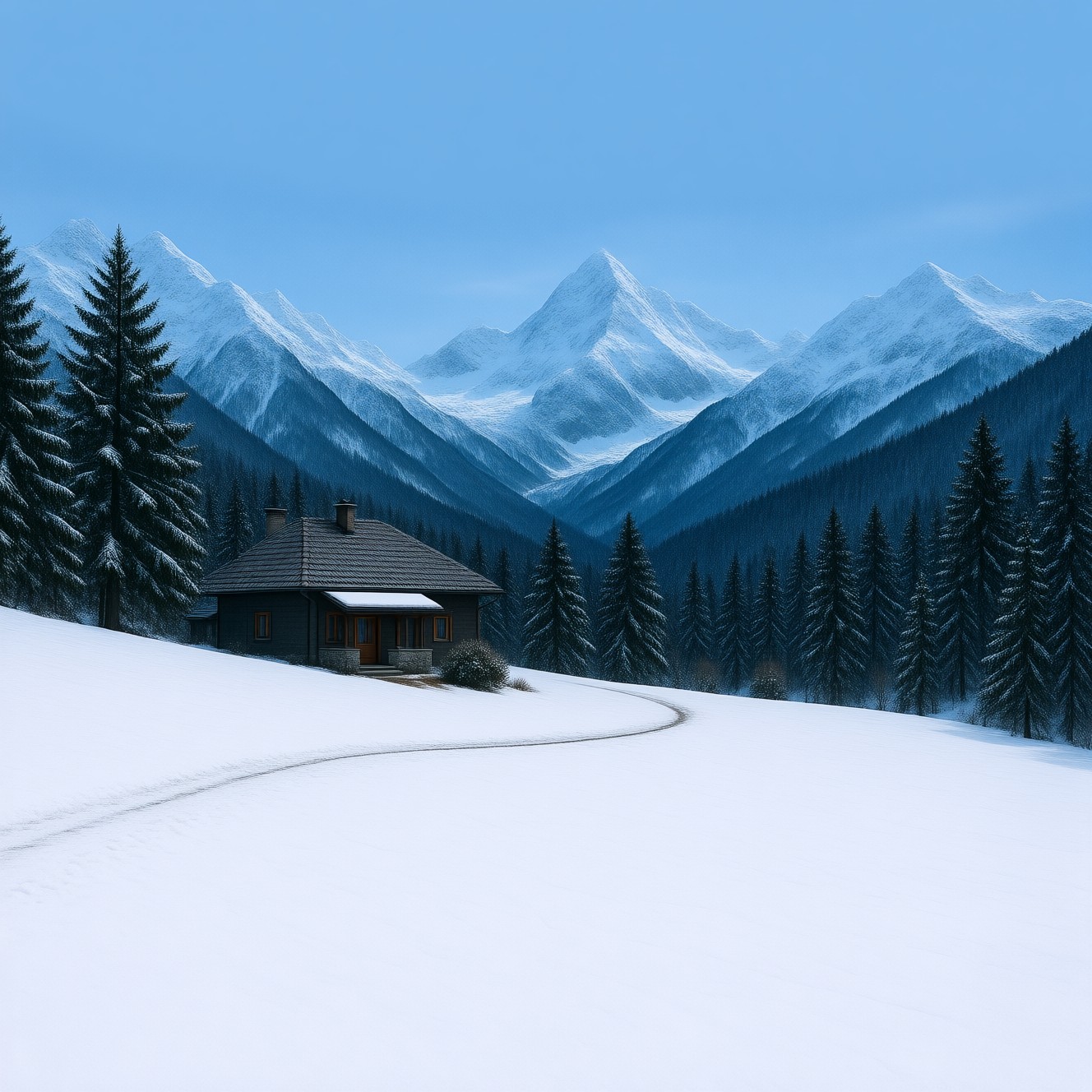}
    \end{subfigure}
    \begin{subfigure}[b]{0.28\textwidth} 
        \includegraphics[width=\linewidth]{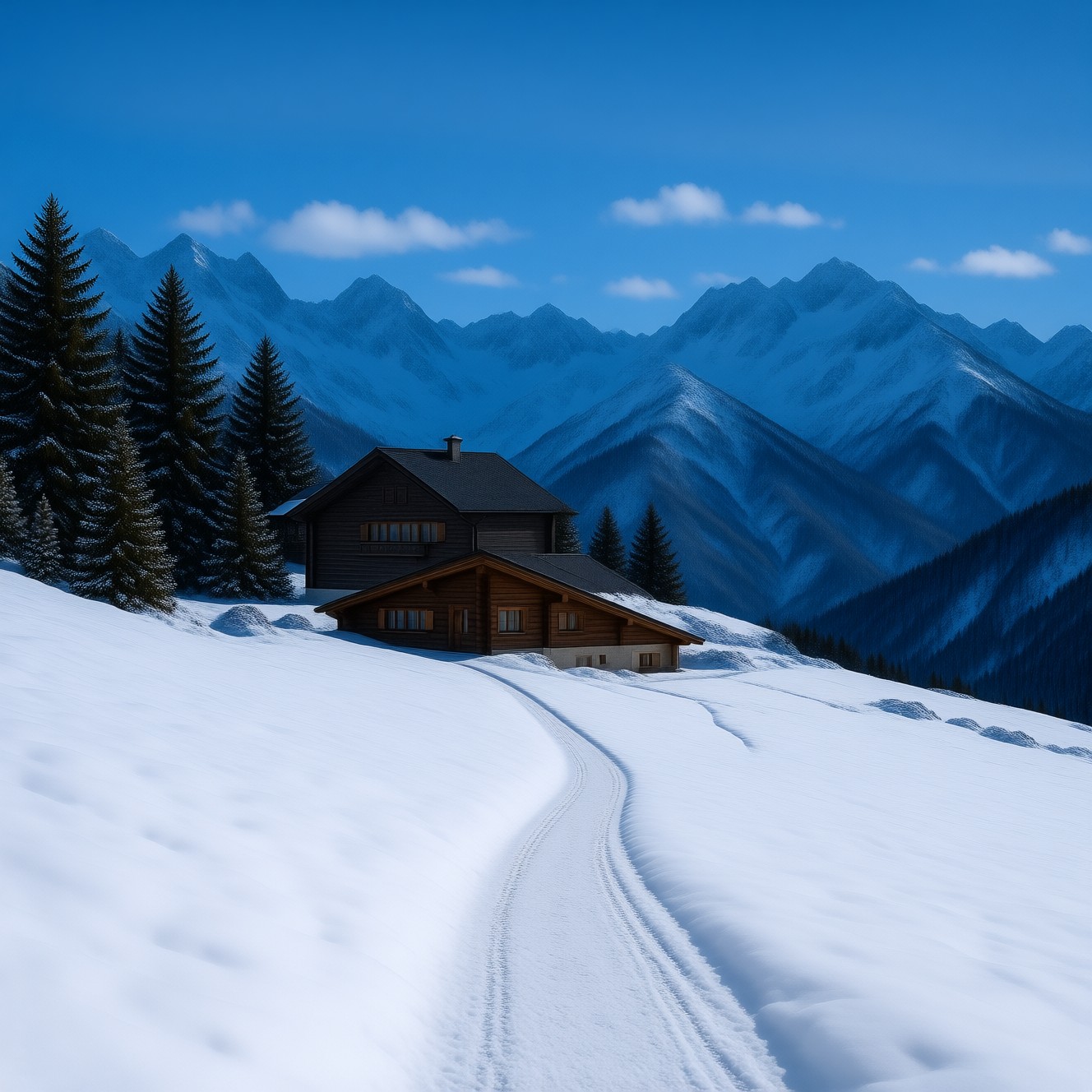}
    \end{subfigure}
    \begin{subfigure}[b]{0.28\textwidth}
        \includegraphics[width=\linewidth]{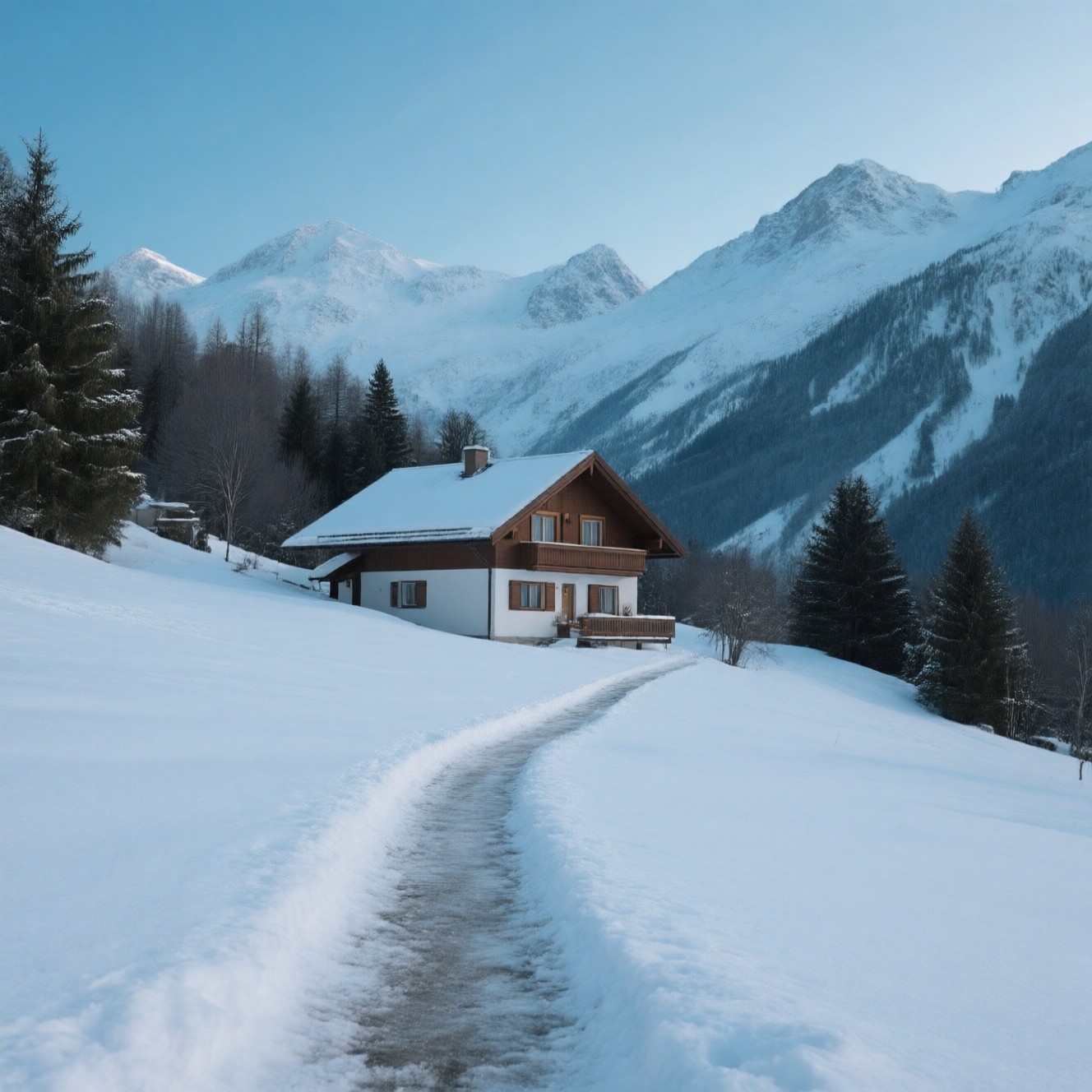}
    \end{subfigure}
    \begin{subfigure}[b]{0.28\textwidth}
        \includegraphics[width=\linewidth]{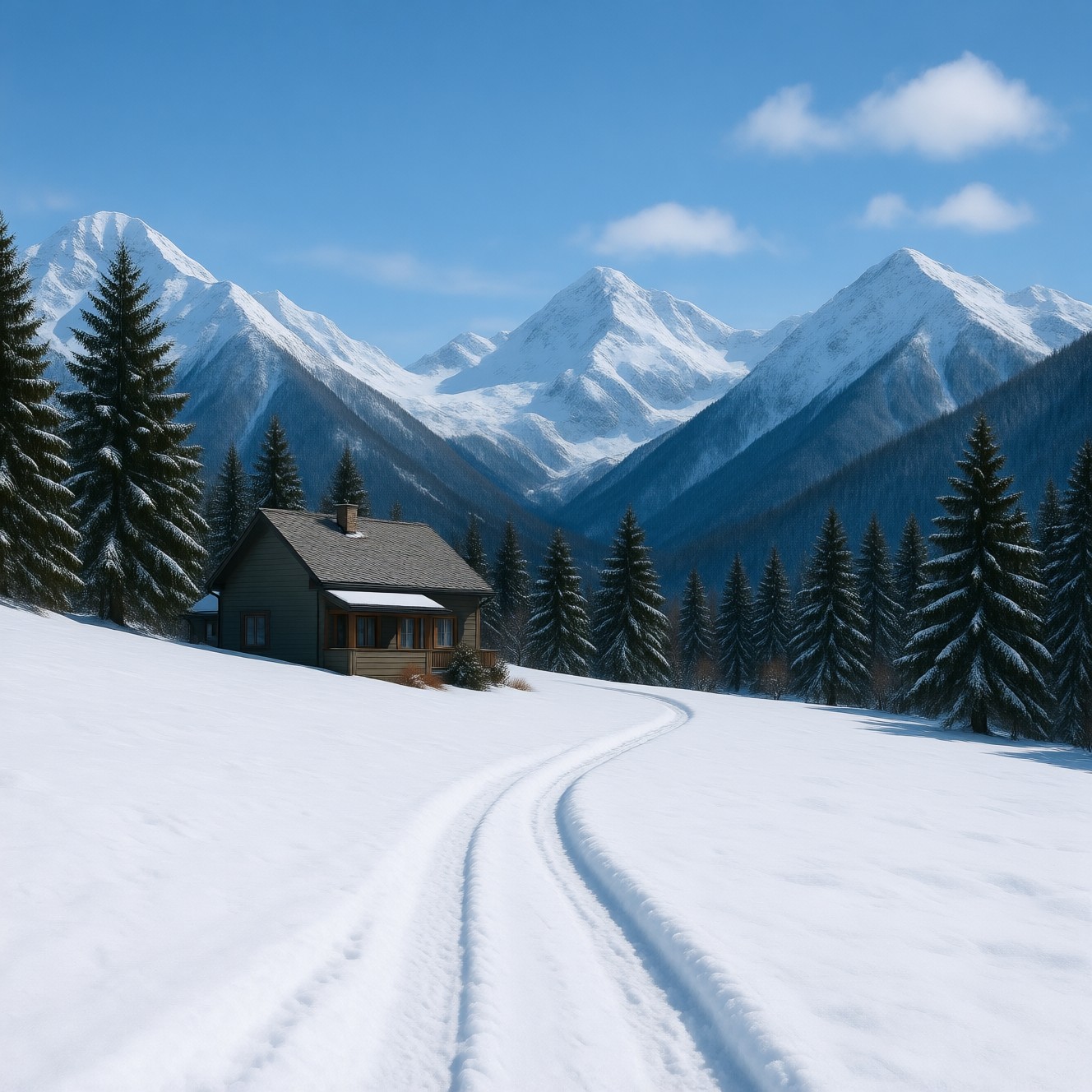}
    \end{subfigure}
    \begin{subfigure}[b]{0.28\textwidth}
        \includegraphics[width=\linewidth]{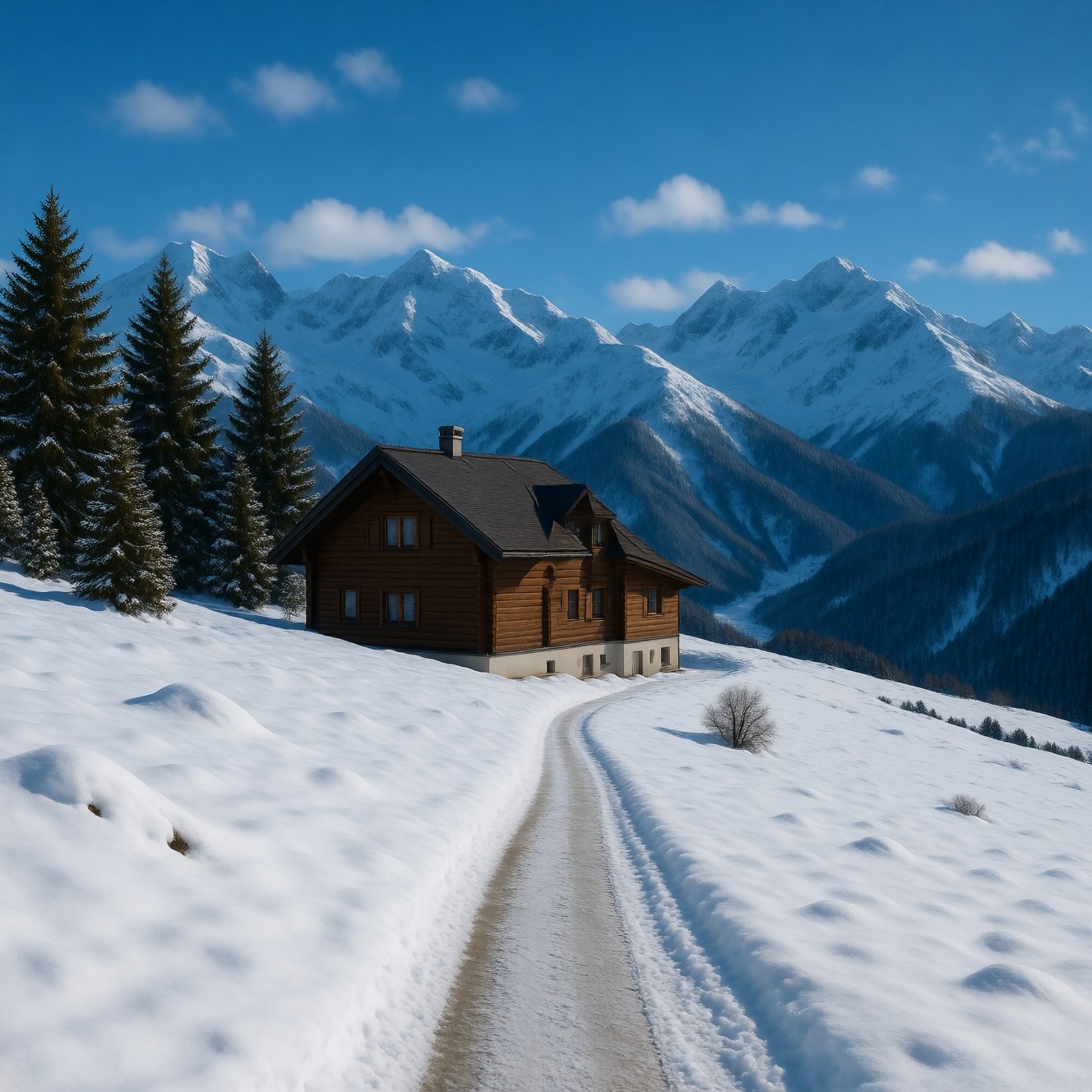}
    \end{subfigure}
    % \begin{subfigure}[b]{0.15\textwidth}
    %     \includegraphics[width=\linewidth]{figures/show/d1.jpg}
    % \end{subfigure}
    \caption{Qualitative comparison between \textbf{Qwen-Image} (first row) and \textbf{Qwen-Image-DCW} (second row) using \textbf{10 steps}, where the prompt is ``There is a house and a path on a snowy mountain".}
    \label{fig:qwen_show2}
\end{figure*}

\begin{figure*}[h] 
    \centering
    \begin{subfigure}[b]{0.28\textwidth} 
        \includegraphics[width=\linewidth]{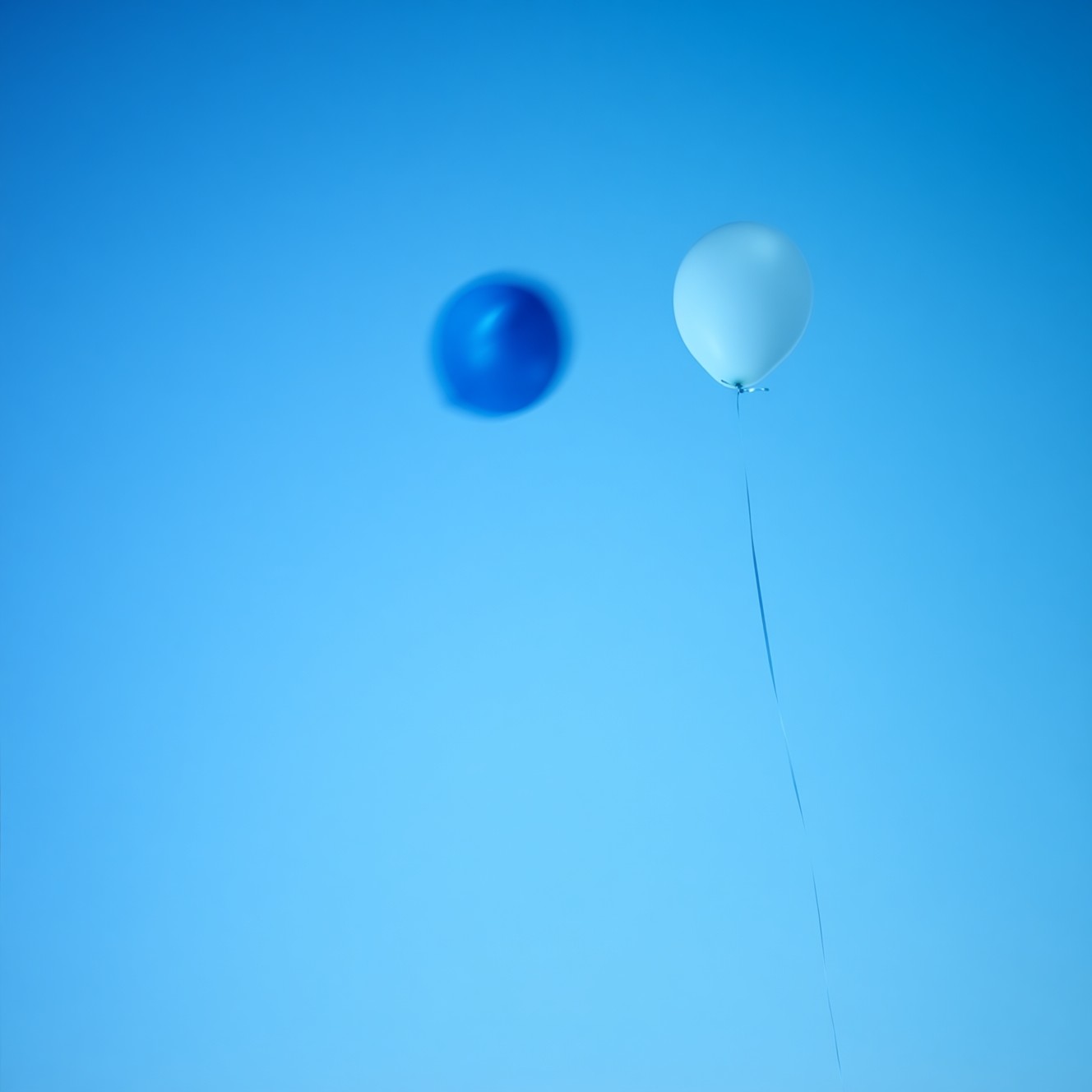}
    \end{subfigure}
    \begin{subfigure}[b]{0.28\textwidth} 
        \includegraphics[width=\linewidth]{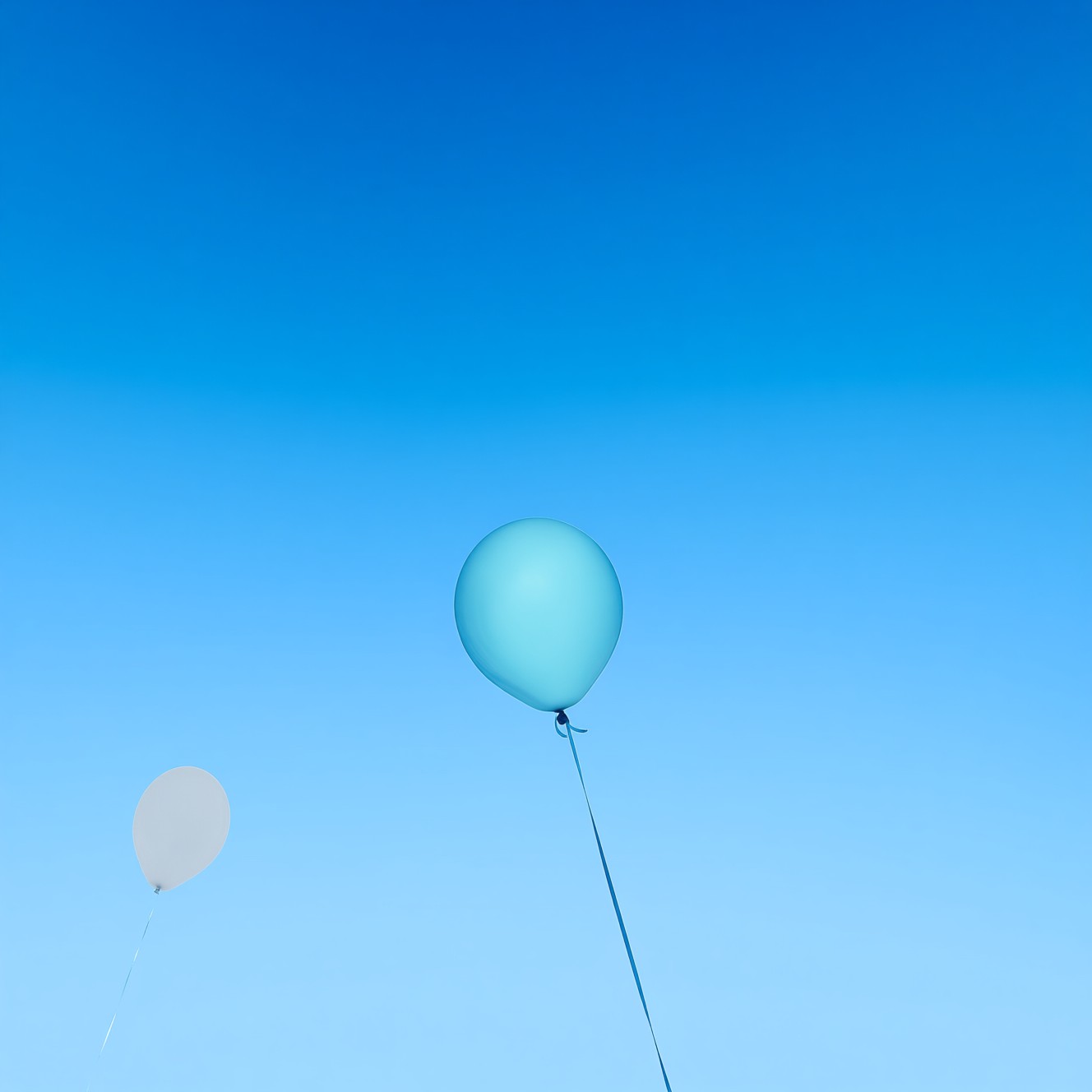}
    \end{subfigure}
    \begin{subfigure}[b]{0.28\textwidth} 
        \includegraphics[width=\linewidth]{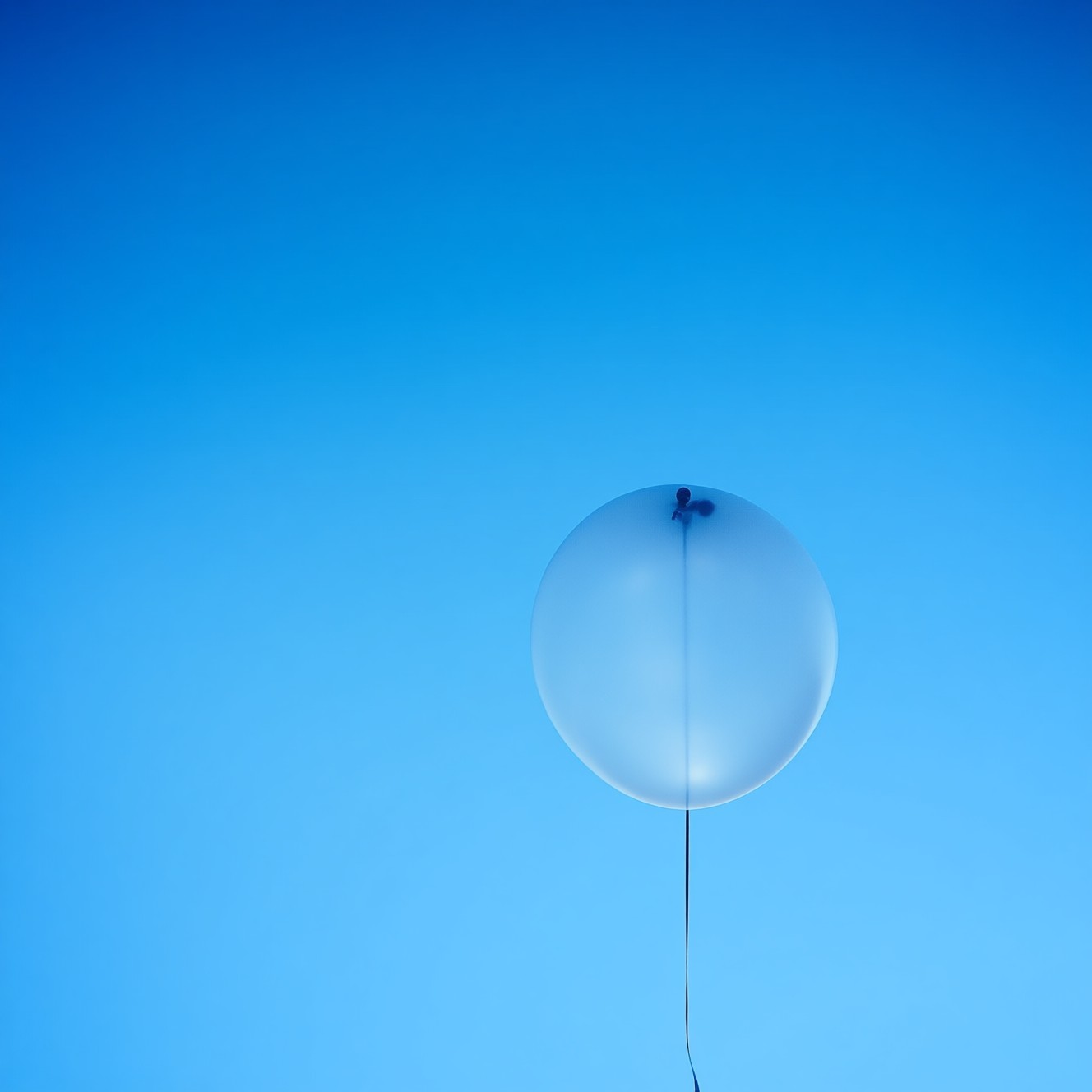}
    \end{subfigure}
    \begin{subfigure}[b]{0.28\textwidth}
        \includegraphics[width=\linewidth]{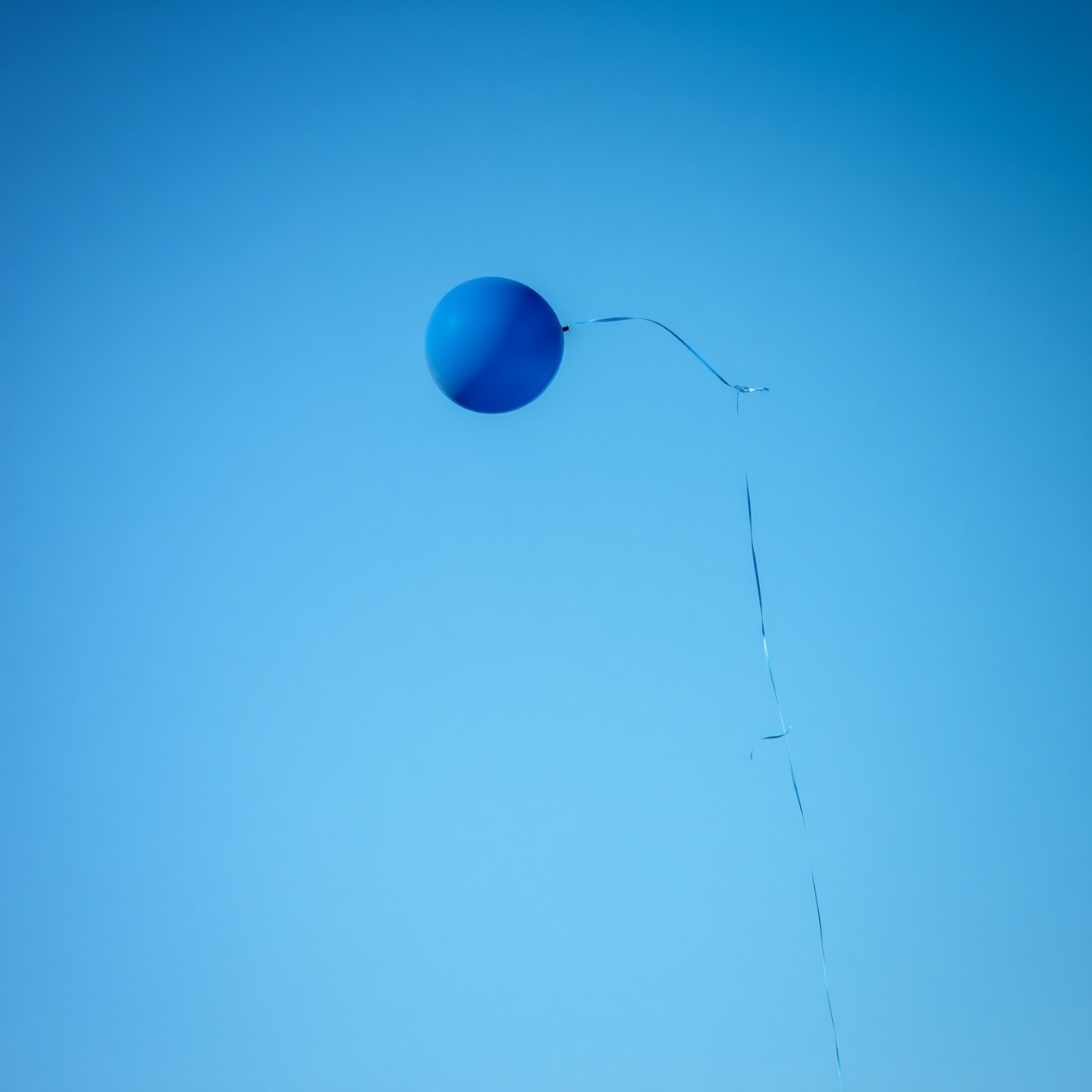}
    \end{subfigure}
    \begin{subfigure}[b]{0.28\textwidth}
        \includegraphics[width=\linewidth]{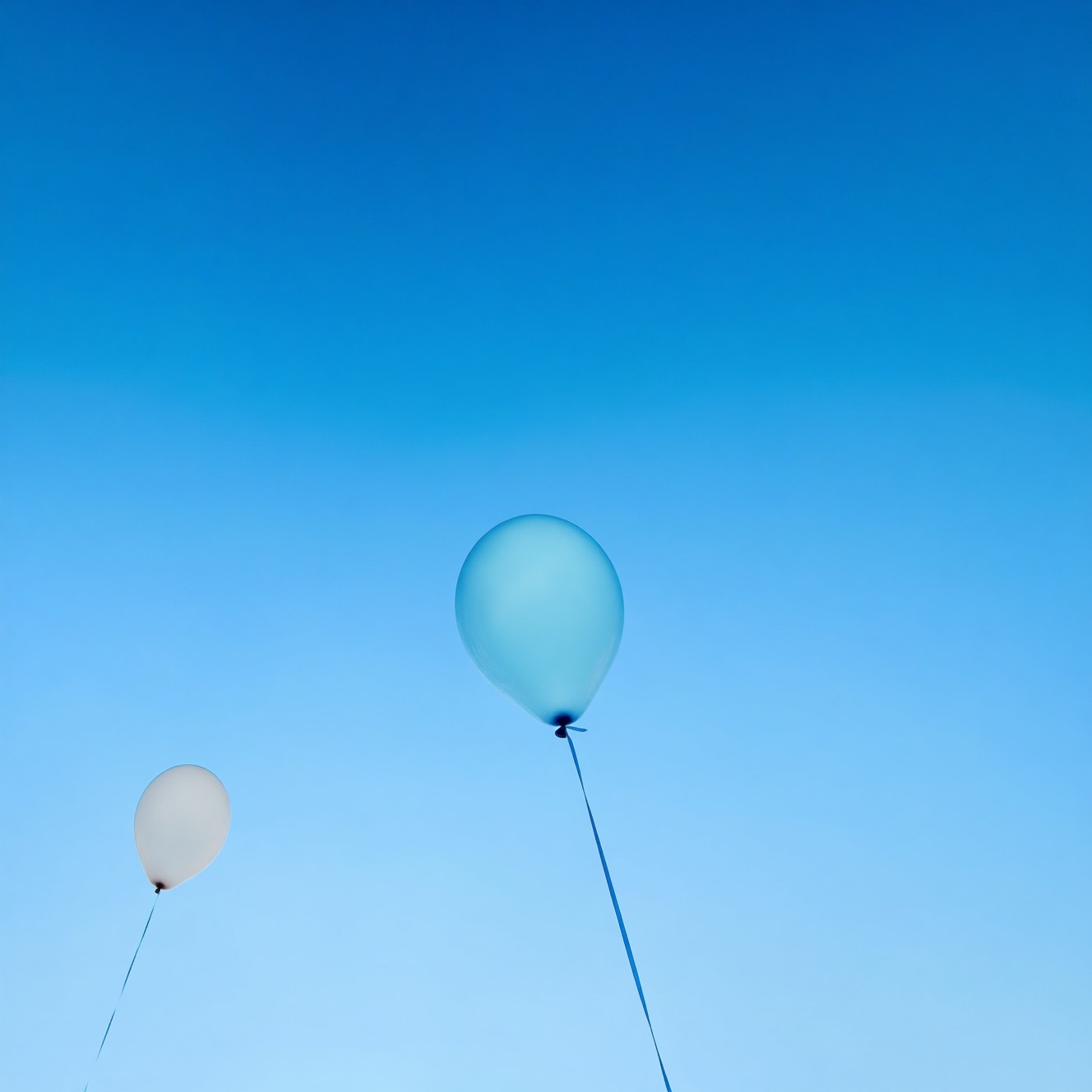}
    \end{subfigure}
    \begin{subfigure}[b]{0.28\textwidth}
        \includegraphics[width=\linewidth]{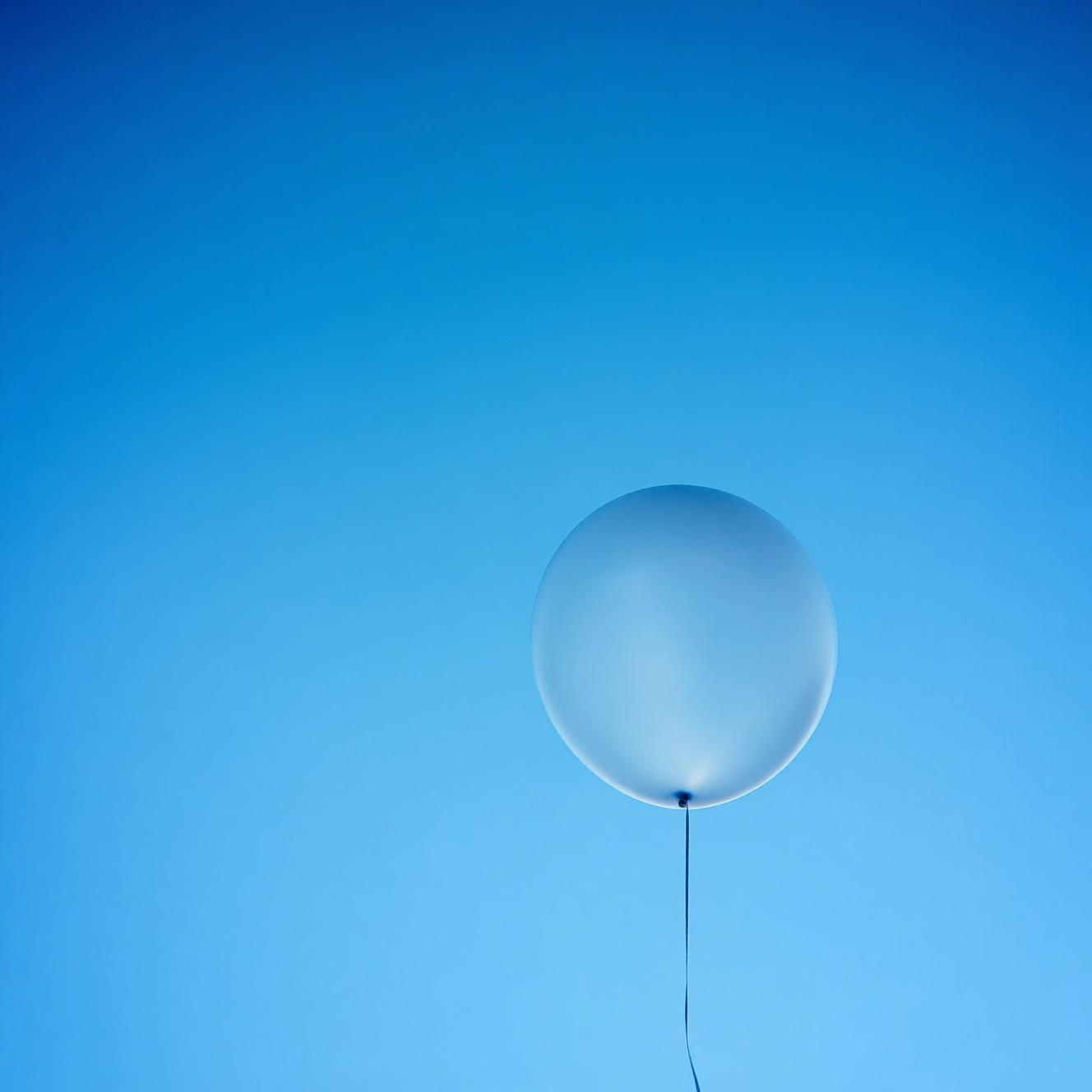}
    \end{subfigure}
    \caption{Qualitative comparison between \textbf{Qwen-Image} (first row) and \textbf{Qwen-Image-DCW} (second row) using \textbf{10 steps}, where the prompt is ``\textbf{A balloon} gently climbs into a serene blue sky".}
    \label{fig:qwen_show3}
\end{figure*}

\begin{figure*}[t] 
    \centering
    \begin{subfigure}[b]{0.28\textwidth} 
        \includegraphics[width=\linewidth]{figures/show/flux/b1.jpg}
    \end{subfigure}
    \begin{subfigure}[b]{0.28\textwidth} 
        \includegraphics[width=\linewidth]{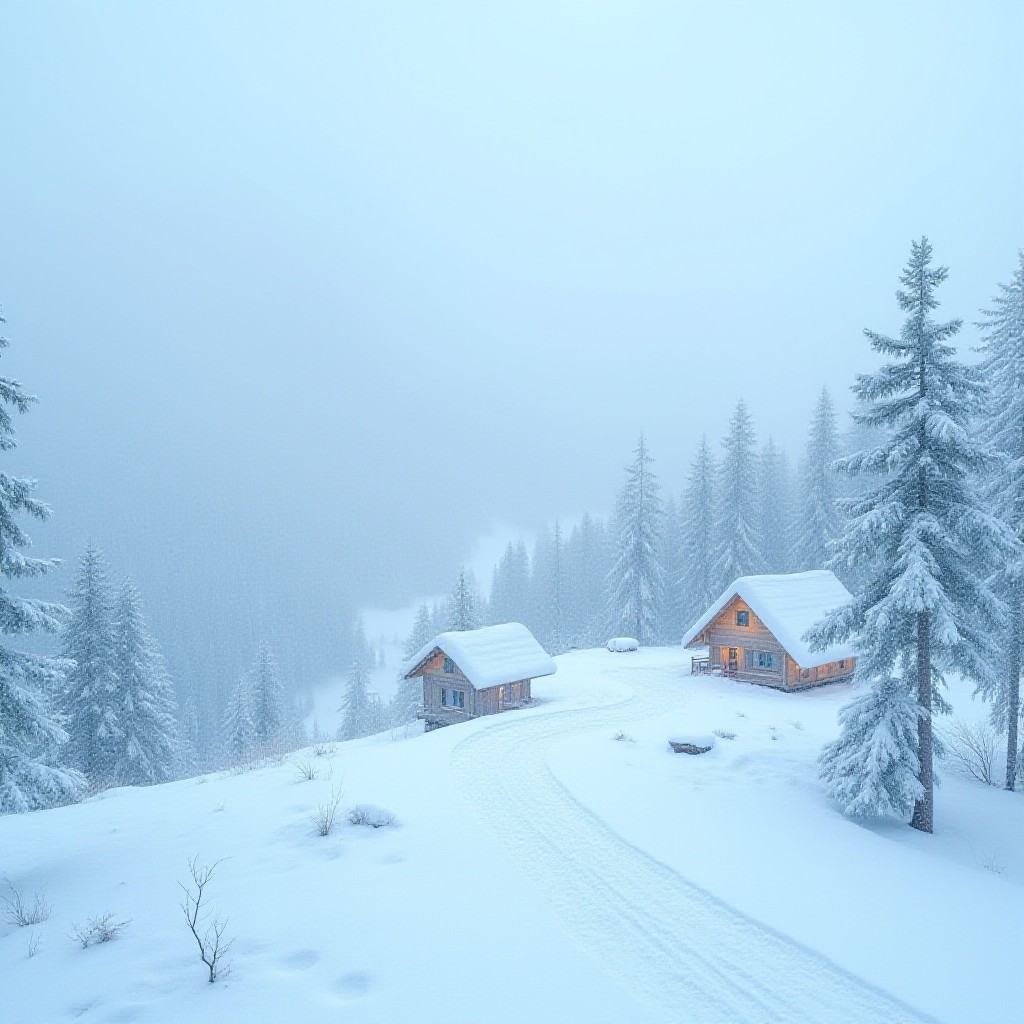}
    \end{subfigure}
    \begin{subfigure}[b]{0.28\textwidth} 
        \includegraphics[width=\linewidth]{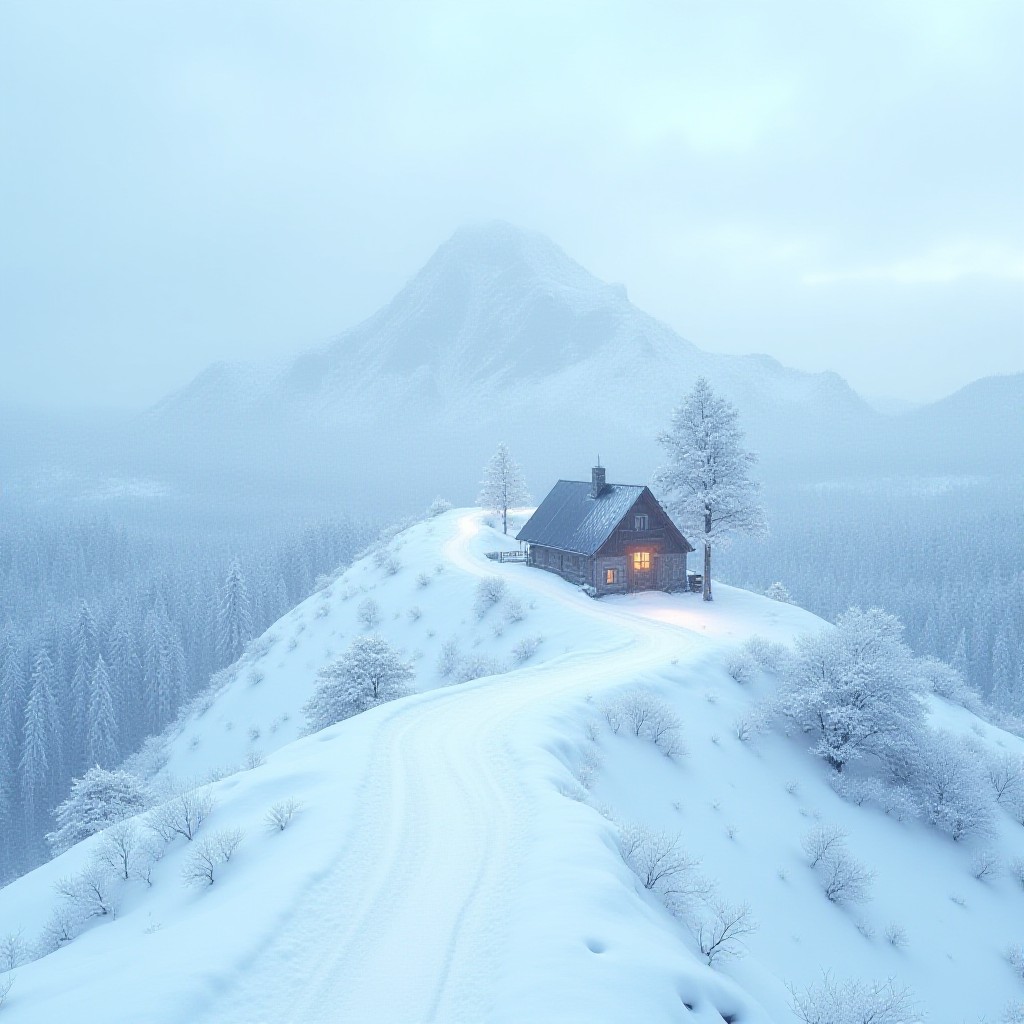}
    \end{subfigure}
    \begin{subfigure}[b]{0.28\textwidth}
        \includegraphics[width=\linewidth]{figures/show/flux/d1.jpg}
    \end{subfigure}
    \begin{subfigure}[b]{0.28\textwidth}
        \includegraphics[width=\linewidth]{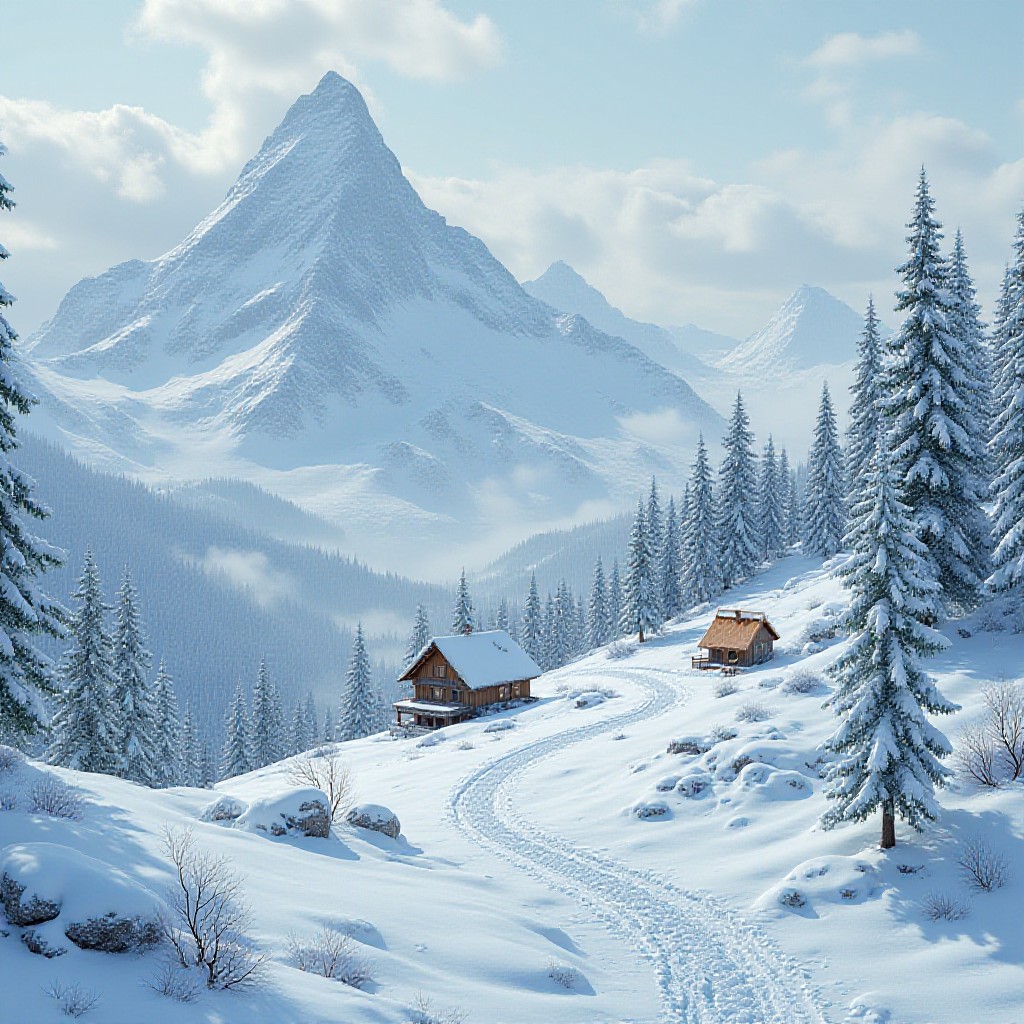}
    \end{subfigure}
    \begin{subfigure}[b]{0.28\textwidth}
        \includegraphics[width=\linewidth]{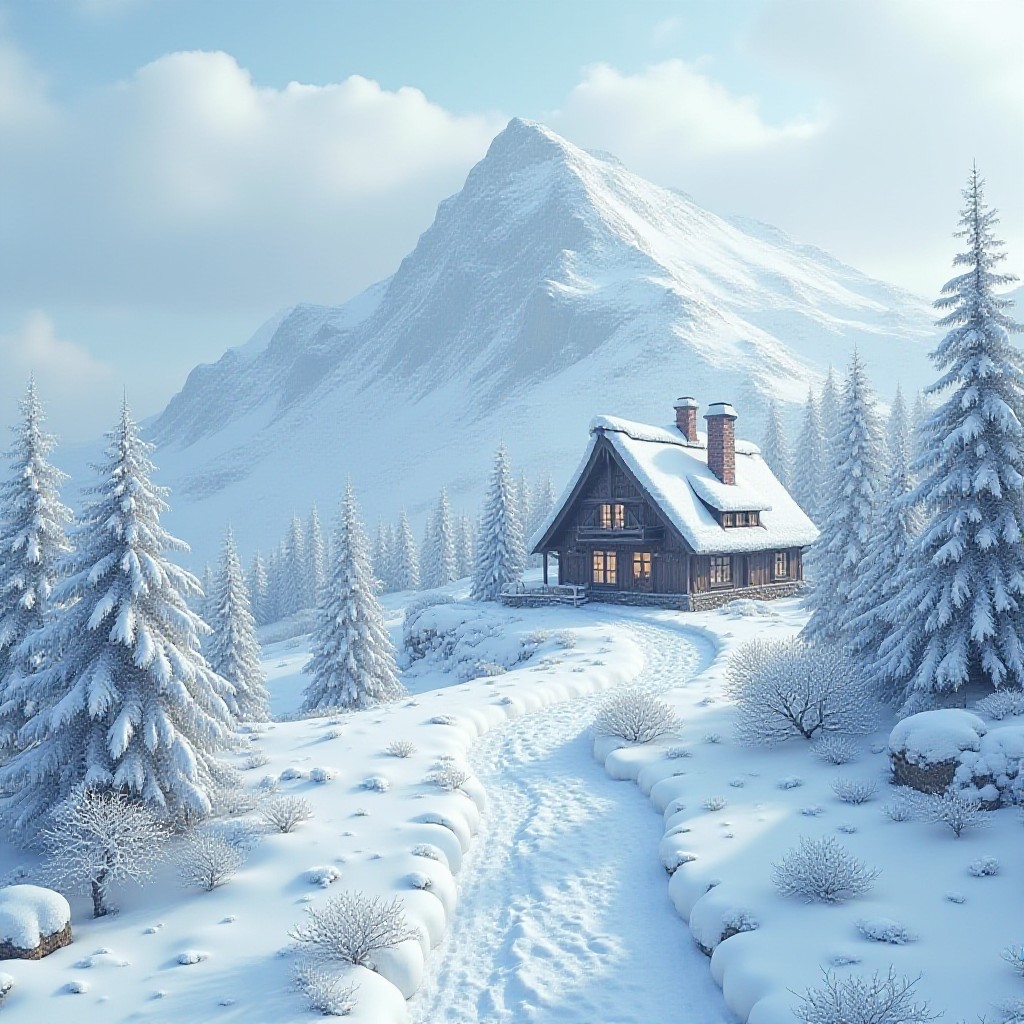}
    \end{subfigure}
    \caption{Qualitative comparison between \textbf{FLUX} (first row) and \textbf{FLUX-DCW} (second row) using \textbf{10 steps}, where the prompt is ``There is a house and a path on a snowy mountain".}
    \label{fig:flux_show4}
\end{figure*}

\begin{figure*}[t] 
    \centering
    \begin{subfigure}[b]{0.28\textwidth} 
        \includegraphics[width=\linewidth]{figures/show/flux/b4.jpg}
    \end{subfigure}
    \begin{subfigure}[b]{0.28\textwidth} 
        \includegraphics[width=\linewidth]{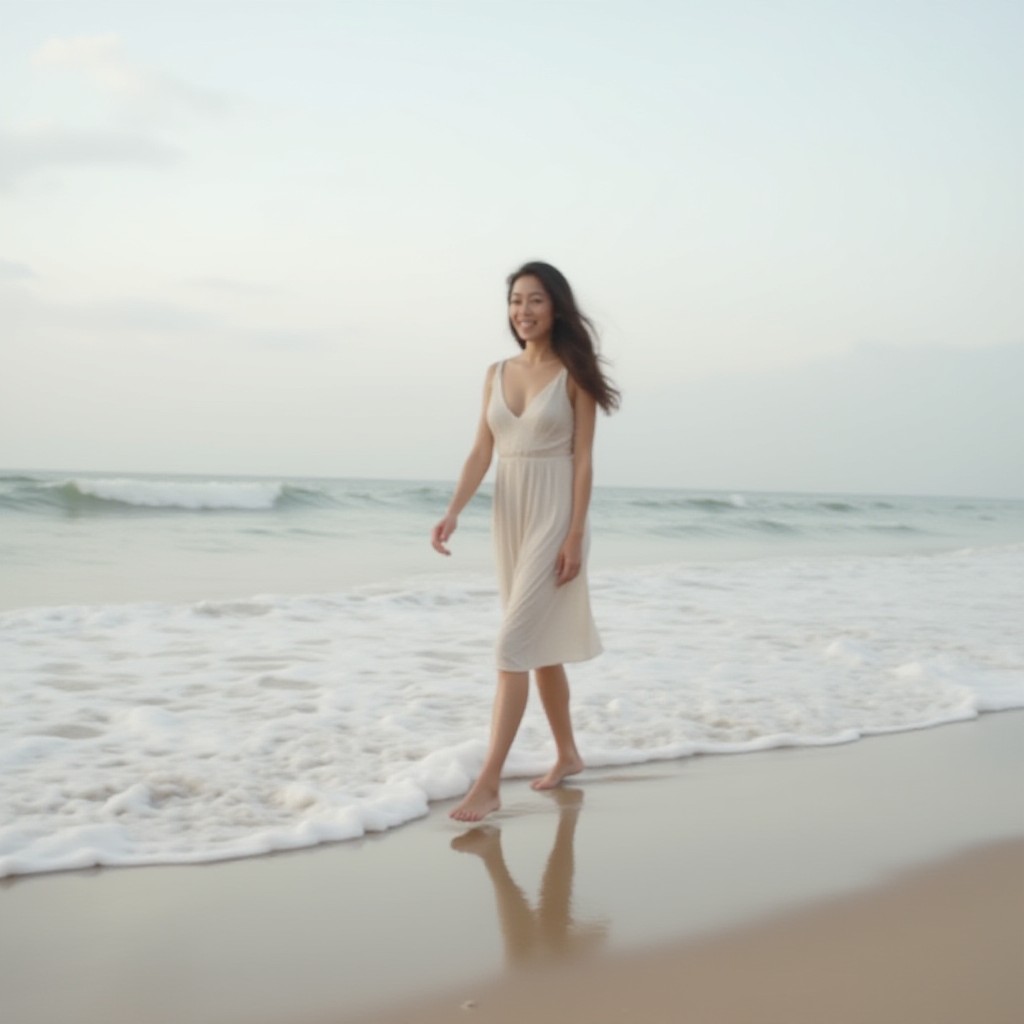}
    \end{subfigure}
    \begin{subfigure}[b]{0.28\textwidth} 
        \includegraphics[width=\linewidth]{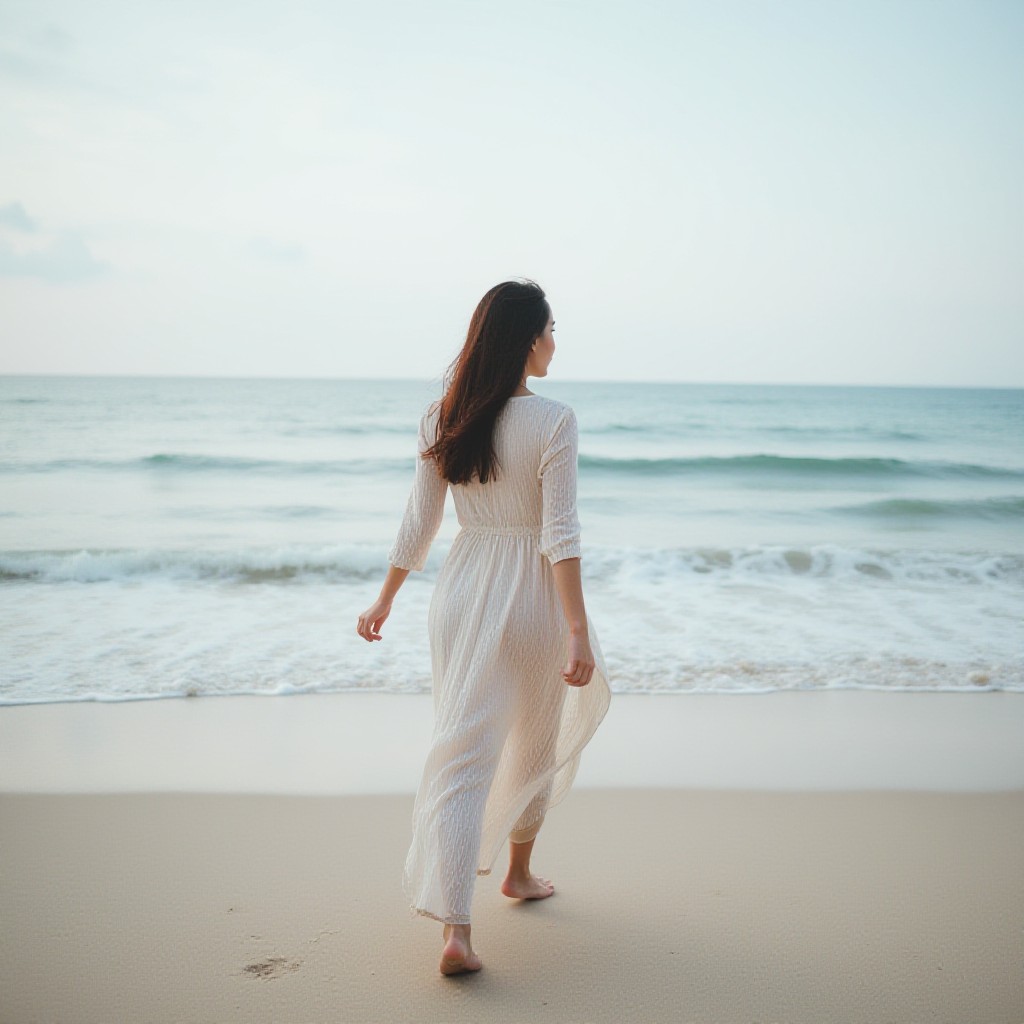}
    \end{subfigure}
    \begin{subfigure}[b]{0.28\textwidth}
        \includegraphics[width=\linewidth]{figures/show/flux/d4.jpg}
    \end{subfigure}
    \begin{subfigure}[b]{0.28\textwidth}
        \includegraphics[width=\linewidth]{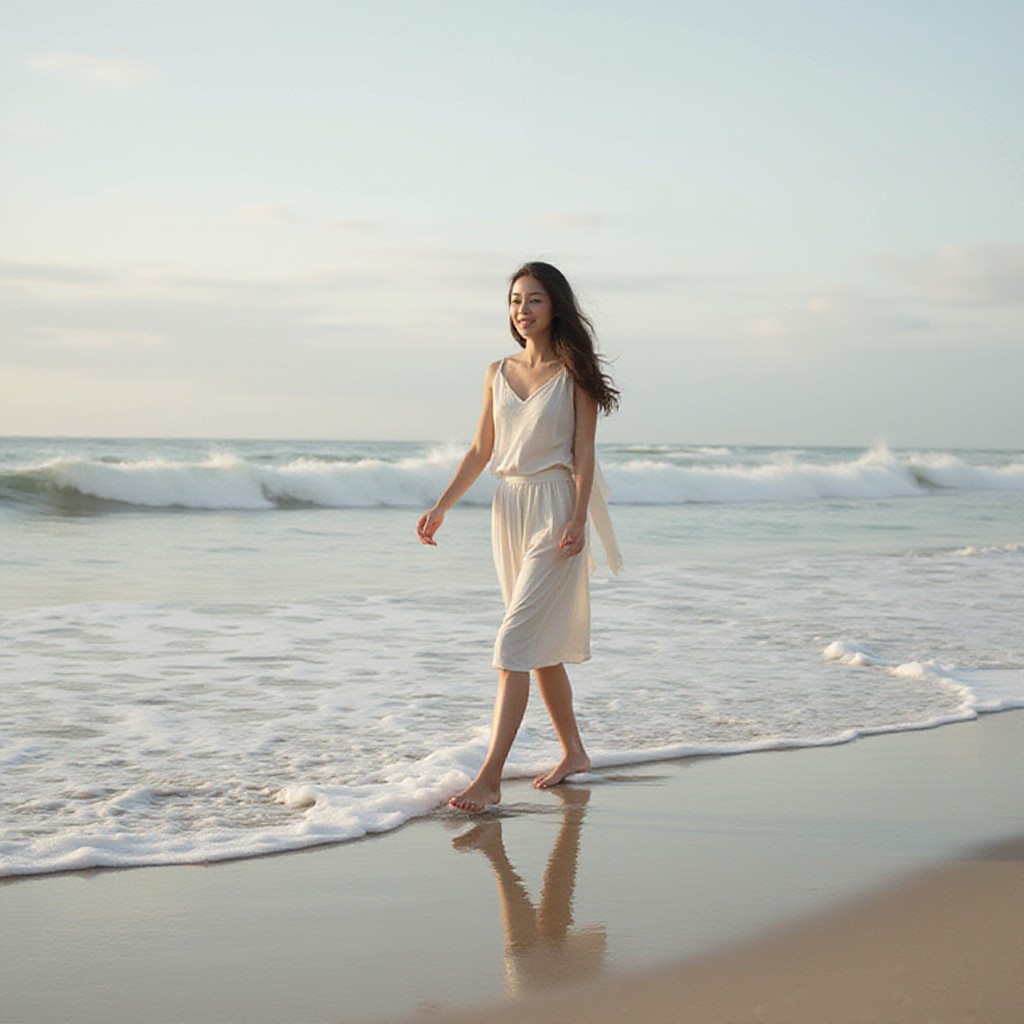}
    \end{subfigure}
    \begin{subfigure}[b]{0.28\textwidth}
        \includegraphics[width=\linewidth]{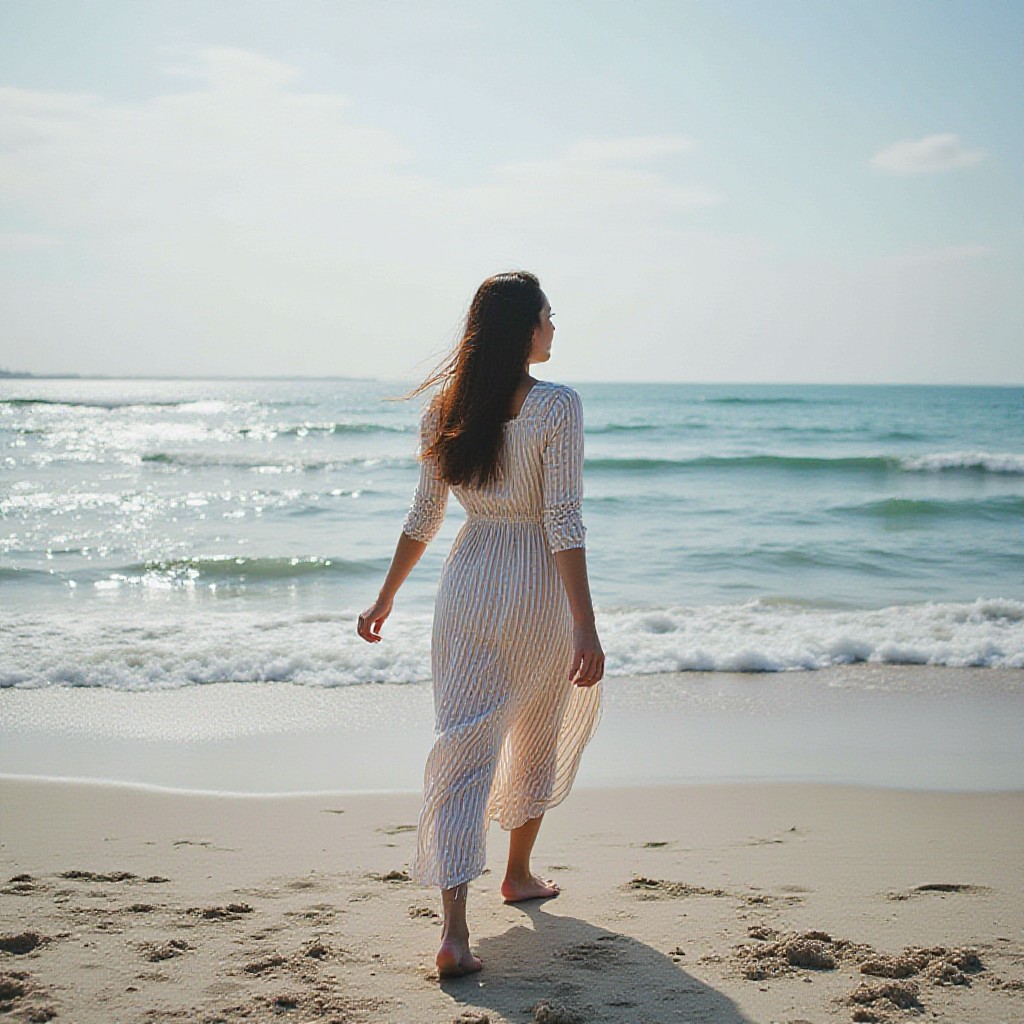}
    \end{subfigure}
    \caption{Qualitative comparison between \textbf{FLUX} (first row) and \textbf{FLUX-DCW} (second row) using \textbf{10 steps}, where the prompt is ``A woman is walking on the beach by the sea".}
    \label{fig:flux_show5}
\end{figure*}

\begin{figure*}[h] 
    \centering
    \begin{subfigure}[b]{0.28\textwidth} 
        \includegraphics[width=\linewidth]{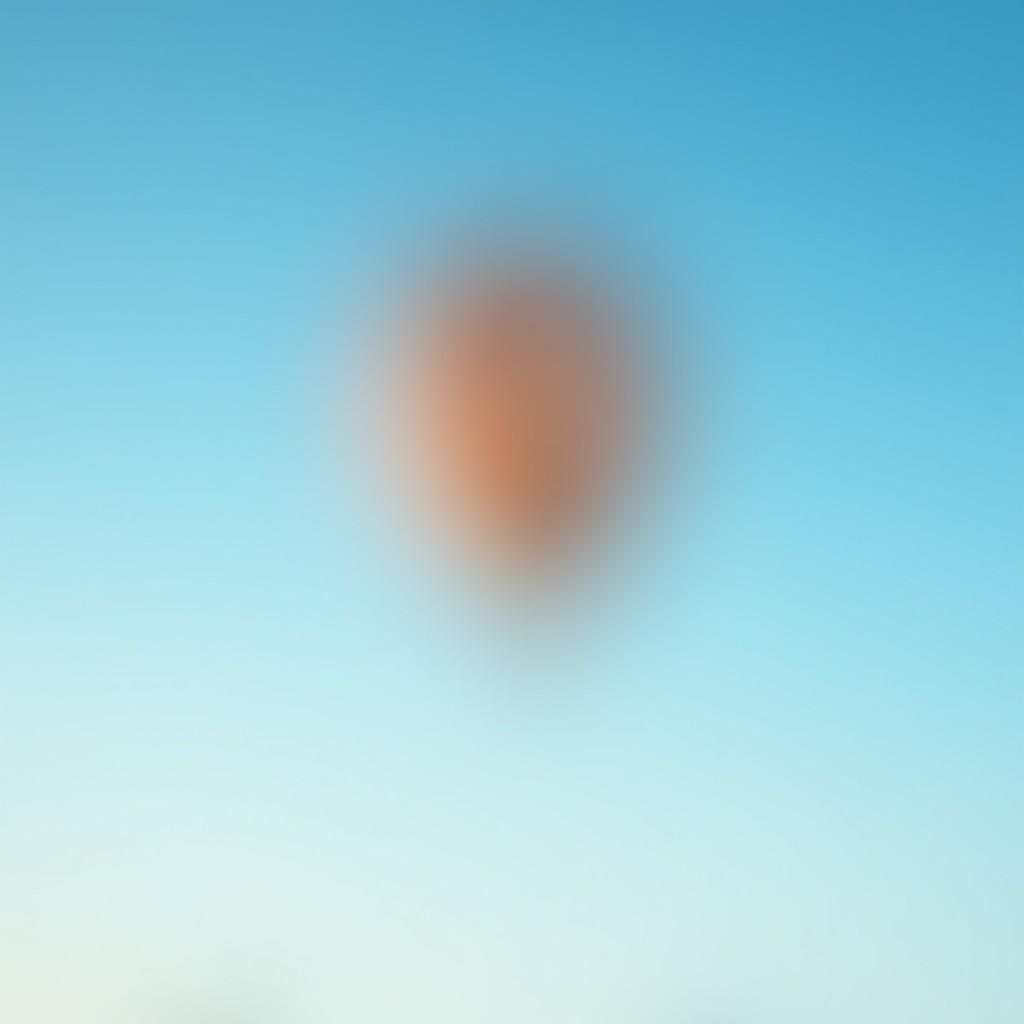}
    \end{subfigure}
    \begin{subfigure}[b]{0.28\textwidth} 
        \includegraphics[width=\linewidth]{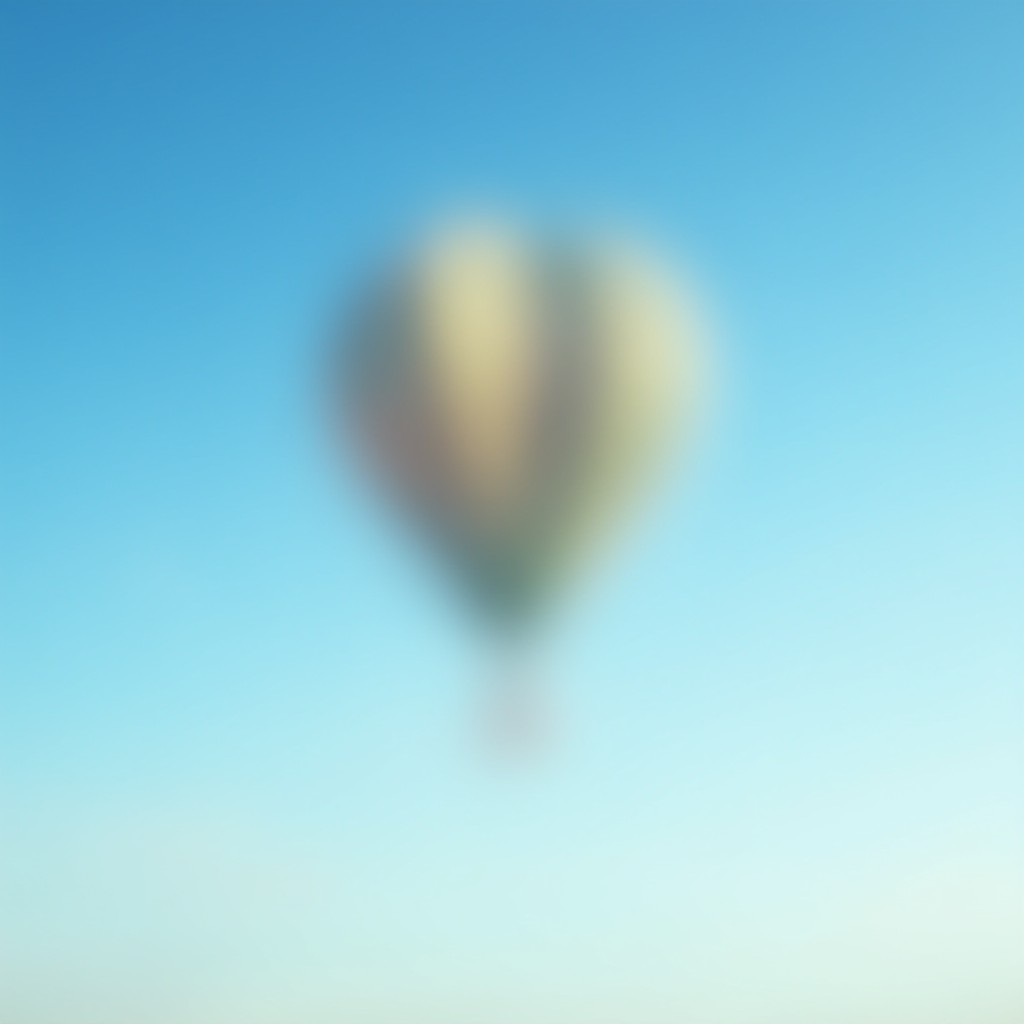}
    \end{subfigure}
    \begin{subfigure}[b]{0.28\textwidth} 
        \includegraphics[width=\linewidth]{figures/show/flux/b9.jpg}
    \end{subfigure}
    \begin{subfigure}[b]{0.28\textwidth}
        \includegraphics[width=\linewidth]{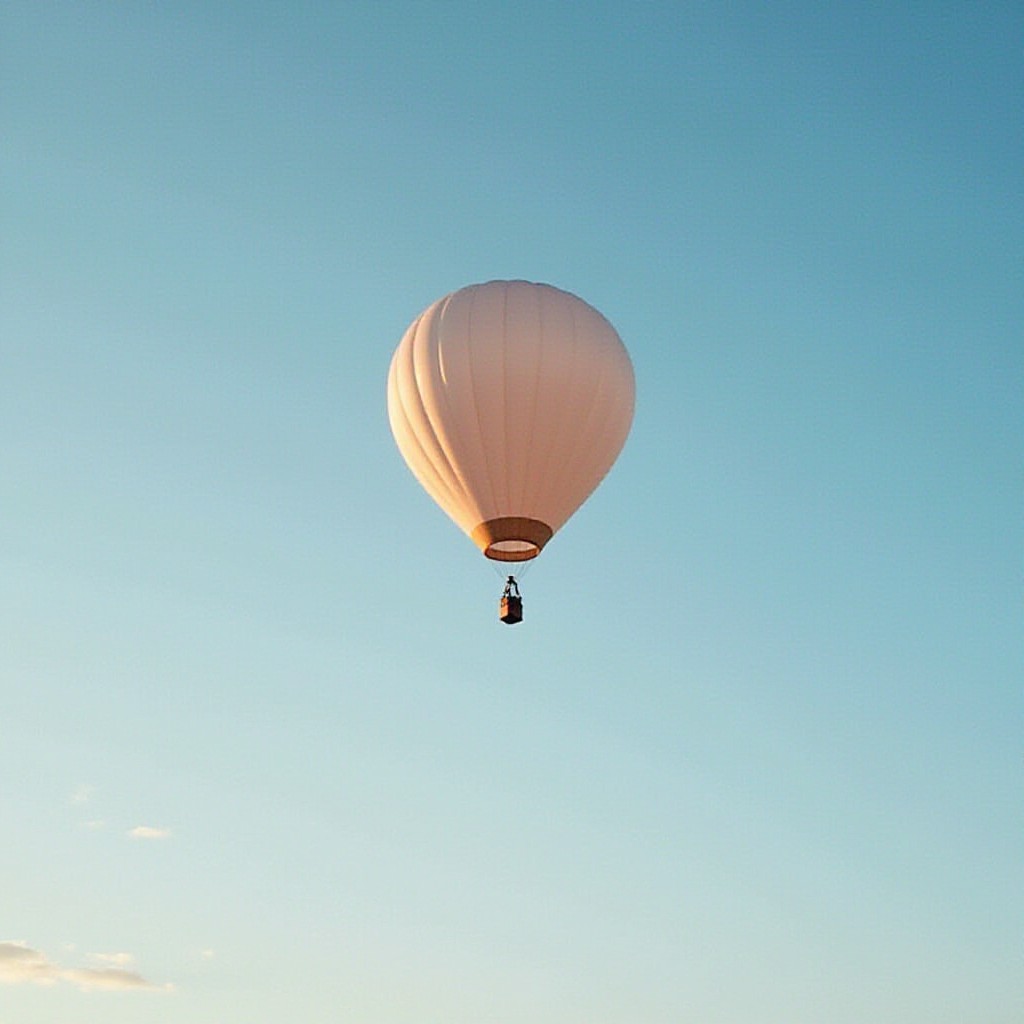}
    \end{subfigure}
    \begin{subfigure}[b]{0.28\textwidth}
        \includegraphics[width=\linewidth]{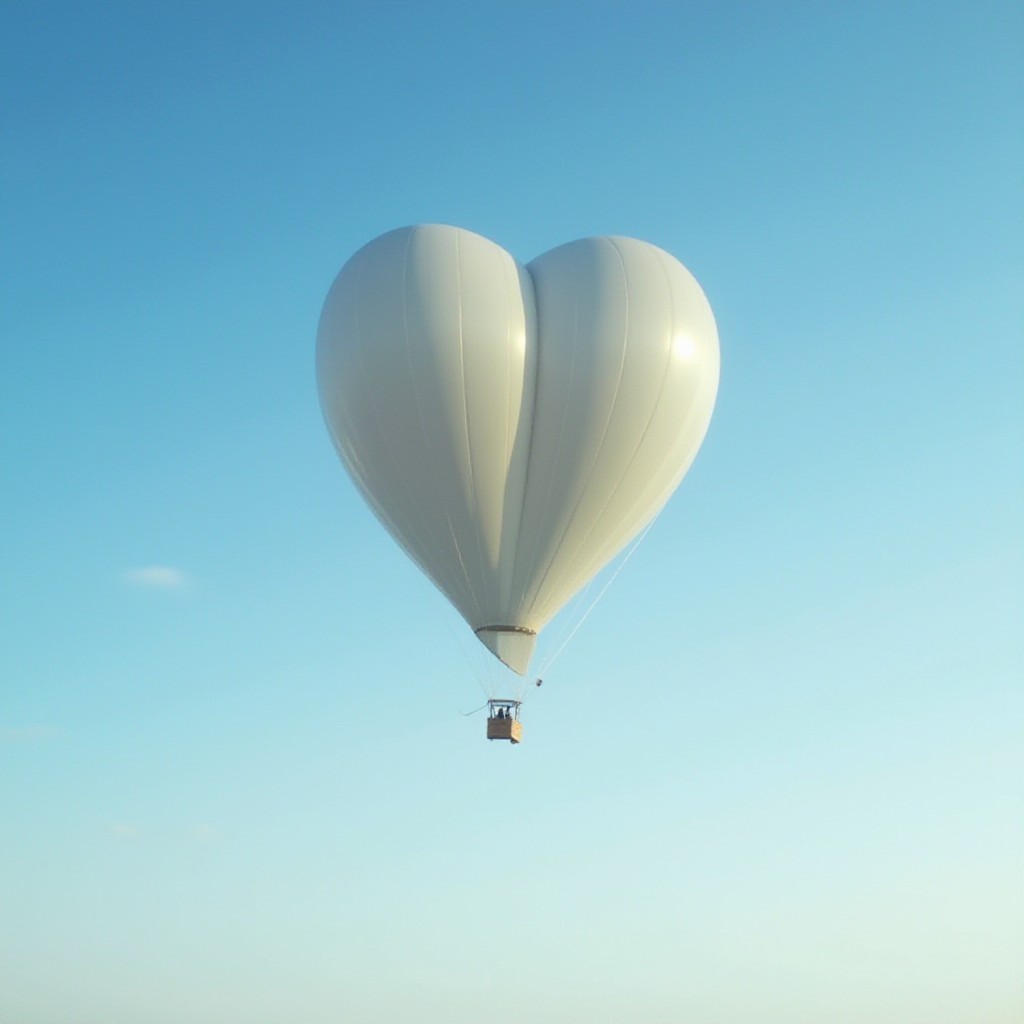}
    \end{subfigure}
    \begin{subfigure}[b]{0.28\textwidth}
        \includegraphics[width=\linewidth]{figures/show/flux/d9.jpg}
    \end{subfigure}
    \caption{Qualitative comparison between \textbf{FLUX} (first row) and \textbf{FLUX-DCW} (second row) using \textbf{10 steps}, where the prompt is ``A balloon gently climbs into a serene blue sky".}
    \label{fig:flux_show6}
\end{figure*}

\begin{figure*}[h] 
    \centering
    \begin{subfigure}[b]{0.28\textwidth} 
        \includegraphics[width=\linewidth]{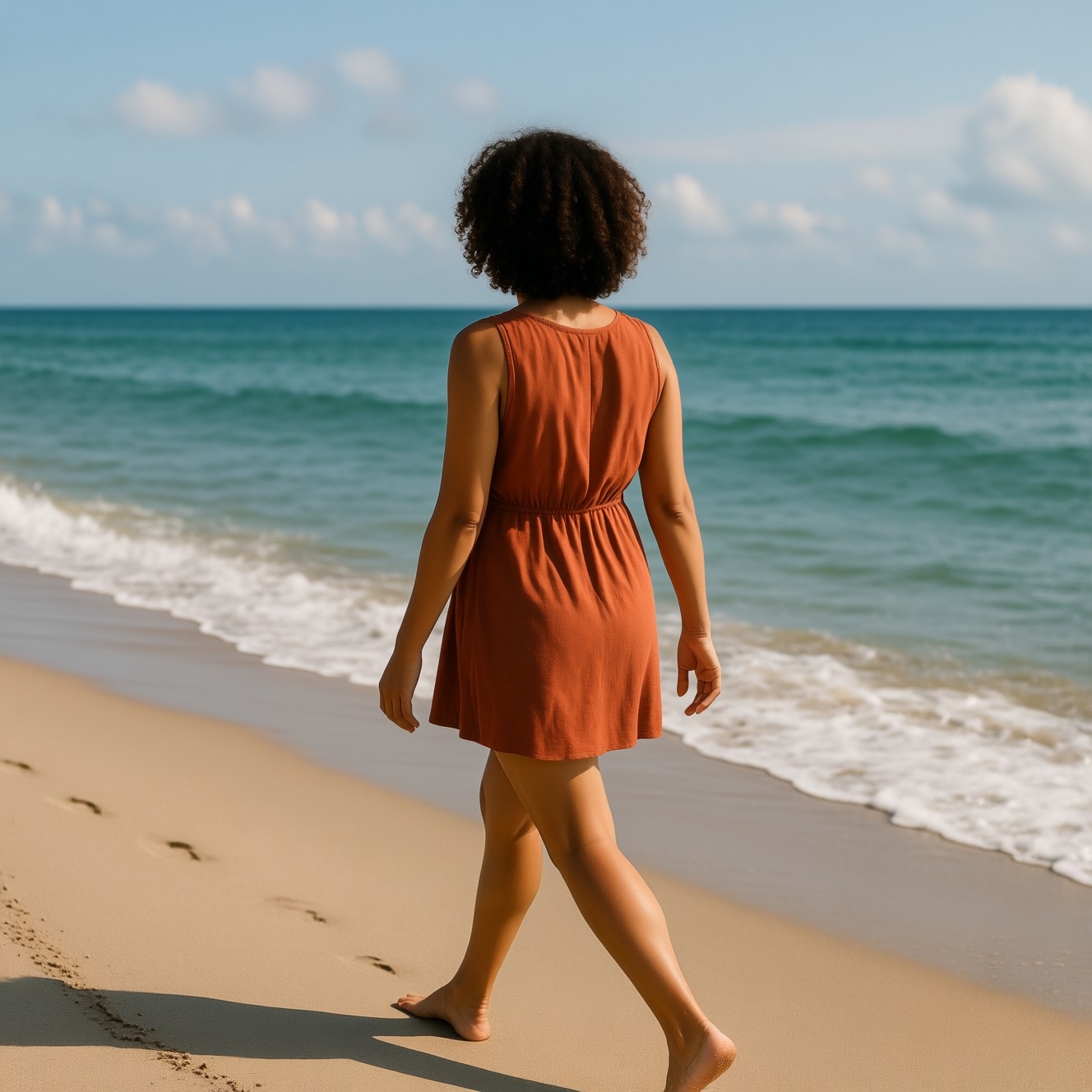}
    \end{subfigure}
    \begin{subfigure}[b]{0.28\textwidth} 
        \includegraphics[width=\linewidth]{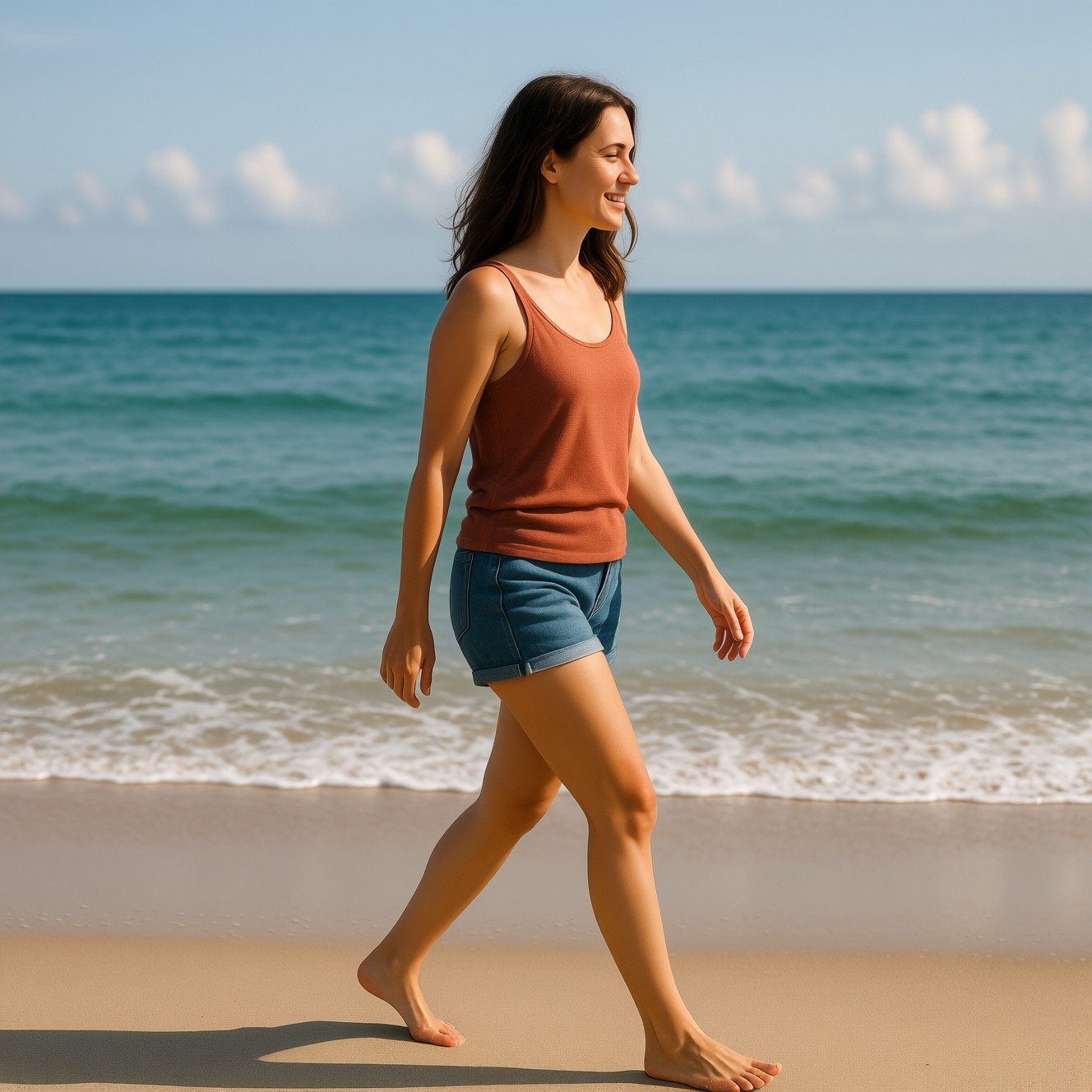}
    \end{subfigure}
    \begin{subfigure}[b]{0.28\textwidth} 
        \includegraphics[width=\linewidth]{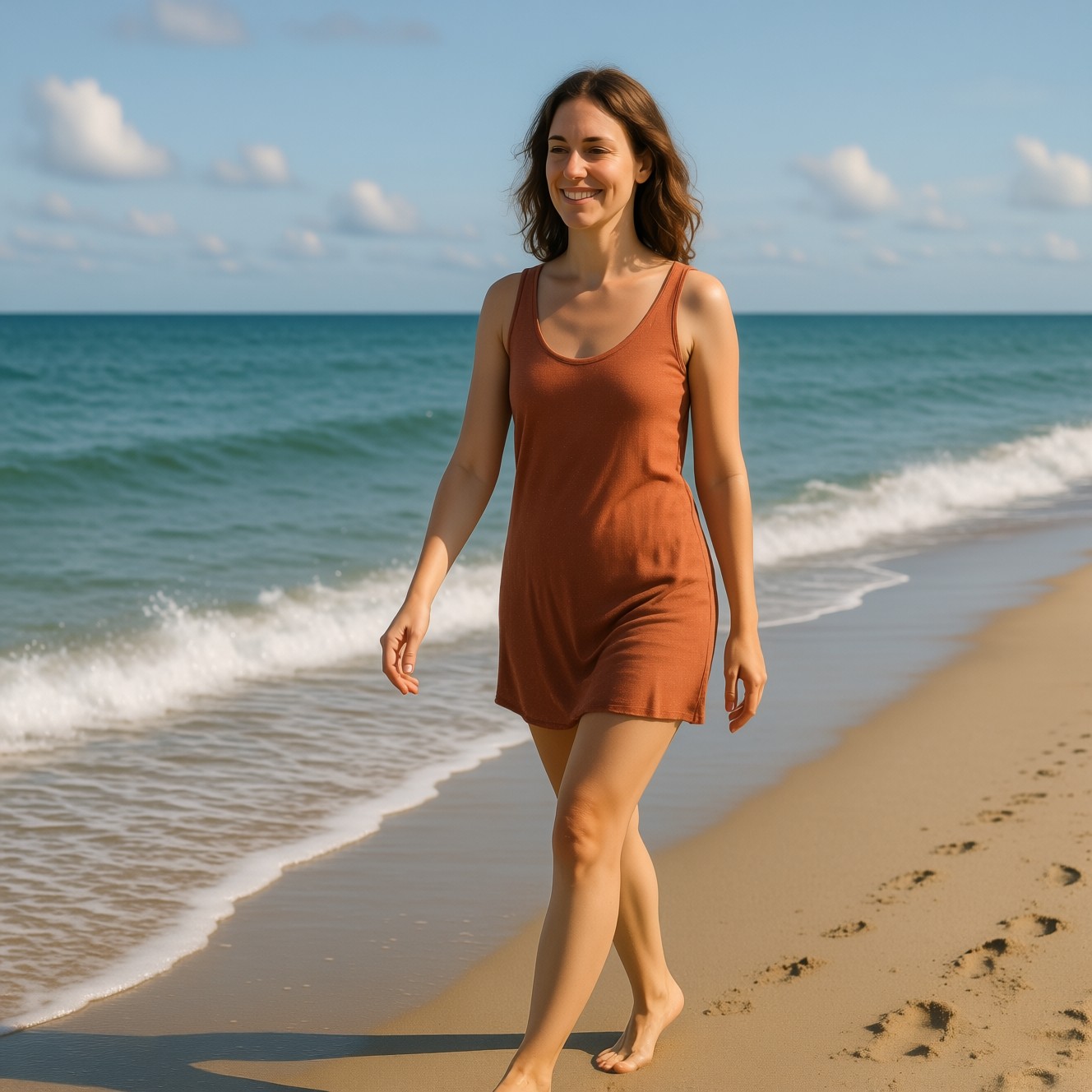}
    \end{subfigure}
    \begin{subfigure}[b]{0.28\textwidth}
        \includegraphics[width=\linewidth]{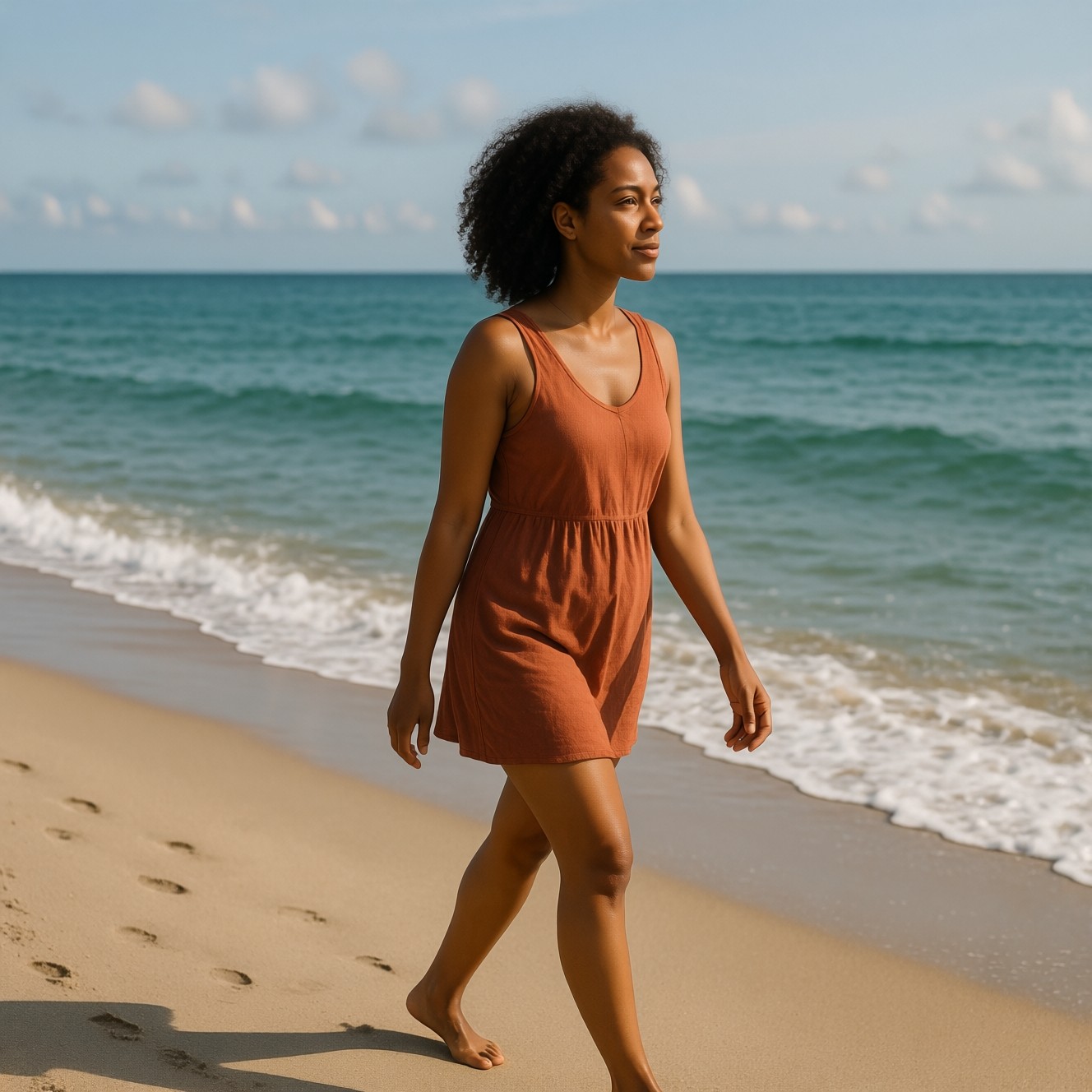}
    \end{subfigure}
    \begin{subfigure}[b]{0.28\textwidth}
        \includegraphics[width=\linewidth]{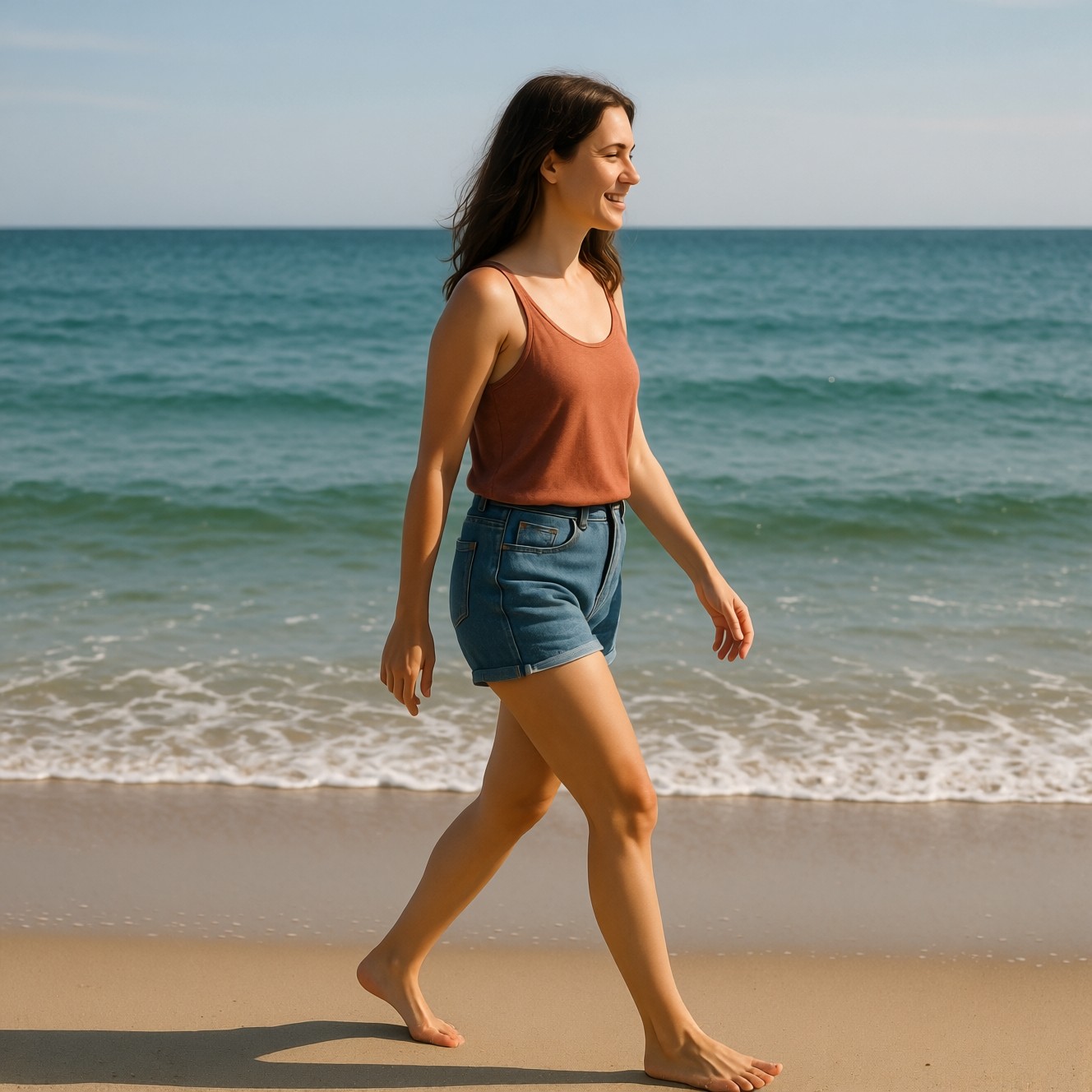}
    \end{subfigure}
    \begin{subfigure}[b]{0.28\textwidth}
        \includegraphics[width=\linewidth]{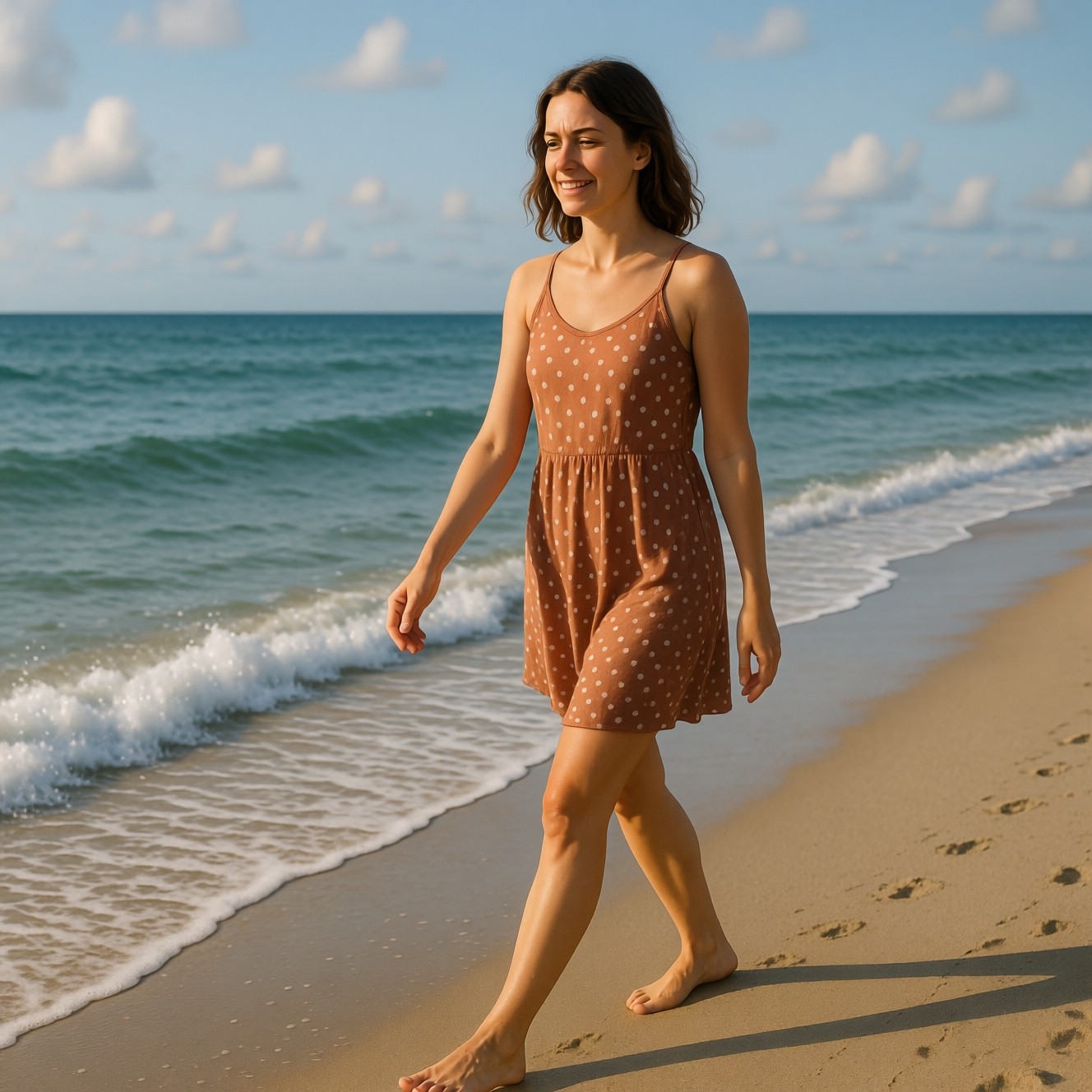}
    \end{subfigure}
    \caption{Qualitative comparison between \textbf{Qwen-Image} (first row) and \textbf{Qwen-Image-DCW} (second row) using \textbf{20 steps}, where the prompt is ``A woman is walking on the beach by the sea".}
    \label{fig:qwen_show7}
\end{figure*}

\begin{figure*}[h] 
    \centering
    \begin{subfigure}[b]{0.28\textwidth} 
        \includegraphics[width=\linewidth]{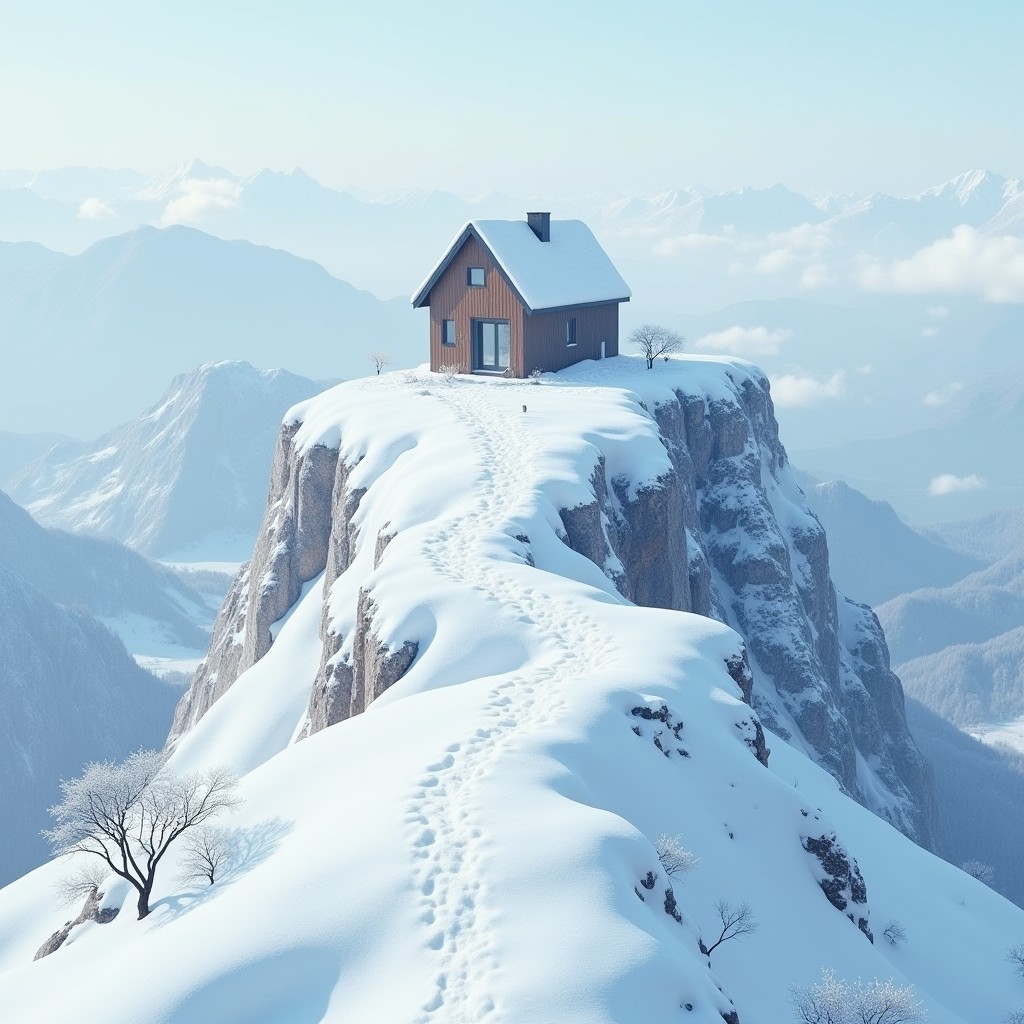}
    \end{subfigure}
    \begin{subfigure}[b]{0.28\textwidth} 
        \includegraphics[width=\linewidth]{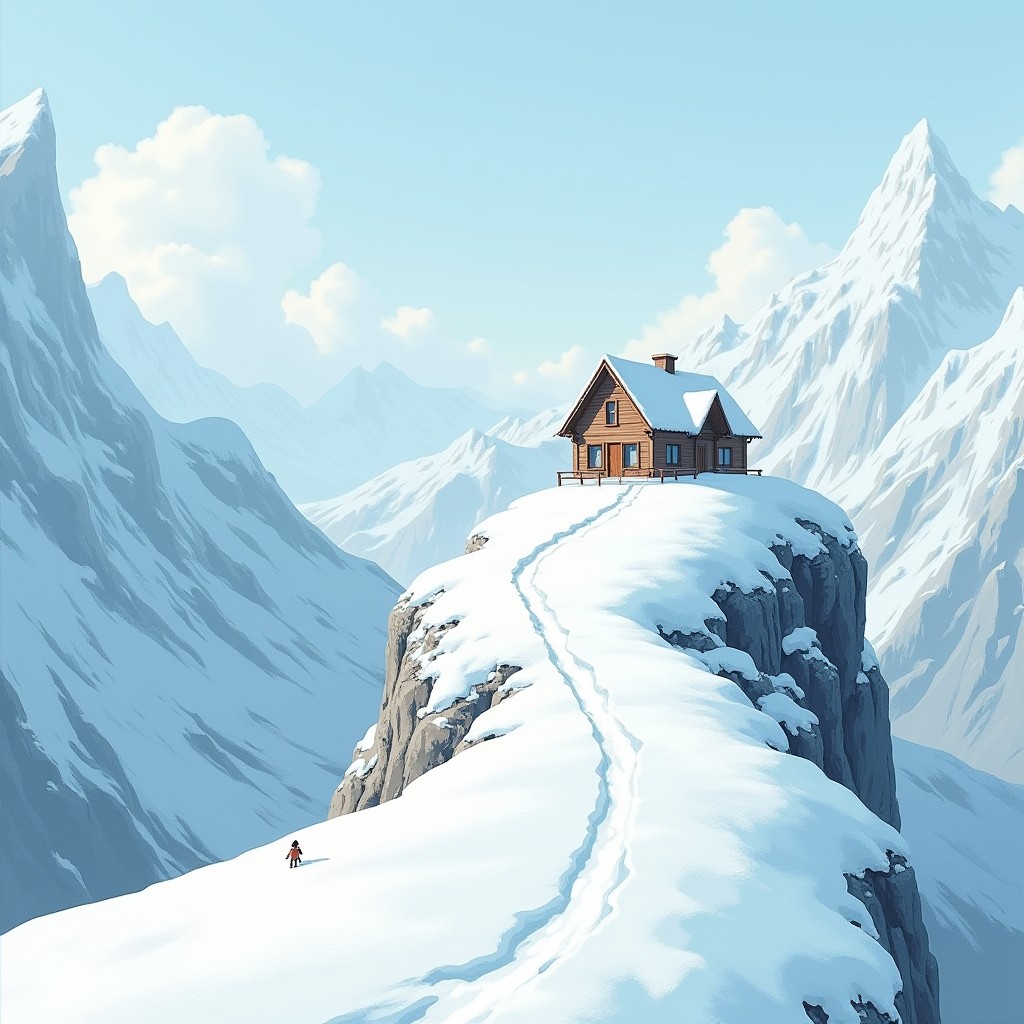}
    \end{subfigure}
    \begin{subfigure}[b]{0.28\textwidth} 
        \includegraphics[width=\linewidth]{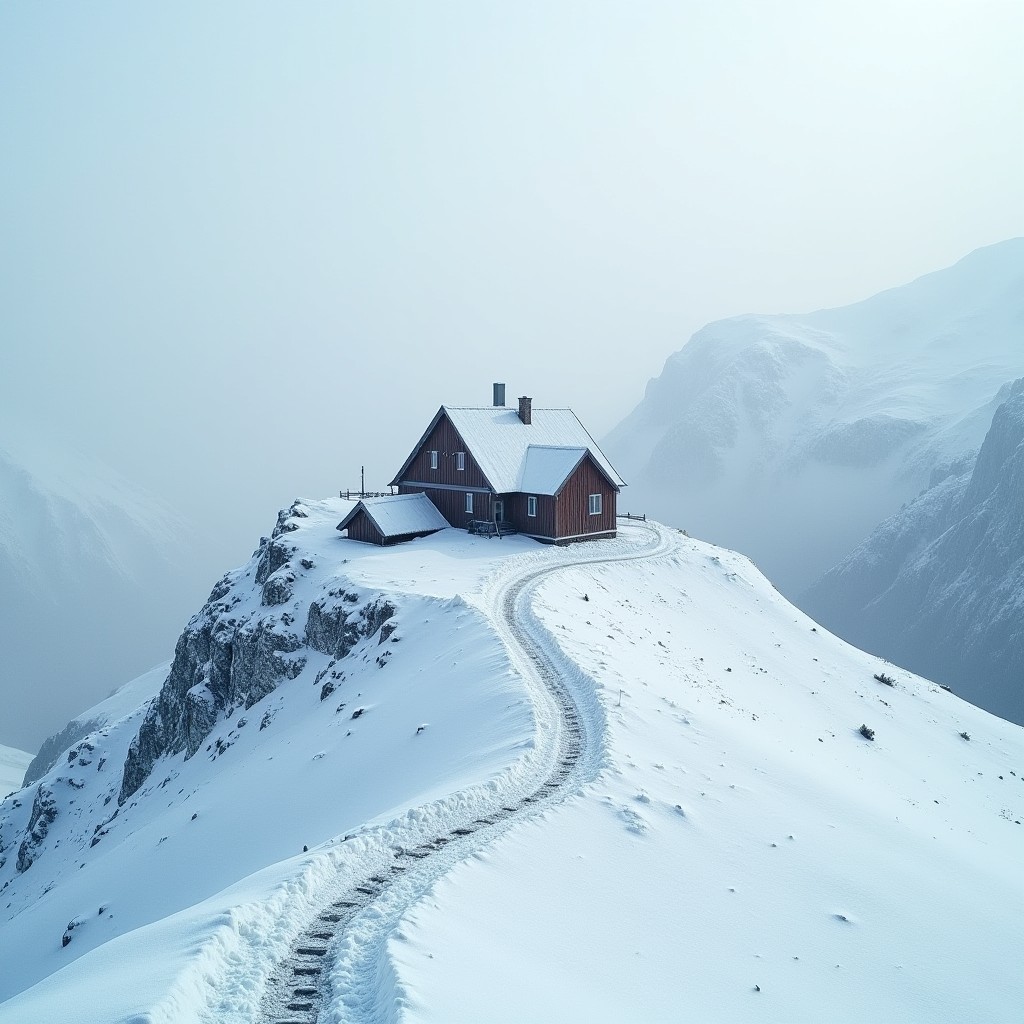}
    \end{subfigure}
    \begin{subfigure}[b]{0.28\textwidth}
        \includegraphics[width=\linewidth]{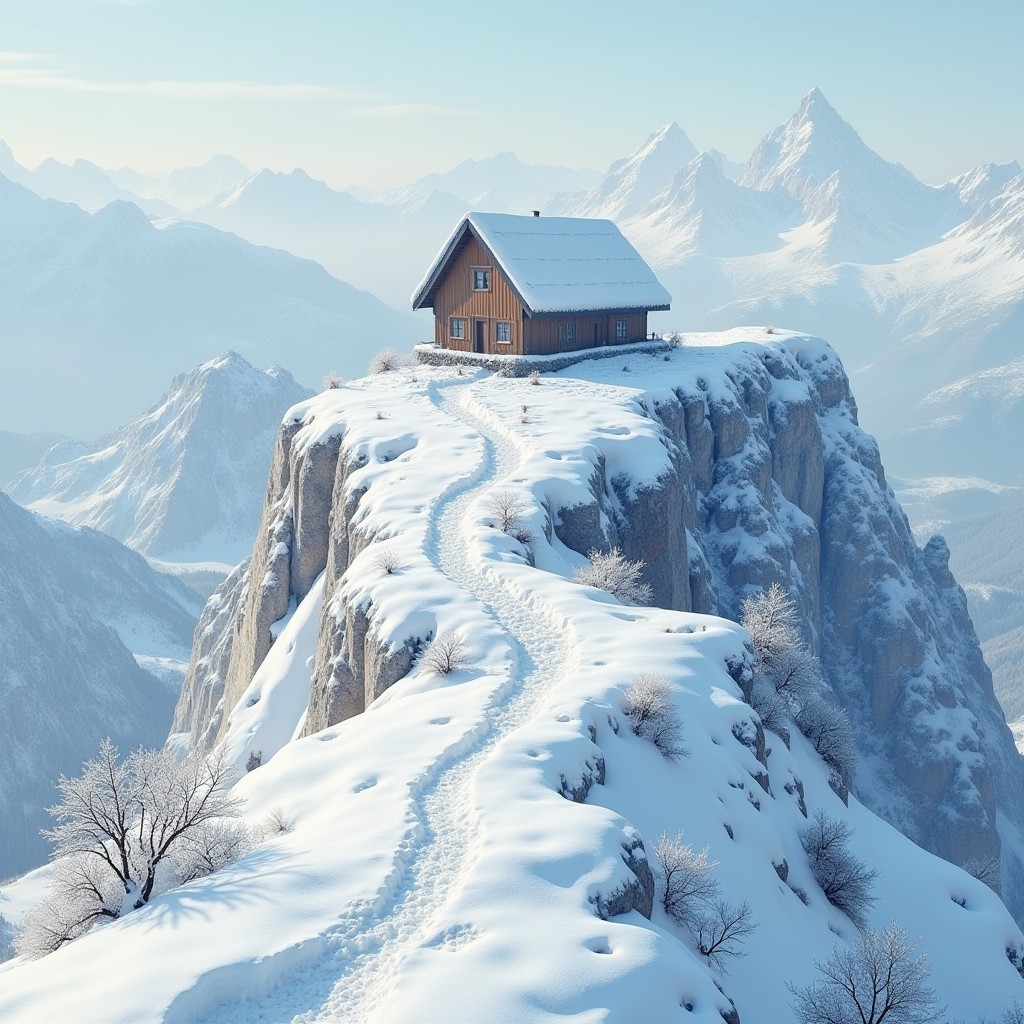}
    \end{subfigure}
    \begin{subfigure}[b]{0.28\textwidth}
        \includegraphics[width=\linewidth]{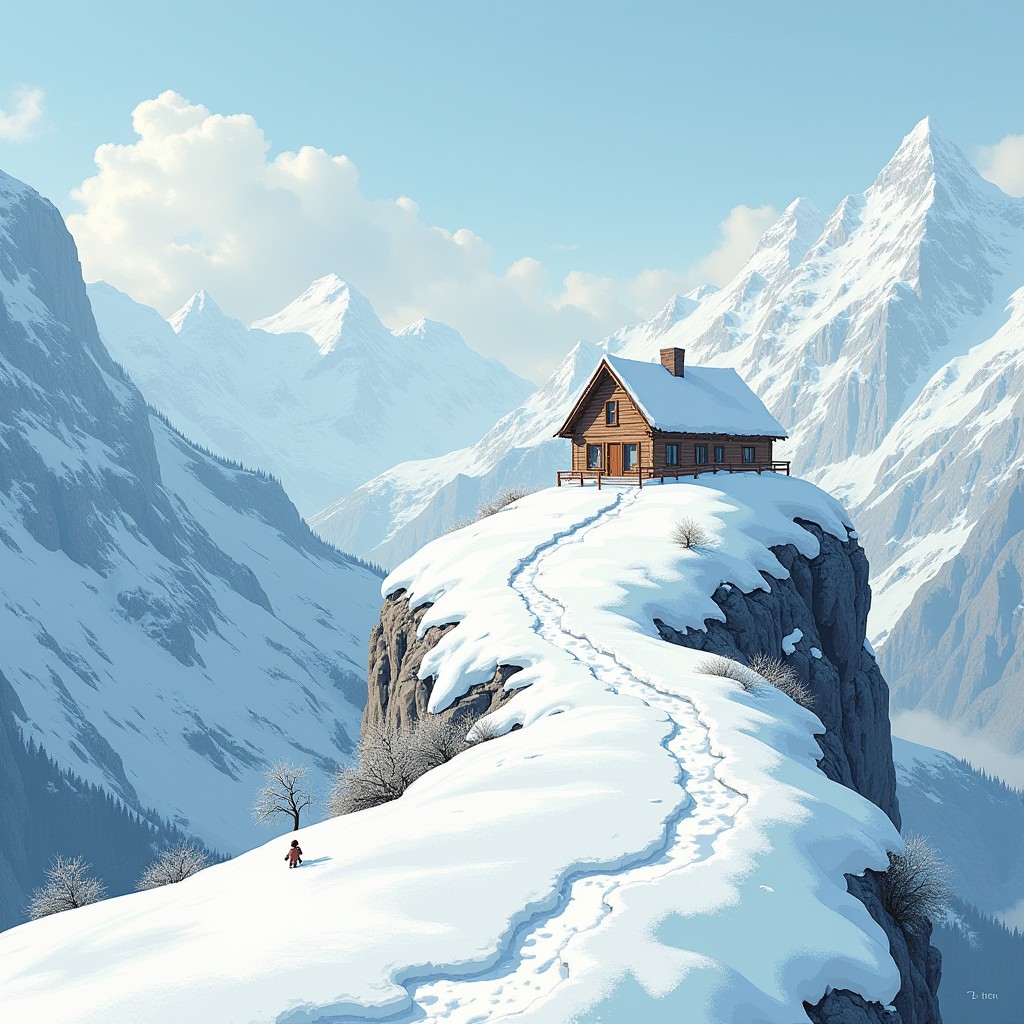}
    \end{subfigure}
    \begin{subfigure}[b]{0.28\textwidth}
        \includegraphics[width=\linewidth]{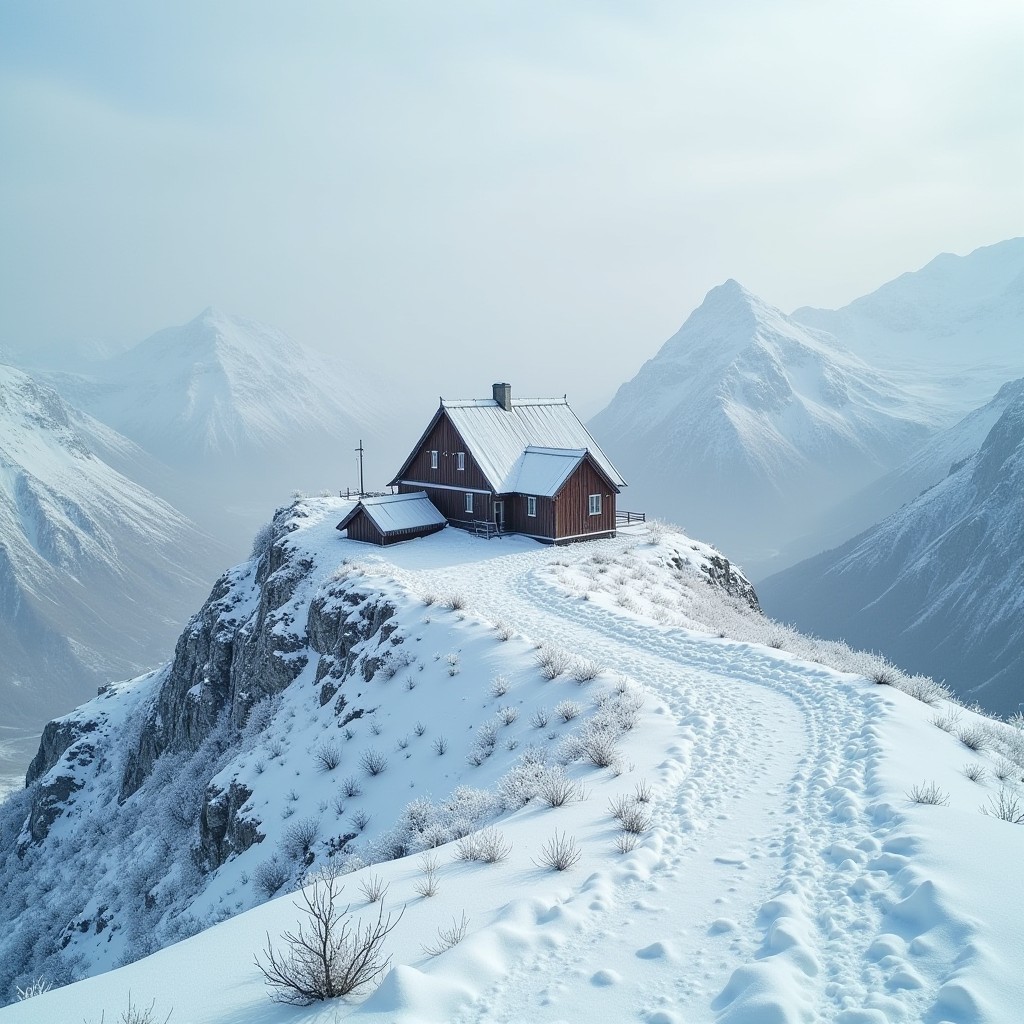}
    \end{subfigure}
    \caption{Qualitative comparison between \textbf{FLUX} (first row) and \textbf{FLUX-DCW} (second row) using \textbf{20 steps}, where the prompt is ``There is a house and a path on a snowy mountain".}
    \label{fig:flux_show8}
\end{figure*}

\begin{figure*}[h] 
    \centering
    \begin{subfigure}[b]{0.28\textwidth} 
        \includegraphics[width=\linewidth]{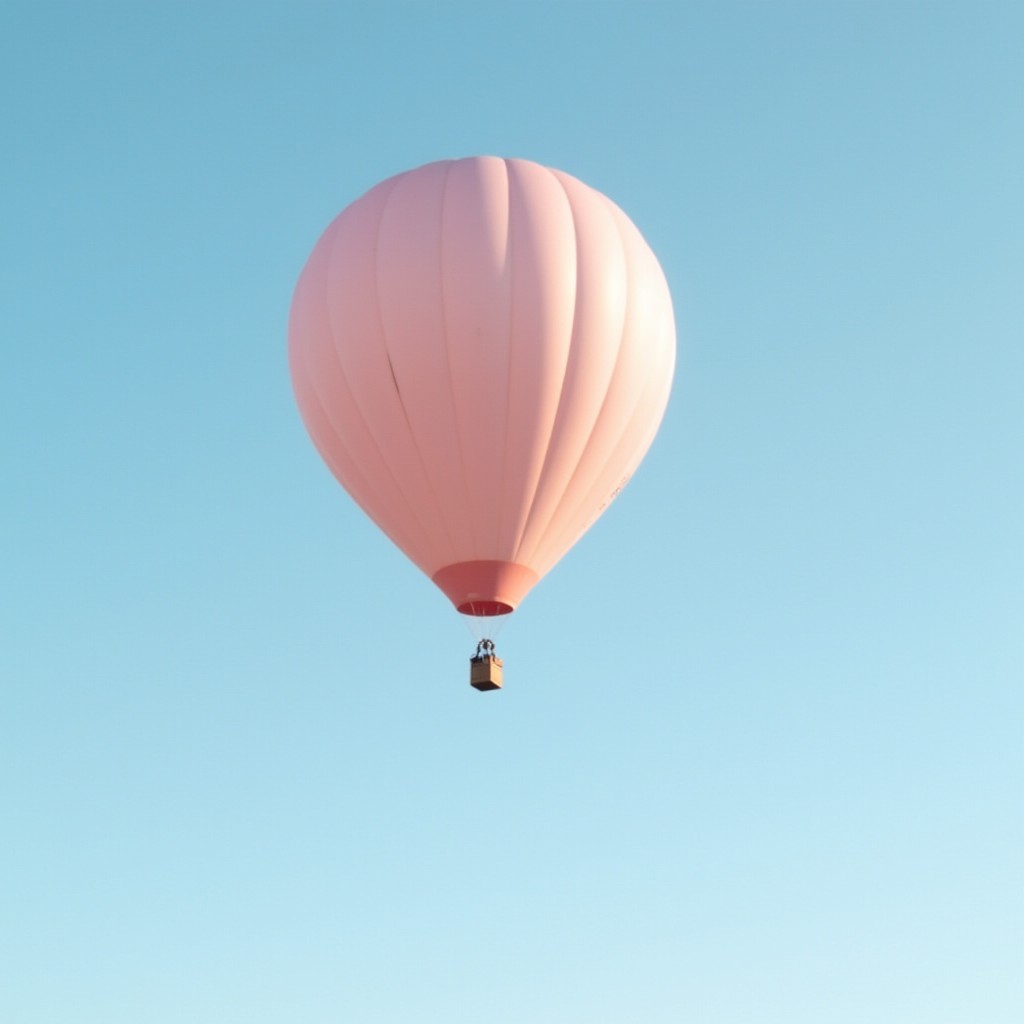}
    \end{subfigure}
    \begin{subfigure}[b]{0.28\textwidth} 
        \includegraphics[width=\linewidth]{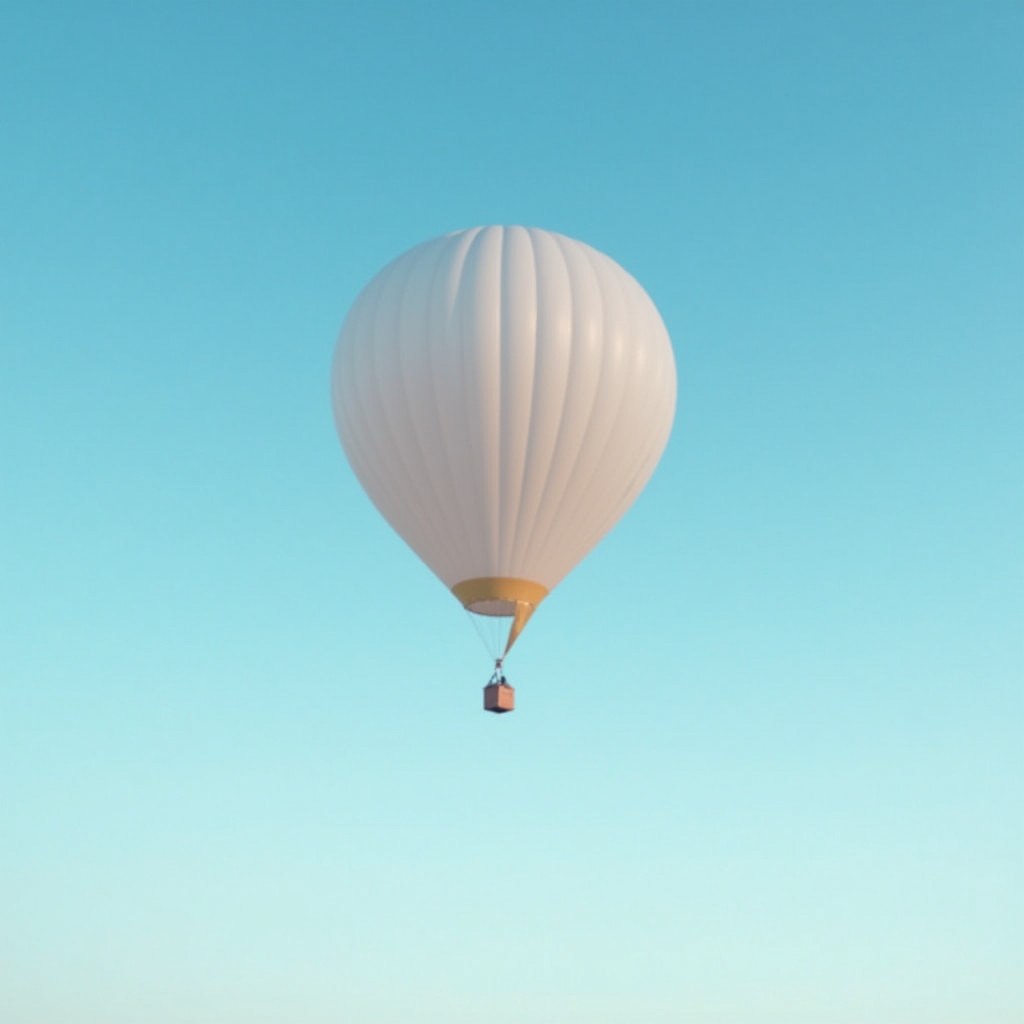}
    \end{subfigure}
    \begin{subfigure}[b]{0.28\textwidth} 
        \includegraphics[width=\linewidth]{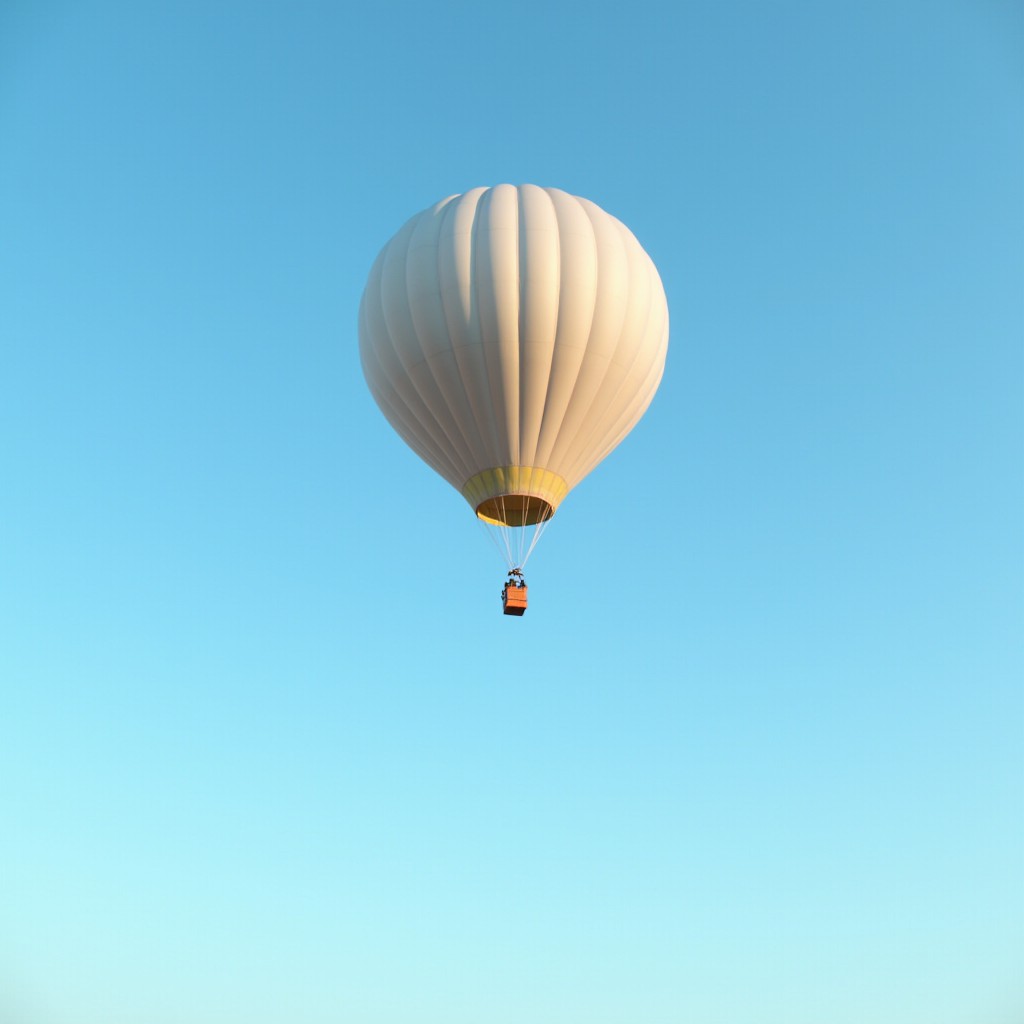}
    \end{subfigure}
    \begin{subfigure}[b]{0.28\textwidth}
        \includegraphics[width=\linewidth]{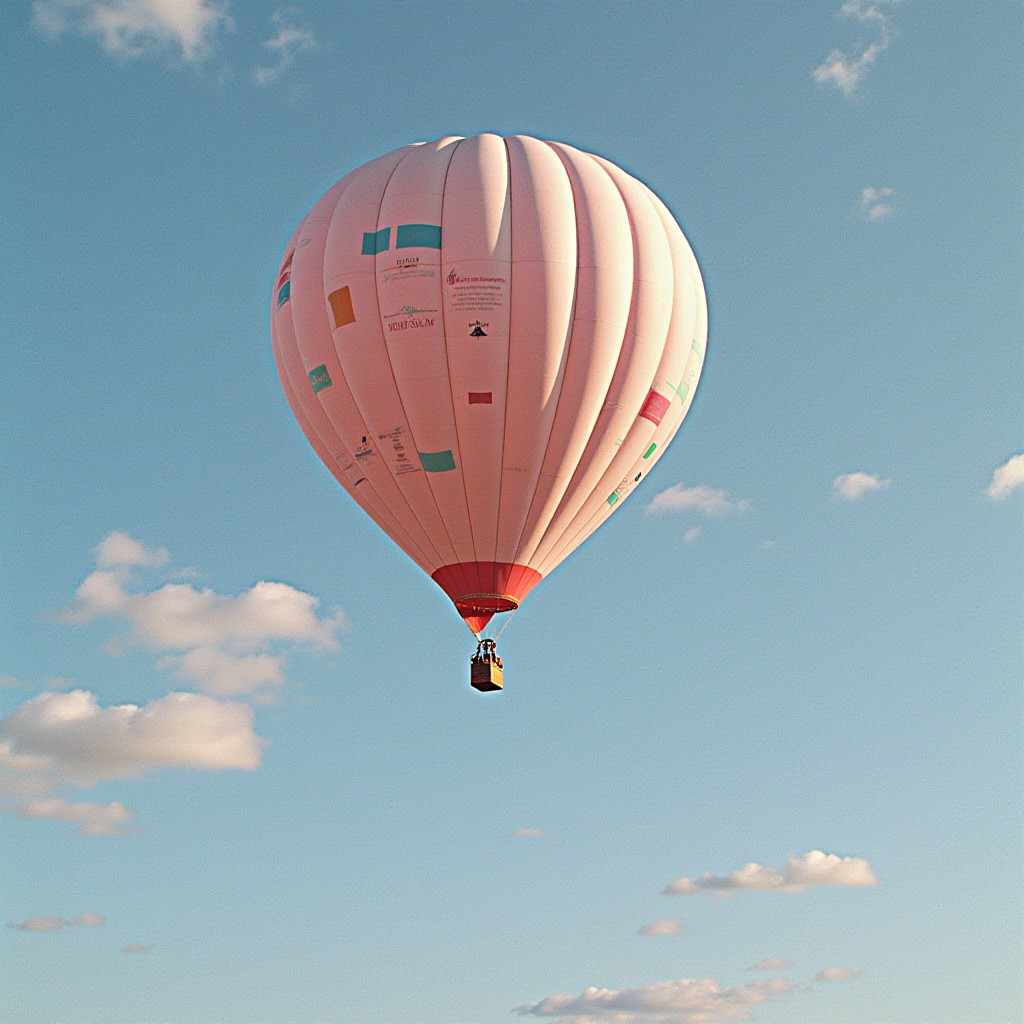}
    \end{subfigure}
    \begin{subfigure}[b]{0.28\textwidth}
        \includegraphics[width=\linewidth]{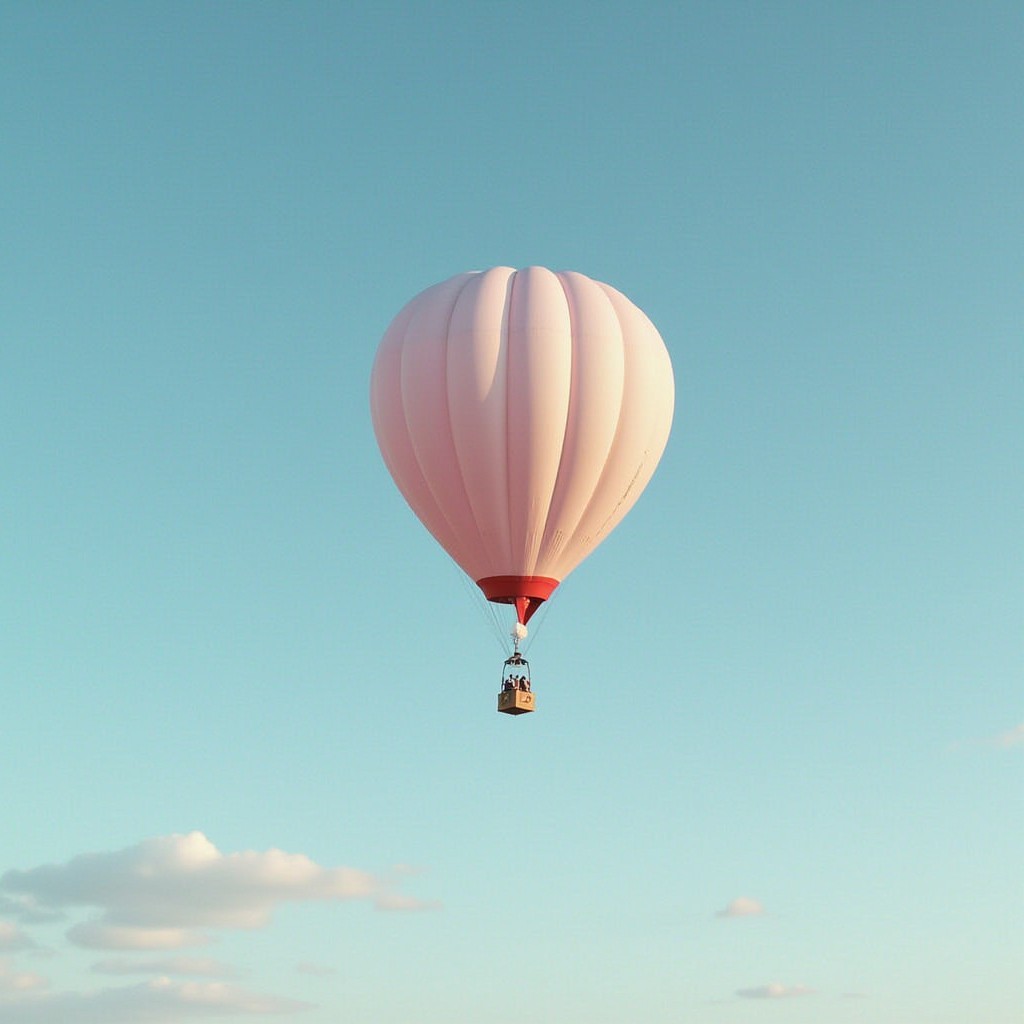}
    \end{subfigure}
    \begin{subfigure}[b]{0.28\textwidth}
        \includegraphics[width=\linewidth]{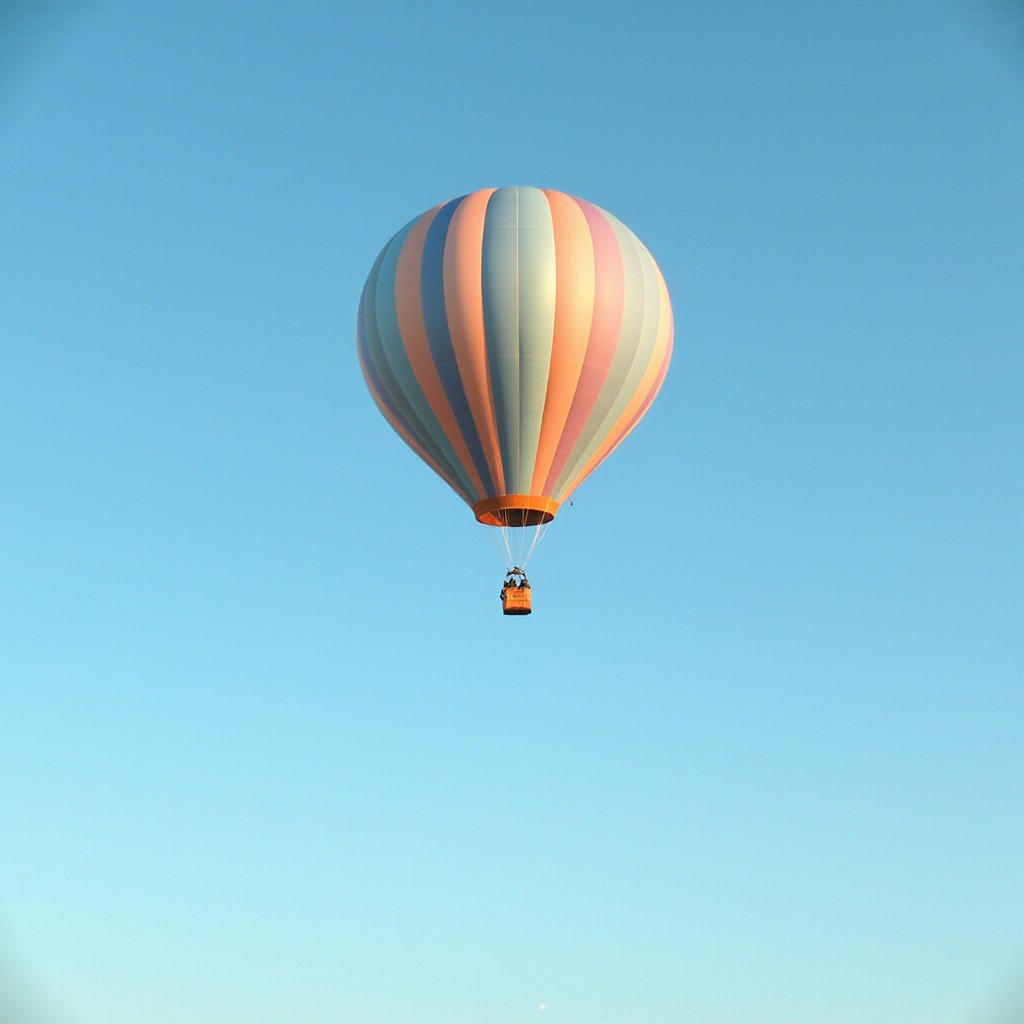}
    \end{subfigure}
    \caption{Qualitative comparison between \textbf{FLUX} (first row) and \textbf{FLUX-DCW} (second row) using \textbf{20 steps}, where the prompt is ``A balloon gently climbs into a serene blue sky".}
    \label{fi".}
    \label{fig:flux_show9}
\end{figure*}

\section{Qualitative Comparison}
\label{app:f}
To show the improvement effect of DCW on the generation quality of DPMs, we select two \textbf{state-of-the-art} text-to-image models, namely \textbf{Qwen-Image}, which demonstrates strong instruction following and text rendering ability, and \textbf{FLUX}, which is known for its high visual fidelity, to conduct extensive experiments. Given that our study focuses on the SNR-t bias, we conduct tests with a small number of steps to amplify the sampling errors of the baseline models as much as possible, thereby verifying how effectively DCW corrects such bias. As shown in Figs. \ref{fig:qwen_show1}, \ref{fig:qwen_show2}, \ref{fig:qwen_show3}, \ref{fig:flux_show4}, \ref{fig:flux_show5}, \ref{fig:flux_show6}, \ref{fig:qwen_show7}, \ref{fig:flux_show8}, and \ref{fig:flux_show9}, our method can significantly enhance the aesthetic quality across different models and time steps.

Specifically, as shown in Figs.~\ref{fig:qwen_show1} and~\ref{fig:flux_show4}, our method consistently improves the visual quality of the generated images under a small number of sampling steps. Compared with the original models, our method produces results with more coherent scene structure, better semantic fidelity, and clearer details. It also alleviates common artifacts caused by sampling bias, leading to images that are more natural and visually appealing. These results demonstrate that DCW is effective across different baseline models and can reliably enhance generation quality in low-step sampling settings. Moreover, the improvements are consistently observed across diverse scenes and content types, further highlighting the robustness and generality of our method.

\section{Parameter sensitivity}
\label{app:g}
To demonstrate the insensitivity of DCW to hyperparameters $\la_l$ and $\la_h$, we first apply DCW to A-DPM to obtain the optimal parameter $\la_l$ on CIFAR-10 (CS). Then, based on the optimal parameter $\la_l$, we apply DCW to obtain the optimal parameter $\la_h$. Fig.~\ref{fig4:parameter} clearly shows that DCW can achieve performance gains over a wide range of $\la_l$ and $\la_h$, indicating the insensitivity of DCW to hyperparameters. 

Benefiting from the strong robustness of the proposed method to hyperparameter perturbations, the parameter search process is fast via the two-stage search. Firstly, a coarse search with a step size of 0.01 was performed. After identifying a turning point in the FID curve around 0.05, we conducted a fine-grained search with a step size of 0.001 and quickly determined the optimal value to be 0.052, as shown in Tab~\ref{tab:search}. Then, after fixing the optimal $\la^*_l$ at 0.052, quickly derive the optimal parameter $\la^*_h=0.010$ using the same method. In summary, the above experimental process further demonstrates the robustness and practicality of our method with respect to hyperparameters.

\begin{table}[t]
\centering
\caption{The search process of $\la_l$ and $\la_h$ on CIFAR-10 (CS) using A-DPM-DCW with 25 sampling steps.}
\small
\setlength{\tabcolsep}{3pt} % 可微调列间距
\begin{tabular}{@{}l c c c c c c c c c c@{}} % 1+8列，消除两侧空白
\toprule
Value & 0.02 & 0.03 & 0.04 & 0.05 & 0.06 & 0.07 & 0.08 & \\
\midrule
FID & 7.64 & 7.37 & 7.24 & 7.18 & 7.19 & 7.35 & 7.66 & \\
\bottomrule
\end{tabular}
\label{tab:search}
\end{table}

\end{document}